\documentclass[twoside,11pt]{article}

\usepackage{jmlr2e}

\usepackage{amsmath,amssymb,url,braket,bbm}
\usepackage{hyperref}
\usepackage{mathrsfs}
\usepackage{overpic}
\usepackage{tikz}

\usepackage{adjustbox}

\usepackage{comment}
\usepackage{subfigure}
\usepackage{mathtools}
\usepackage{etoolbox}
\AtBeginEnvironment{dcases}{\selectfont}
\usepackage{enumitem}
\usepackage{graphicx}
\usepackage{algorithm}
\usepackage{booktabs}
\usepackage{algpseudocode}
\usepackage{soul}
\usepackage{xcolor}
\DeclareMathOperator{\E}{\mathbb{E}}

\newcommand{\as}{\overset{\mathrm{a.s.}}{\longrightarrow}}
\DeclareMathOperator*{\argmax}{arg\,max}
\DeclareMathOperator*{\argmin}{arg\,min}

\newtheorem{assumption}{Assumption}

\allowdisplaybreaks[1]

\usepackage{lastpage}
\jmlrheading{25}{2024}{1-\pageref{LastPage}}{9/22; Revised
3/24}{7/24}{22-1088}{Junwen Yang, Zixin Zhong and Vincent Y. F. Tan}

\ShortHeadings{Optimal Clustering with Bandit Feedback}{Yang, Zhong, and Tan}

\firstpageno{1}

\begin{document}

\title{Optimal Clustering with Bandit Feedback}

\author{\name Junwen Yang \email junwen\_yang@u.nus.edu\\
       \addr Institute of Operations Research and Analytics\\
       National University of Singapore\\
       117602, Singapore 
       \AND
       \name Zixin Zhong \email zixinzhong@hkust-gz.edu.cn \\
       \addr Thrust of Data Science and Analytics\\
       Hong Kong University of Science and Technology (Guangzhou)\\
       511453, Guangzhou, Guangdong, China
       % \name Zixin Zhong \email zixin.zhong@u.nus.edu \\
       % \addr Department of Computing Science\\
       % University of Alberta\\
       % T6G 2E8, Edmonton, AB, Canada
       \AND
       \name Vincent Y. F. Tan \email vtan@nus.edu.sg \\
       \addr Department of Mathematics\\
        Department of Electrical and Computer Engineering\\
        Institute of Operations Research and Analytics\\
       National University of Singapore\\
       119076, Singapore 
       }

\editor{Chris Wiggins}

\maketitle
\begin{abstract}%   <- trailing '%' for backward compatibility of .sty file
This paper considers the problem of online clustering with bandit feedback. A set of arms (or items) can be partitioned into various groups that are unknown. Within each group, the observations associated to each of the arms follow the same distribution with the same mean vector. At each time step, the agent queries or pulls an arm and obtains an independent  observation from the distribution it is associated to. Subsequent pulls depend on previous ones as well as the previously obtained samples. The agent's task is to uncover the underlying partition of the arms with the least number of arm pulls and with a probability of error not exceeding a prescribed constant $\delta$. The problem proposed finds numerous applications from clustering of variants of viruses to online market segmentation. We present an instance-dependent information-theoretic lower bound on the expected sample complexity for this task, and design a computationally efficient and asymptotically optimal algorithm, namely \textsc{Bandit Online Clustering}  (\textsc{BOC}). The algorithm includes a novel stopping rule for adaptive sequential testing that circumvents the need to exactly solve any NP-hard weighted clustering problem as its subroutines.  We show through extensive simulations on synthetic and real-world datasets that \textsc{BOC}'s performance matches the lower bound asymptotically, and significantly outperforms a non-adaptive baseline algorithm. 
%\st{that samples each arm uniformly at each time}. 
\end{abstract}

\begin{keywords}
  clustering, $K$-means, online learning, multi-armed bandits, pure exploration
\end{keywords}

\section{Introduction}
\label{section_intro}
Clustering, the task of partitioning a set of items into smaller clusters based on their commonalities, is one of the most fundamental tasks in data analysis and machine learning with a rich and diverse history \citep{driver1932quantitative, cattell1943description, ruspini1969new, jain1999data, celebi2013comparative}. It has numerous applications in a wide variety of areas including business analytics, bioinformatics, pattern recognition, and social sciences.
In this era of abundance of medical data, clustering is a powerful tool to uncover the underlying patterns of unknown treatments or diseases when related systematic knowledge is underdeveloped \citep{maccuish2010clustering}. In commercial decision making, aiming at increasing customer satisfaction and maximizing  potential benefit, marketers utilize clustering strategies to partition the inclusive business market into more narrow market segments with similar characteristics \citep{chaturvedi1997feature}. Due to its enormous importance in practical applications, clustering has been studied extensively in the literature from multidisciplinary perspectives. A plethora of algorithms (e.g., $K$-means and spectral clustering) have been proposed for the task of clustering (see \citet{jain1999data} or \citet{saxena2017review} for  comprehensive reviews). In particular, the $K$-means algorithm by \citet{macqueen1967some} and \citet{lloyd1982least} is arguably the most ubiquitous algorithm due to its simplicity, efficiency, and empirical successes \citep{jain2010data}. 

Although modern data analysis has benefited immensely from the abundance and richness of data, there is an urgent need to develop new techniques that are adapted to the sequential and uncertain nature of data collection. In this paper, we are interested in {\em online clustering with bandit feedback}, which is an online variant of the classical offline clustering problem. With bandit feedback, the agent only observes a noisy measurement on the selected arm (or item) at each time step. However, the agent can  decide which arm to pull adaptively, so as to minimize the expected number of total arm pulls it takes to correctly partition the given arm set with a given (high) probability. 
 
\paragraph{Two Motivating Examples.} Our online clustering model captures various contemporary real-world scenarios, in which data contaminated by some degree of measurement noise become available in a sequential and adaptive fashion. We are firstly motivated by medical and public health professionals' arduous battles against new viruses (e.g., COVID-19). In the face of an unknown type of virus that has different variants, let us assume that there is only  one dominant variant in each sub-region of a particular state. When accurate laboratory analysis is not available especially in  underdeveloped regions, how can   healthcare professionals partition the virus samples into  specific dominant variants based on    noisy measurements of  infectious patients from various sub-regions? This realistic and critical problem can be well modelled  by our framework, namely online clustering with bandit feedback, where the healthcare personnel adaptively obtains independent observations of  patients from the selected sub-regions and finally partitions the whole state into various groups based on the types of dominant variants. Due to the prohibitive costs  in obtaining the measurements, this must be done with as few of them as possible.  

\begin{figure}[t]

\begin{minipage}{1\textwidth}
\centering
%\vspace{0.25cm}
\begin{overpic}[width=1\textwidth]{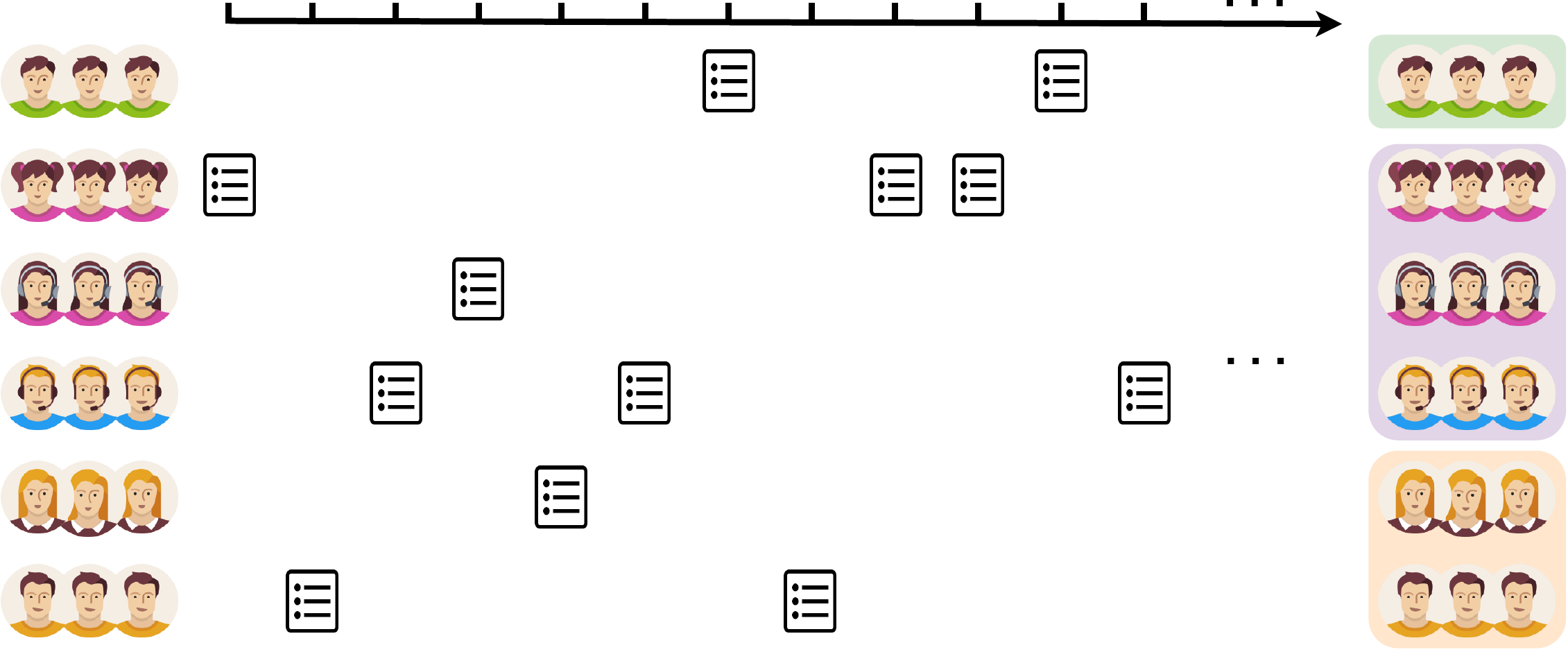}
\put(83.7,42){Time} % from: left, bottom
\put(14,42){$1$} % from: left, bottom
\put(19.3,42){$2$} % from: left, bottom
\put(24.6,42){$3$} % from: left, bottom
\put(29.9,42){$4$} % from: left, bottom
\put(35.2,42){$5$} % from: left, bottom
\put(40.5,42){$6$} % from: left, bottom
\put(45.8,42){$7$} % from: left, bottom
\put(51.1,42){$8$} % from: left, bottom
\put(56.4,42){$9$} % from: left, bottom
\put(61.1,42){$10$} % from: left, bottom
\put(66.4,42){$11$} % from: left, bottom
\put(71.7,42){$12$} % from: left, bottom
\end{overpic}
%\vspace{0.05cm} %add space
\end{minipage}
\caption{Example involving partitioning $6$ sub-groups of customers into $3$ market segments with bandit feedback. {Initially, customers are preliminarily divided into sub-groups, treated as arms, based on some basic characteristics such as age and gender. At each time step, an algorithm chooses a sub-group to query and receives a multidimensional sample (e.g., product ratings) from a single customer within that sub-group.} Based on the previously chosen sub-groups and their samples, the algorithm decides which sub-group to query next. Finally, when it is sufficiently confident of producing a partition of the sub-groups into market segments that share similar preferences, the algorithm terminates. }
\label{figure_intro}
\end{figure} 

Our second motivation is  in  digital marketing in which customer feedback on certain products are  collected in an online manner and always accompanied by random or systematic noise. For market segmentation, one important objective is to reduce the cost of feedback collection while maintaining a high quality of clustering, so that 
subsequent recommendations are  suitably tailored to particular groups of consumers. See Figure~\ref{figure_intro} for a protocol of online market segmentation that our framework is able to model well.

\paragraph{Main Contributions.} Our main contributions are as follows:
\begin{enumerate}[label = (\roman*)]
%\begin{enumerate}[label = (\roman*), noitemsep, nolistsep]
    \item  We formulate the online clustering with bandit feedback problem in Section~\ref{section_setup}. We identify some subtleties of the framework, which may be of independent interest for future research. In particular, there are multiple ways that a partition of an arm set can be represented. This is further complicated by the fact that each cluster is also identified by a mean vector. We propose a precise  expression for  an instance of cluster bandits, and establish two equivalence relations for the partitions and bandit instances, respectively. %\red{\st{By further stating some natural assumptions on the mean vectors, we identify the instance pairs that can be learned up their equivalence classes.}}
    
    \item In Section~\ref{section_lower}, we derive an instance-dependent (information-theoretic) lower bound on the expected sample complexity for the online clustering problem; this lower bound, however, involves a tricky optimization problem. By exploiting the structure of the problem and leveraging an interesting combinatorial property, we simplify the optimization to a {\em finite convex} minimax problem, which can be solved efficiently. Further analyses of the lower bound provide fundamental insights and essential tools for the design of our algorithm.
    
    \item In Section~\ref{section_algo}, we propose and analyze \textsc{Bandit Online Clustering} (or \textsc{BOC}). We show that it is not only computationally efficient but also asymptotically optimal in the sense that its expected sample complexity attains the lower bound as the error probability tends to zero. This result is somewhat surprising since  solving the corresponding \emph{offline} clustering problem exactly is   NP-hard  \citep{aloise2009np, mahajan2012planar}. En route to overcoming this combinatorial challenge and demonstrating desirable properties of \textsc{BOC}, we utilize a variant of the classical $K$-means algorithm in the sampling rule and propose a novel stopping rule for adaptive sequential testing. In particular, the ``natural'' stopping rule based on the generalized likelihood ratio (GLR) statistic \citep{chernoff1959sequential,garivier2016optimal} is intractable. Our workaround involves another statistic that exploits our insights  on the lower bound.
    Finally in Section~\ref{section_exp}, we show via numerical experiments that \textsc{BOC} is indeed asymptotically optimal and significantly outperforms a uniform sampling strategy on both synthetic and real-world benchmark datasets. 
    
\end{enumerate}

\section{Literature Review} 
\label{section_literature}
\paragraph{Clustering and the $K$-means algorithm. }Although we consider an online version of the clustering problem, a certain variant of the classical (offline) $K$-means algorithm  serves as a subroutine of  \textsc{BOC}. Given the $d$-dimensional observations of $M$ items, the $K$-means algorithm   \citep{macqueen1967some, lloyd1982least}  aims at partitioning the items into $K$ disjoint clusters in order to minimize the sum of squared Euclidean distances between each item to the center of its associated cluster. If each item is associated to a certain {\em weight}, then a reasonable  objective is the minimization of  the weighted sum of squared Euclidean distances. Due to the prominence of the \emph{$K$-means} algorithm, the corresponding clustering problem is also referred to as the \emph{(weighted) $K$-means clustering} problem in the literature. This non-convex $K$-means clustering problem has been shown to be NP-hard even for $d = 2$ \citep{mahajan2012planar} or $K=2$ \citep{aloise2009np}. In fact, for general $d$, $K$ and $M$, the problem can be exactly solved in $O(M^{dK+1})$  time \citep{inaba1994applications}. When used as a heuristic algorithm, the performance of $K$-means depends to a large extent on how it is initialized. To improve the stability and the quality of the eventual solution that   $K$-means produces, various initialization methods have been proposed (e.g., Forgy’s method \citep{forgy1965cluster}, Maximin \citep{gonzalez1985clustering}, $K$-means++ \citep{arthur2006k}, PCA-Part \citep{su2007search}). See \citet{celebi2013comparative} for detailed comparisons on initialization methods for $K$-means.

There have been some attempts in the literature to adapt the vanilla $K$-means algorithm to an online framework involving streams of incoming data \citep{choromanska2012online, liberty2016algorithm, cohen2021online}. In this line of works, each item only has {\em one} observation. In particular, at each time step $t$, the agent receives an observation of one item and has to determine its cluster index before the arrival of next observation. This is vastly different from our setting of bandit feedback, where the arm pulls are determined adaptively by the agent and the observations are stochastic.
In addition, another related work by \citet{khaleghi2012online} addressed the case where every item is associated with an infinite sequence generated by one of $K$ unknown stationary ergodic processes. At each time step $t$, some observations arrive, each being either a new sequence or the continuation of some previously observed sequence, and the agent needs to partition the observed sequences into $K$ groups. Finally, \citet{mazumdar2017clustering} studied clustering with noisy queries in both the adaptive
and non-adaptive settings, where the agent receives a potentially incorrect answer on whether two items belong to the same cluster at each time step. To the best of our knowledge, the online clustering with bandit feedback problem (formally described in Section~\ref{section_setup}) has not been considered before. 

\paragraph{Bandit Algorithms. }The stochastic multi-armed bandit problem,  originally introduced by \citet{thompson1933likelihood},  provides a simple but powerful online learning framework. This problem has been studied extensively in recent years. While the \emph{regret minimization} problem aims at maximizing the cumulative reward by balancing  the trade-off between exploration and exploitation \citep{auer2002finite, abbasi2011improved, bubeck2012regret,agrawal2012analysis}, the \emph{pure exploration} problem focuses on efficient exploration with specific goals, e.g., best (top-$k$) arm identification \citep{even2006action,audibert2010best,karnin2013almost,garivier2016optimal, jun2016top}, and its variants in linear bandits \citep{soare2014best, jedra2020optimal,yang2022minimax}, and cascading bandits \citep{zhong2020best}, among others.
Our online clustering task can also be viewed as a pure exploration problem although the rewards are \emph{multi-dimensional}  and the arm set has an inherent cluster structure.
It is worth mentioning that \citet{prabhu2020sequential} introduced a framework of sequential multi-hypothesis testing with bandit feedback, which is a generalization of the odd arm identification problem \citep{vaidhiyan2017learning}. Our online clustering problem falls within this framework if each  partition of the arm set is viewed as a hypothesis. However, the methodology proposed in \citet{prabhu2020sequential} relies heavily on a strong continuity assumption on the proportions of arm pulls (Assumption~A therein). Even if one accepts continuity assumption, the total number of hypotheses (i.e., the number of possible partitions) is prohibitively large and the corresponding computation of the \emph{modified GLR} is intractable (see  Remark~\ref{remark_odd_arm} for more details). Our work does away with the continuity assumption and, in fact, {\em proves} that the required  continuity property  holds (see  Proposition~\ref{prop_continuity2}). 
We refer to \citet{lattimore2020bandit} for a comprehensive review on bandit algorithms. 

There is also a line of works that incorporates   cluster structures into  multi-armed bandits.  In particular, \citet{nguyen2014dynamic, gentile2014online}, \citet{li2016collaborative}, and \citet{carlsson2021thompson} assumed that users  can be divided into groups and the users within each group  receive similar rewards for  each arm. Besides, \citet{bouneffouf2019optimal} and \citet{singh2020multi} assumed that the arm set is pre-clustered and the reward distributions of the arms within each cluster are similar.
However, all the works mentioned above focus on {\em leveraging} the cluster structure to improve the performance of regret minimization, which differs from our objective, i.e., to {\em uncover} the underlying partition of the arms. Finally, \citet{wang2022max} considered  a min-max grouped bandits problem in which the objective is to  find a sub-group (among possibly overlapping groups) whose worst arm has the highest mean reward.

\section{Problem Setup and Preliminaries}
\label{section_setup}
\paragraph{A Bandit Feedback Model with Cluster Structure.} We consider a bandit feedback model, in which the arm set has an inherent cluster structure. In particular, the agent is given an arm set $\mathcal A = [M]$, which can be partitioned into $K$ disjoint nonempty clusters, and the arms in the same cluster share the same $d$-dimensional mean vector, also referred to as the \emph{center} of the corresponding cluster. Without loss of generality, we assume that $K < M$, otherwise there is only one possible partition. Therefore, an instance of cluster bandits can be fully characterized by a pair $(c,\mathcal{U})$, where $c=[c_1, c_2, \ldots, c_M]\in[K]^{M}$ consists of the cluster indices of the arms and $\mathcal{U}=[\mu(1), \mu(2), \ldots, \mu(K)] \in \mathbb R^{d\times K}$ represents the $K$ centers of the clusters. Since each cluster has at least one arm, for any cluster $k \in[K]$, there exists an arm $m\in[M]$ such that $c_m = k$.  To reduce clutter and ease the reading,  we always index the arms and the clusters by subscripts and numbers in parentheses, respectively.

\begin{figure}[t]
\begin{minipage}{1\textwidth}
%\centering
\vspace{-.08in} \hspace{0.8in}
\begin{overpic}[width=.8\textwidth]{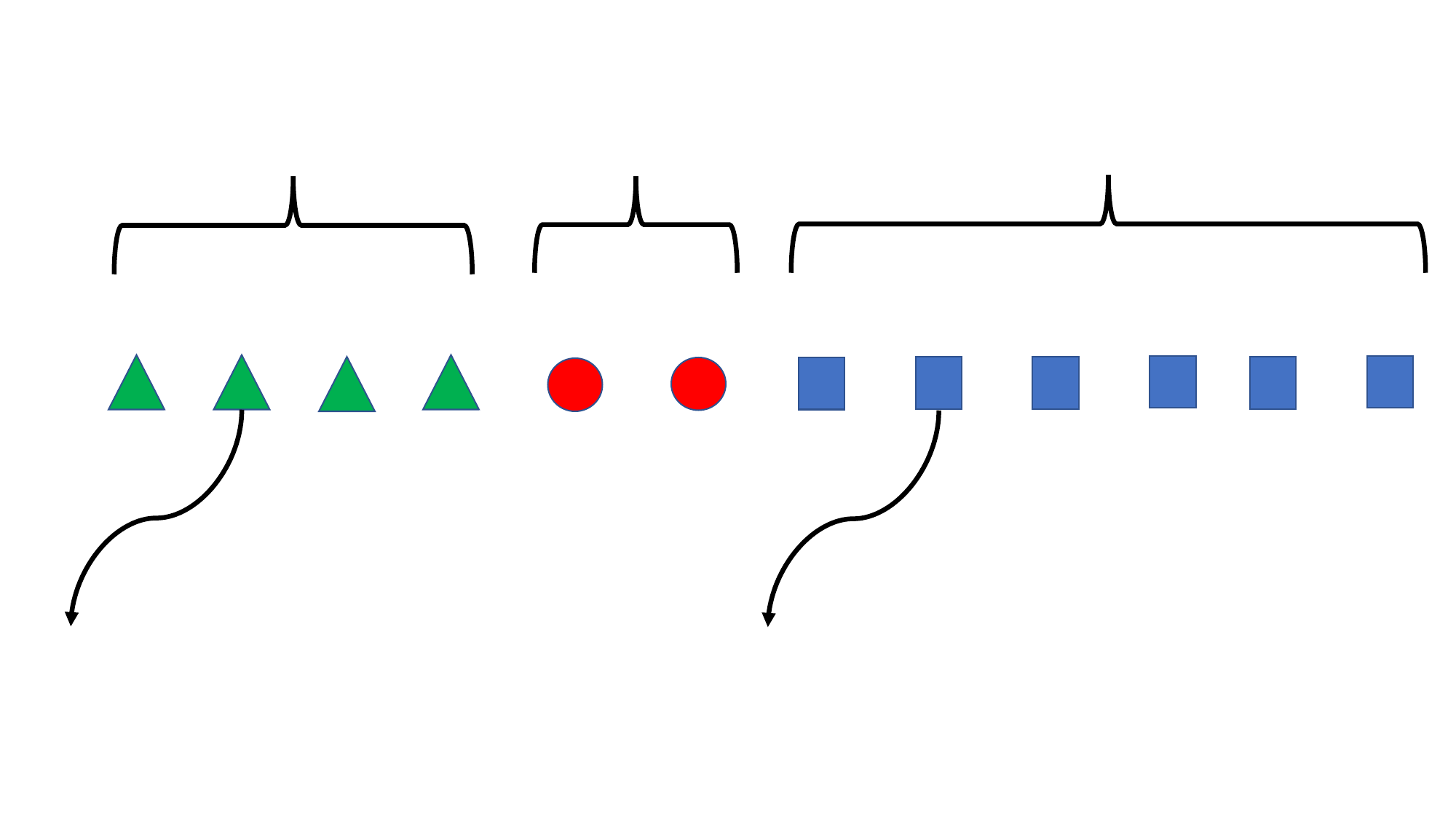}
\put(7,46){$\{m \in [M]: c_m=1\}$}
\put(34,46){$\{m\in [M]: c_m=2\}$}
\put(66,46){$\{m\in [M]: c_m=3\}$}
\put(-12,33){Arm set $[M]$:}
\put(9,33){$1$}
\put(16,33){$2$}
\put(23,33){$3$}
\put(30.5,33){$4$}
\put(39,33){$5$}
\put(47,33){$6$}
\put(56,33){$7$}
\put(64,33){$8$}
\put(72,33){$9$}
\put(79,33){$10$}
\put(86,33){$11$}
\put(94,33){$12$}
\put(-12,24){Mean vectors:}
\put(18,24){$\mu(1)$}
\put(42,24){$\mu(2)$}
\put(72,24){$\mu(3)$}
\put(0,10){If $A_t=2$, then $X_t = \mu(c_2)+\eta_t$.}
\put(46,10){If $A_{t+1}=8$,  then  $X_{t+1} = \mu(c_8)+\eta_{t+1}$.}
\end{overpic}
\vspace{-.4in}
\end{minipage}
\caption{Online clustering with bandit feedback with $K=3$ and $M=12$.} \label{fig:model}
\end{figure}
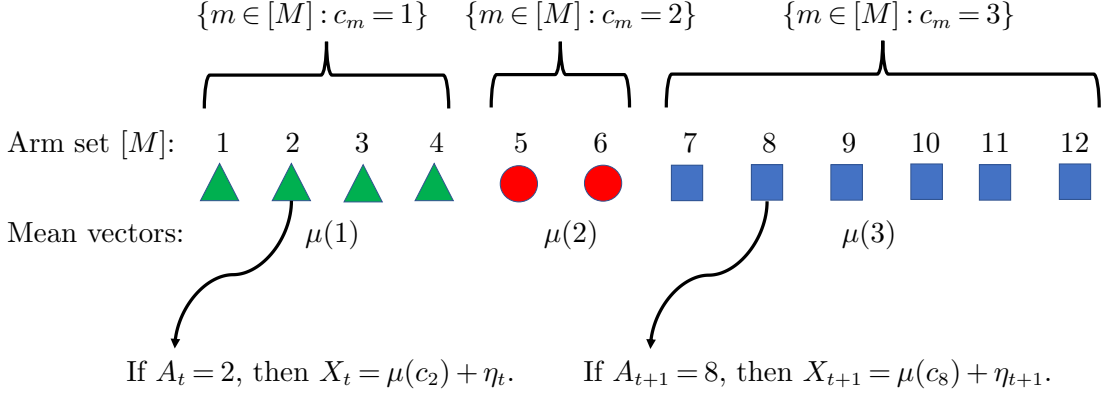 
At each time $t$, the agent selects an arm $A_t$ from the arm set $\mathcal A$, and then observes an noisy measurement on the mean vector of $A_t$, i.e., 
$$
X_t = \mu(c_{A_t})+\eta_t
$$
where $\{\eta_t\}_{t=1}^{\infty} \subset \mathbb R ^ d$ is a sequence of  independent (noise) random variables, each following the standard $d$-dimensional Gaussian distribution $\mathcal N (\mathbf 0, \mathbf I_d)$. See Figure~\ref{fig:model} for a schematic of the model.
\begin{remark}

In this work, we assume that the noise at each time $t$ is a standard $d$-dimensional Gaussian random vector. For general (e.g., non-diagonal)  noise covariance matrices, one can use the Cholesky decomposition to transform the raw observations in an affine manner into ones that have an identity covariance matrix, without loss of generality. 

\end{remark}

\paragraph{The Equivalences of Partitions and Instances.} The representation of a partition $c$ or an instance $(c,\mathcal{U})$ is not unique and we can accordingly define two equivalence relations. For a permutation $\sigma$ on $[K]$, let $\sigma(c) := [\sigma(c_1), \sigma(c_2),\ldots, \sigma(c_M)]$ and $\sigma(\mathcal{U}) := [\mu(\sigma(1)), \mu(\sigma(2)), \ldots, \mu(\sigma(K))]$. Similarly, we define $\sigma(c,\mathcal{U}):= (\sigma(c),\sigma(\mathcal{U}))$. 

For two partitions $c$ and $c'$, if there exists a permutation $\sigma$ on $[K]$ such that $c = \sigma(c')$, then we write $c \sim c'$. Due to the bijectivity of permutations, there also exists another permutation $\sigma^{-1}$ on $[K]$ such that $c' = \sigma^{-1}(c)$. Therefore, it is straightforward to verify this is indeed an equivalence relation.

For two instances $(c,\mathcal{U})$ and $(c',\mathcal{U}')$,\footnote{{Throughout this paper, we denote the columns of $\mathcal{U}' \in \mathbb{R}^{d \times K}$ as $\{\mu(k)\}_{k=1}^K$. The same convention applies to other forms of ${\mathcal{U}}$, such as $\bar{\mathcal{U}}$, ${\mathcal{U}}^*$, and ${\mathcal{U}}^\dagger$.}} if $\mu(c_m)=\mu'(c'_m)$ for all $ m \in [M]$, then the two instances are equivalent and we denote this as $(c,\mathcal{U}) \sim (c',\mathcal{U}')$. Note that $(c',\mathcal{U}') = \sigma(c,\mathcal{U})$ for some permutation $\sigma$ indicates $(c,\mathcal{U}) \sim (c',\mathcal{U}')$, but the reverse implication may not be true. That is to say, if $(c,\mathcal{U}) \sim (c',\mathcal{U}')$, there may not exist a permutation $\sigma$ such that $(c',\mathcal{U}') = \sigma(c,\mathcal{U})$. See Example~\ref{example_instance}.

\begin{example}
\label{example_instance}
Let $K = 2$, $M = 3$ and $d = 1$. Consider the three instances $(c,\mathcal{U})$, $(c',\mathcal{U}')$ and $(c'',\mathcal{U}'')$ where 
$c = [1, 1, 2]$ , $c' = [2, 2, 1]$,  $ c''=[1,2,2]$ and $ \mathcal{U} = \mathcal{U}' =  \mathcal{U}'' = [1 , 1]$. Although $(c,\mathcal{U}) \sim (c',\mathcal{U}') \sim (c'',\mathcal{U}'')$, it holds that $c \sim c' \not \sim c''$.
\end{example}

\paragraph{Online Clustering with Bandit Feedback.} We consider a pure exploration task, aiming to find the unknown cluster structure $c$ by pulling arms adaptively. In the fixed-confidence setting where a confidence level $\delta\in (0,1)$ is given, the agent is required to find a \emph{correct} partition $c^{\mathrm{out}}$ of the arm set $[M]$ (i.e., $c^{\mathrm{out}}\sim c$) with a probability of at least $1-\delta$ in the smallest number of time steps. 

More formally, the agent uses an \emph{online algorithm} $\pi$ to decide the arm $A_t$ to pull at each time step $t$, to choose a time $\tau_\delta$ to stop pulling arms, and to recommend $c^{\mathrm{out}}$ as the partition to output eventually. Let $\mathcal F_t =\sigma(A_1,X_1,\ldots,A_t,X_t)$ denote the $\sigma$-field generated by the past measurements up to and including time $t$. Thus, the online algorithm $\pi$ consists of three components, namely, 
\begin{itemize}
    \item the \emph{sampling rule} selects $A_t$, which is $\mathcal F_{t-1}$-measurable;
    \item the \emph{stopping rule} determines a stopping time $\tau_\delta$ adapted to the filtration $(\mathcal F_t)_{t=1}^{\infty}$;
    \item the \emph{recommendation rule} outputs a partition $c^{\mathrm{out}}$, which is $\mathcal F_{\tau_\delta}$-measurable.
\end{itemize}

\begin{definition}
For a fixed confidence level $\delta\in (0,1)$, an online clustering algorithm $\pi$ is said to be {\em $\delta$-PAC (probably approximately correct)} if for all instances $(c,\mathcal{U})$,
$\Pr(\tau_\delta<\infty) = 1$ and the probability of error $\Pr(c^{\mathrm{out}} \not \sim c) \le \delta$. 
\end{definition}

Our overarching goal is to design a computationally efficient online $\delta$-PAC clustering algorithm while minimizing the expected sample complexity $\E[\tau_\delta]$. To rule out   pathological cases that might lead to infinite expected sample complexities for any algorithm, throughout this work, we only consider partitioning the instances  $(c,\mathcal U)$  that satisfy the following natural property:  the mean vectors for different clusters are {\em distinct} (i.e., the instances $(c,\mathcal U)$ subject to $\mathcal U \in \mathscr U := \{ \bar{\mathcal{U}} \in \mathbb R^{d\times K}: \bar \mu(k_1)\neq \bar \mu(k_2) \text{ for all } 1\le k_1 <k_2\le K\}$). However, this property might not hold for general instances of cluster bandits (including the alternative instances $\mathrm{Alt}(c)$ that we introduce in the next section). Note that the odd arm identification problem \citep{vaidhiyan2017learning,karthik2020learning,karthik2021detecting}) is not a special case of our online clustering problem with $K=2$ since we do not require the knowledge of the number of arms in each cluster.

\paragraph{Other Notations.} Let $\mathbb N$ denote the set of positive integers and $\mathbb R^+ $ denote the set of non-negative real numbers. For any positive integer $N$, $\mathcal P _{N} := \{x\in [0, 1]^N: \|x\|_1=1 \}$ denotes the probability simplex in $\mathbb R ^ N$ while $\mathcal P _{N} ^+ := \{x\in (0, 1)^N: \|x\|_1=1 \}$ denotes the open probability simplex in $\mathbb R ^ N$. 

For two partitions $c$ and $c'$, let $d_{\mathrm{H}}(c,c')$ denote the Hamming distance between $c$ and $c'$, i.e., $d_{\mathrm{H}}(c,c') := \sum_{m=1}^ M \mathbbm 1 \{ c_m \neq c'_m\} $. For any $a,b\in(0,1)$, the binary relative entropy, which is the KL-divergence between Bernoulli distributions with means $a$ and $b$, is denoted as $d_{\mathrm{KL}}(a, b):= a \log (a / b)+(1-a) \log ((1-a) /(1-b))$. 

When we write $i^* = \argmin_{i\in A} f(i)$ where $f(i)$ is a function of $i$ and $A$ is a finite set of integers or a finite set of vectors of integers, we are referring to the minimum index (in lexicographic order) in the set $\{i\in A: f(i) = \min_{j\in A} f(j)\}$ if it is not a singleton.

\section{Lower Bound}
\label{section_lower}
In this section, we leverage the ubiquitous change-of-measure argument for deriving impossibility results to derive an instance-dependent lower bound on the expected sample complexity $\E[\tau_\delta]$ for the online clustering problem. The lower bound  is  closely related to a combinatorial optimization problem. 
Although the  optimization in its original form appears to be intractable, we prove an interesting combinatorial property and reformulate the optimization as a finite convex minimax problem. Moreover, we further present some results on the computation and other useful properties (e.g., the continuity of the optimizer and the optimal value) of the optimization problem (and its sub-problem) embedded in the lower bound, which are fundamental and essential in our algorithm design (see Section~\ref{section_algo}).

The change-of-measure argument, of which the key idea dates back to \citet{chernoff1959sequential}, is ubiquitous in showing various lower bounds in (and beyond) bandit problems (e.g., regret minimization \citep{lai1985asymptotically, lattimore2017end}, pure exploration \citep{kaufmann2016complexity, vaidhiyan2017learning}). Using this technique, the probabilities of the same event under different probability measures are related via the KL-divergence between the two measures. 

For any fixed instance $(c,\mathcal{U})$, we define $\mathrm{Alt}(c) := \{(c',\mathcal{U}'): c''\neq c \text{ for any } (c'',\mathcal{U}'')\sim (c',\mathcal{U}')\}$, which is the set of alternative instances where $c$ is not a correct partition. In particular, we consider the probabilities of correctly identifying the cluster structures under $(c,\mathcal{U})$ and the instances in $\mathrm{Alt}(c)$, and apply the \emph{transportation} inequality \citep[Lemma~1]{kaufmann2016complexity}. The instance-dependent lower bound on $\E[\tau_\delta]$ is presented in the following theorem; see Appendix~\ref{appendix_proof_theorem_lowerbound} for the proof.

\begin{theorem}
\label{theorem_lowerbound}
For a fixed confidence level $\delta\in (0,1)$ and instance $(c,\mathcal{U})$, any $\delta$-PAC online clustering algorithm satisfies 
$$
    \E[\tau_\delta] \ge  { d_{\mathrm{KL}}(\delta, 1-\delta)}  { D^*(c,\mathcal{U})}
$$
where
\begin{align}
\label{equation_D}
    D^*(c,\mathcal{U}) :=\bigg(\frac 1 2 \sup_{\lambda \in \mathcal P _{M}} \inf_{(c',\mathcal{U}')\in \mathrm{Alt}(c)} \sum_{m=1}^{M} \lambda_m\|\mu(c_m)-\mu'(c'_m)\|^2\bigg)^{-1}.
\end{align}
Furthermore, 
\begin{align}
\label{equation_lowerbound}
\liminf_{\delta \rightarrow 0 } \frac {\E[\tau_\delta]} {\log(1/\delta)} \ge D^*(c,\mathcal{U}).
\end{align}
\end{theorem}

We refer to $D^*(c,\mathcal{U})$ as the {\em hardness parameter} of the online clustering task in the sequel. The asymptotic version of the instance-dependent lower bound given in Equation~\eqref{equation_lowerbound} in  Theorem~\ref{theorem_lowerbound} is tight in view of the expected sample complexity of the efficient algorithm we present in Section~\ref{section_algo}. %Thus, the limit inferior in the bound~\eqref{equation_lowerbound} can be directly replaced by a limit.  \textcolor{red}{this is not true. Note that the statement says ``for any $\delta$-PAC algorithm''. This (sequence of) algorithm could result in $ \frac {\E[\tau_\delta]} {\log(1/\delta)}$ not having a limit, because it's arbitrary! However, any optimal algorithm $\{ \tau_\delta^* \}_{\delta>0}$ will be such that  $\frac {\E[\tau_\delta^*]} {\log(1/\delta)}$ converges. }
Intuitively, any ${\lambda \in \mathcal P _{M}}$ in Equation~\eqref{equation_D} can be understood as the proportion of arm pulls, which inspires the design of the sampling rule of a $\delta$-PAC online clustering algorithm. The agent wishes to find the optimal proportion of arm pulls to distinguish the instance $c$ from the \emph{most confusing} alternative instances in $\mathrm{Alt}(c)$ (for which $c$ is not a correct partition). Therefore, with the knowledge of the instance $(c,\mathcal{U})$, the optimization problem embedded in \eqref{equation_D} naturally unveils the optimal sampling rule, which is the basic idea behind the   design of our sampling rule in Section~\ref{section_algo}.

For ease of description, we refer to the entire optimization problem and the inner infimization in \eqref{equation_D} as Problem~($\Box$) and Problem~($\bigtriangleup$), respectively, i.e., 
\begin{align*}
    &\text{Problem~($\Box$):}   && \sup_{\lambda \in \mathcal P _{M}} \inf_{(c',\mathcal{U}')\in \mathrm{Alt}(c)} \sum_{m=1}^{M} \lambda_m\|\mu(c_m)-\mu'(c'_m)\|^2\\*
    &\text{Problem~($\bigtriangleup$):} &&\inf_{(c',\mathcal{U}')\in \mathrm{Alt}(c)} \sum_{m=1}^{M} \lambda_m\|\mu(c_m)-\mu'(c'_m)\|^2.
\end{align*}
{These two optimization problems will also feature in the sampling rule and the stopping rule, respectively, of our proposed algorithm in Section~\ref{section_algo}.}
However, even given the full information of the instance $(c,\mathcal{U})$, solving Problem~($\Box$) is tricky although the inner objective function is a weighted sum of quadratic functions. In particular, the alternative instances in $\mathrm{Alt}(c)$ need to be identified by {\em both} $c$ and $\mathcal{U}'$, which  is rather involved  as  the definition of $\mathrm{Alt}(c)$ is combinatorial  and the number of instances in it is obviously infinite. For Problem~($\bigtriangleup$), one may consider first fixing $c'$, then  optimizing over different $\mathcal{U}'$. We remark that this idea is only theoretically but not practically feasible since for a fixed number of clusters $K$, the total number of possible partitions (which is called \emph{the Stirling number of the second kind} \citep{graham1989concrete}) grows asymptotically as $K^M/K!$ as the number of arms $M$ tends to infinity. Nevertheless, we show a natural and useful property of \text{Problem~($\bigtriangleup$).}  

\begin{lemma}
\label{theorem_opt1}
For any $\lambda \in \mathcal P _{M}$ and  $(c,\mathcal{U})$,
$$
\inf_{(c',\mathcal{U}')\in \mathrm{Alt}(c)} \sum_{m=1}^{M} \lambda_m\|\mu(c_m)-\mu'(c'_m)\|^2 = \inf_{\substack{(c',\mathcal{U}')\in \mathrm{Alt}(c):\\d_{\mathrm{H}}(c', c)=1 }} \sum_{m=1}^{M} \lambda_m\|\mu(c_m)-\mu'(c'_m)\|^2.
$$
\end{lemma}

Lemma~\ref{theorem_opt1} provides a useful combinatorial property of Problem~($\bigtriangleup$), which makes it possible to solve for the hardness parameter $D^*(c,\mathcal{U})$ efficiently. Instead of considering {\em all} the alternative instances in $\mathrm{Alt}(c)$, Lemma~\ref{theorem_opt1} shows it suffices to consider the instances whose partitions have a {\em Hamming distance of $1$} from the given partition $c$. The complete proof of Lemma~\ref{theorem_opt1} is deferred to Appendix~\ref{appendix_proof_theorem_opt1}, which consists of four steps (see Figure~\ref{figure_lemma1} therein for an illustration). In particular, we show for any instance $(c^\dagger,\mathcal{U}^\dagger) \in \mathrm{Alt}(c)$ such that $d_{\mathrm{H}}(c^\dagger, c) > 1$, there exists another instance $(c^*,\mathcal{U}^*) \in \mathrm{Alt}(c)$ such that $d_{\mathrm{H}}(c^*, c) = 1$ and the objective function under $(c^*,\mathcal{U}^*)$ is not larger than that under $(c^\dagger,\mathcal{U}^\dagger)$. This desirable combinatorial property depends strongly on the specific structure of a valid partition. In fact, in Example~\ref{example_counter} in Section~\ref{section_algo}, we will see an example where a similar combinatorial property no longer holds if the true mean vectors $\{\mu(c_m)\}_{m=1}^M$ in \eqref{equation_D}
are replaced by some empirical estimates which may be  obtained as the algorithm proceeds.
{Thanks to Lemma~\ref{theorem_opt1}, Problem~($\bigtriangleup$) turns out to be equivalent to a much simpler finite minimization problem, as shown in Proposition~\ref{theorem_opt2}.}

\begin{proposition}
\label{theorem_opt2}
For any $\lambda \in \mathcal P _{M}$ and  $(c,\mathcal{U})$, 
$$\inf_{(c',\mathcal{U}')\in \mathrm{Alt}(c)} \sum_{m=1}^{M} \lambda_m\|\mu(c_m)-\mu'(c'_m)\|^2 = \begin{cases} \min\limits_{\substack{k,k'\in[K]:\\ n(k)>1,k'\neq k}}  \frac {\bar  w(k) w(k')} { \bar w(k) +w(k')} \|\mu(k)-\mu(k')\|^2 &\text{ if } \lambda \in \mathcal P _{M}^+\\ 0 &\text{ otherwise }\end{cases}
$$
where 
$
w(k) :=   \sum_{m=1}^{M} \lambda_m \mathbbm 1\{c_m=k\}
$,
$
n(k) :=   \sum_{m=1}^{M} \mathbbm 1\{c_m=k\}
$
and
$
\bar  w(k) := \min_{m\in[M]:c_m=k} \lambda_m.
$  Moreover, if $\lambda \in \mathcal P _{M}^+$, the infimum in Problem~($\bigtriangleup$) can be replaced with a minimum.
\end{proposition}

%\red{$\mathcal U \in \mathscr U$ is necessary. Otherwise, there \emph{might} be some special cases that the LHS is not zero even if $\lambda \in \mathcal P _{M} \setminus \mathcal P _{M}^+$. I don't want to bother with it. Maybe we can just say in the whole paper, $\mathcal U \in \mathscr U$. However, $\mathcal U'$ might not be in $\mathscr U$.}

As a corollary of our former intuition where $\lambda$ represents the proportion of arm pulls, given the knowledge of the instance $(c,\mathcal{U})$ and the proportion of arm pulls, Problem~($\bigtriangleup$) tells us how similar the true instance $c$ and the most confusing alternative instances in $\mathrm{Alt}(c)$ are. In fact, Proposition~\ref{theorem_opt2} plays an essential role in the computation of the stopping rule of our method in Section~\ref{section_algo}, which succeeds in circumventing the need to solve NP-hard optimization problems. In addition, Proposition~\ref{prop_continuity1} below asserts the continuity of the optimal value of Problem~($\bigtriangleup$), which will help to assert that the stopping rule proposed in Section~\ref{section_algo} is asymptotically optimal. Refer to Appendices~\ref{appendix_proof_theorem_opt2} and \ref{appendix_proof_prop_continuity1} for the proofs of Proposition~\ref{theorem_opt2} and Proposition~\ref{prop_continuity1}, respectively.

\begin{proposition}
\label{prop_continuity1} For any fixed $c$, define $g:\mathcal P _{M}\times \mathbb R^{d\times K}\to \mathbb R^+$ as 
$$
g( \lambda,\mathcal{U}) : = \inf_{(c',\mathcal{U}')\in \mathrm{Alt}(c)} \sum_{m=1}^{M} \lambda_m\|\mu(c_m)-\mu'(c'_m)\|^2.
$$
Then $g$ is continuous on $\mathcal P _{M} \times \mathscr U$.
\end{proposition}

As a consequence of Proposition~\ref{theorem_opt2}, Proposition~\ref{theorem_opt3} below (proved in Appendix~\ref{appendix_proof_theorem_opt3}), transforms Problem~($\Box$)
into a finite convex minimax problem, which has been studied extensively in the optimization  literature (e.g., \cite{gigola1990regularization} \citet{herrmann1999genetic}, \citet{gaudioso2006incremental}). 

\begin{proposition}
\label{theorem_opt3}
For any $(c,\mathcal{U})$, 
\begin{align*}
 D^*(c,\mathcal{U})
 &=  2 \min_{w \in \mathcal P _{K}^+} \max_{\substack{k,k'\in[K]:\\ n(k)>1,k'\neq k}}  \left( \frac {n(k)} {w(k)} + \frac 1 { w(k')}\right) \|\mu(k)-\mu(k')\|^{-2}.
\end{align*}
\end{proposition}

A by-product of Proposition~\ref{theorem_opt3} is that the outer supremum in Problem~($\Box$) can be replaced with a maximum. Intuitively, the maximizer of Problem~($\Box$) in $\mathcal P _{M}$ represents the optimal proportion of arm pulls, which will be of considerable importance in our design of the sampling rule. In fact, there exists a bijection between the solution to the finite convex minimax problem above and Problem~($\Box$), as shown in Proposition~\ref{theorem_opt4} below. Although the finite convex minimax problem is not strictly convex in ${w \in \mathcal P _{K}^+}$, Proposition~\ref{theorem_opt4}   states that the solution to Problem~($\Box$) is unique. This, together with Proposition~\ref{prop_continuity2} concerning the continuity of the solution to Problem~($\Box$),   guarantees the computationally efficiency and the asymptotic optimality of our sampling rule in Section~\ref{section_algo}. The proofs of Proposition~\ref{theorem_opt4} and Proposition~\ref{prop_continuity2} are deferred to Appendices~\ref{appendix_proof_theorem_opt4} and \ref{appendix_proof_prop_continuity2}, respectively.

\begin{proposition}
\label{theorem_opt4}
For any $(c,\mathcal{U})$, the solution to
\begin{align}
    \argmax\limits_{\lambda \in \mathcal P _{M}} \inf\limits_{(c',\mathcal{U}')\in \mathrm{Alt}(c)} \sum\limits_{m=1}\limits^{M} \lambda_m\|\mu(c_m)-\mu'(c'_m)\|^2
        \label{theorem_opt4_equation}
\end{align}
is unique.

If $\lambda^*$ denotes the unique solution to \eqref{theorem_opt4_equation} and $w^*$ denotes the unique solution to
$$
\argmin_{w \in \mathcal P _{K}^+} \max_{\substack{k,k'\in[K]:\\ n(k)>1,k'\neq k}}  \left( \frac {n(k)} {w(k)} + \frac 1 { w(k')}\right) \|\mu(k)-\mu(k')\|^{-2},
$$
then $\lambda^*$ can be expressed in terms of $w^*$ as 
$$
\lambda^*_m =   \frac {w^*(c_m)} {n(c_m)}\, \text{  for all  } \,  m \in [M].
$$
\end{proposition}

\begin{proposition}
\label{prop_continuity2} For any fixed $c$, define  $\Lambda: \mathbb R^{d\times K}\to \mathcal P _{M}$ as 
$$
\Lambda(\mathcal{U} ) :=\argmax_{\lambda \in \mathcal P _{M}} \inf_{(c',\mathcal{U}')\in \mathrm{Alt}(c)} \sum_{m=1}^{M} \lambda_m\|\mu(c_m)-\mu'(c'_m)\|^2 .
$$
Then $\Lambda$ is continuous\footnote{In finite-dimensional spaces, pointwise convergence and convergence in $L^p$ norm are equivalent.} on $\mathscr U$. 
\end{proposition}

\section{Algorithm: Bandit Online Clustering}
\label{section_algo}
% For the online clustering task with bandit feedback, we propose a computationally efficient and asymptotically optimal algorithm, namely  \textsc{Bandit Online Clustering} (or \textsc{BOC}), whose pseudocode is presented in Algorithm~\ref{algo1} and explained in the following subsections. First, we introduce an important subroutine, which returns a guess of both the correct partition and the mean vectors given the past measurements as inputs. Then we explain the methodology of our sampling rule and stopping rule, respectively, focusing on their connections with the optimization problems discussed in Section~\ref{section_lower}. Finally, we show the effectiveness of \textsc{BOC} and prove that  its expected sample complexity $\E[\tau_\delta]$ is asymptotically optimal as the confidence level $\delta$ tends to zero.

For the online clustering task with bandit feedback, we propose a computationally efficient and asymptotically optimal algorithm, namely  \textsc{Bandit Online Clustering} (or \textsc{BOC}), whose pseudocode is presented in Algorithm~\ref{algo1}. {As elucidated in Section~\ref{section_setup}, our algorithm comprises three components: the sampling rule for selecting the arms to pull (Lines~3-9), the stopping rule for determining a stopping time (Lines 11-13), and the recommendation rule for producing a final result (the clustering $c^{\mathrm{out}}$). In the following subsections, we first introduce an important subroutine called \textsc{K-means--Maximin}. This subroutine is employed at each time step to return a guess of both the correct partition and mean vectors based on the past measurements.
We then explain the methodology of our sampling rule and stopping rule, respectively, focusing on their connections with the optimization problems discussed in Section~\ref{section_lower}. Finally, from a theoretical standpoint, we demonstrate the effectiveness of \textsc{BOC} and prove that its expected sample complexity $\E[\tau_\delta]$ is asymptotically optimal as the confidence level $\delta$ tends to zero.}

\begin{algorithm}[t]
\caption{{Bandit Online Clustering (BOC)}}
\label{algo1}
\hspace*{0.02in} {\bf Input:} Number of clusters $K$, confidence level $\delta$ and arm set $[M]$
\begin{algorithmic}[1]
\State Sample each arm once, set $t=M$ and initialize $\hat \mu_m (t)$ and $N_m (t) = 1 $ for all $ m \in [M]$.
\Repeat
\If{$\min_{m\in[M]} N_m (t) \le \max(\sqrt{t} - M/2, 0)$} \label{line:ifAlg1_1}
\State Sample $A_{t+1} = \argmin_{m\in[M]} N_m (t)$, and $(c^t,\mathcal{U}^t) \leftarrow (c^{t-1},\mathcal{U}^{t-1})$ 
\Statex \Comment{Forced exploration}  \label{line:ifAlg1_2}
\Else 
\State $(c^t,\mathcal{U}^t) \leftarrow \textsc{K-means--Maximin} (K,\{\hat \mu_m (t)\}_{m\in [M]}, \{N_m(t)\}_{m\in [M]})$\Comment{Algorithm~\ref{algo2}}
\State Solve \Comment{Proposition~\ref{theorem_opt4}}
\begin{equation}
\label{equation_lambda_star_t}
\lambda^*(t) = \argmax_{\lambda \in \mathcal P _{M}} \inf_{(c',\mathcal{U}')\in \mathrm{Alt}(c^t)} \sum_{m=1}^{M} \lambda_m\|\mu^t(c^t_m)-\mu'(c'_m)\|^2
\end{equation}
\State Sample $A_{t+1} = \argmax_{m\in[M]} (t \lambda^*_m(t) - N_m (t))$ \label{algo1_sampling_rule}
\EndIf
\State  $t\leftarrow t+1$, update $\hat \mu_m (t)$ and $N_m (t)$ for all $ m \in [M]$
\State Compute 
$$
Z_1(t) = \sum_{m=1}^{M} N_m(t)\|\hat \mu_m(t)-\mu^{t-1}(c^{t-1}_m)\|^2
$$
and solve \Comment{Proposition~\ref{theorem_opt2}}
$$
Z_2(t) = \min_{(c',\mathcal{U}')\in \mathrm{Alt}(c^{t-1})} \sum_{m=1}^{M} N_m (t)\|\mu^{t-1}(c^{t-1}_m)-\mu'(c'_m)\|^2 
$$
\State Set
$$
Z(t) = \frac 1 2 \left(\left(-\sqrt{Z_1(t)}+\sqrt{Z_2(t)}\right)_+\right)^2
$$
\Until{$Z(t) \ge \beta(\delta, t)$}
\end{algorithmic}
\hspace*{0.02in} {\bf Output:} $\tau_\delta = t$ and $c^{\mathrm{out}} = c^{t-1}$
\end{algorithm}

\subsection{Weighted \texorpdfstring{$K$}{K}-means with Maximin Initialization}
\label{subsection_weightes_kmeans}
Although we only aim at producing a correct partition in the final recommendation rule as noted in Section~\ref{section_setup}, learning the $K$ unknown mean vectors of the clusters is essential in the sampling rule as well as the stopping rule. Different from other pure exploration problems in bandits (e.g., best arm identification \citep{garivier2016optimal}, odd arm identification \citep{vaidhiyan2017learning}), in the online clustering problem, it is not straightforward to recommend an estimate of the pair $(c,\mathcal{U})$ given some past measurements on the arm set. However, using maximum likelihood estimation, we will see the equivalence between the recommendation subroutine and the classical offline \emph{weighted $K$-means clustering} problem, which has been shown to be NP-hard \citep{aloise2009np, mahajan2012planar}. 

Given the past arm pulls and observations up to time $t$ (i.e., $A_1,X_1,\ldots,A_t,X_t$), the log-likelihood function of the hypothesis that the instance can be identified by the pair $(c',\mathcal{U}')$ can be written as
\begin{align}
 \label{equation_loglikelihood}
     \ell (c',\mathcal{U}' \mid A_1,X_1,\ldots,A_t,X_t) := -\frac 1 2 \sum_{s=1}^t  \| X_{s} - \mu'(c'_{A_{s}}) \|^2 - \frac {td} 2 \log(2\pi).
 \end{align}
 For any arm $m \in [M]$, let $N_m(t) := \sum_{s=1}^t \mathbbm 1 \{ A_{s} = m \}$ and $\hat \mu_m (t):= \sum_{s=1}^t X_{s}\mathbbm 1 \{ A_{s} =m \} / N_m(t)$ denote the number of pulls and the empirical estimate up to time $t$, respectively. By rearranging Equation~\eqref{equation_loglikelihood}, the maximum likelihood estimate of the unknown pair $(c,\mathcal{U})$ can be expressed as
\begin{align}
 \label{equation_mle}
 \argmin_{(c',\mathcal{U}')} \sum_{m=1}^{M} N_m (t)\|\hat \mu_m(t)-\mu'(c'_m)\|^2 
 \end{align}
which consists in   minimizing a weighted sum of squared Euclidean distances between the empirical estimate of each arm and its associated center. Therefore, any algorithm designed for the weighted $K$-means clustering problem is applicable to obtain an approximate (not exact) solution to \eqref{equation_mle}.

We remark that although the weighted variant of the original $K$-means algorithm \citep{macqueen1967some, lloyd1982least} is an efficient heuristic for the weighted $K$-means clustering problem, to the best of our knowledge, there are no theoretical guarantees for finding a global minimum of this problem in general. To establish the asymptotic optimality of our online clustering method \textsc{BOC}, in the initialization stage of $K$-means, we leverage the \emph{Maximin} method, which is a farthest point heuristic proposed by \citet{gonzalez1985clustering}. The complete pseudocode for \textsc{Weighted $K$-means with Maximin Initialization} (abbreviated as \textsc{K-means--Maximin}) is presented in Algorithm~\ref{algo2}. In 
the following, we derive some useful properties of \textsc{K-means--Maximin}; see Appendix~\ref{appendix_proof_theorem_maximum} for the proof of Proposition~\ref{theorem_maximum}.

\begin{algorithm}[t]
\caption{Weighted $K$-means with Maximin Initialization (\textsc{K-means--Maximin})} 
\label{algo2}
\hspace*{0.02in} {\bf Input:} Number of clusters $K$, empirical estimate $\hat \mu_m$ and weighting $N_m$ for all $ m \in [M]$
\begin{algorithmic}[1]
\State Choose the empirical estimate of an arbitrary arm as the first cluster center $\hat \mu(1)$
\For{$k = 2$ \textbf{to} $K$}\Comment{Maximin Initialization}
\State Choose the empirical estimate of the arm that has the greatest Euclidean distance to the nearest existing center as the $k$-th center $\hat \mu(k)$:
$$
\hat \mu(k) = \argmax_{m\in[M]} \min_{1\le k' \le k-1} \|\hat \mu_m -\hat \mu(k') \|
$$ 
\EndFor
\Repeat \Comment{Weighted $K$-means}
\State \label{algo2_step1}Assign each arm to its closest cluster center: $$\hat c _m = \argmin_{k\in[K]} \|\hat \mu_m -\hat \mu(k)\|$$ 
\State \label{algo2_step2}Update each cluster center as the weighted mean of the empirical estimates of the arms in it: $$\hat \mu (k) = \frac {\sum_{m\in[M]} N_m \hat \mu_m \mathbbm 1 \{ \hat c_m = k \}} {\sum_{m\in[M]} N_m  \mathbbm 1 \{ \hat c_m = k \}}$$ 
\Until{Clustering $\hat c$ no longer changes}
\State Set $\mu^{\mathrm{out}}(k) = \hat \mu(k)$ for all $k\in[K]$
\end{algorithmic} 
\hspace*{0.02in} {\bf Output:} $c^{\mathrm{out}} = \hat c$ and $\mathcal U^{\mathrm{out}} = [\mu^{\mathrm{out}}(1),  \mu^{\mathrm{out}}(2), \ldots,  \mu^{\mathrm{out}}(K)]  $
\end{algorithm}

\begin{proposition}
\label{theorem_maximum}
Given an instance $(c,\mathcal{U})$, if the empirical estimates of the arms $\{\hat \mu_m \}_{m\in [M]}$ satisfy 
$$
\max_{m\in[M]} \|\hat \mu_m-\mu(c_m)\| < \frac 1 4 \min_{{k,k'\in[K]: k\neq k'}} \|\mu(k)-\mu(k')\| ,
$$
then \textsc{K-means--Maximin} will output a correct partition $c^{\mathrm{out}}\sim c$. Furthermore, suppose that $c= \sigma(c^{\mathrm{out}})$ for some permutation $\sigma$ on $[K]$. Then 
$$\max_{k\in[K]}\|\mu^{\mathrm{out}}(k) - \mu(\sigma(k))\| \le  \max_{m\in[M]} \|\hat \mu_m-\mu(c_m)\|.$$
\end{proposition}

\begin{remark}
\label{remark_clustering}
The \textsc{K-means--Maximin} subroutine used in Algorithm~\ref{algo1} can be replaced by any offline clustering algorithm that meets the following conditions without affecting the subsequent results including Proposition~\ref{prop_sampling_rule}, Proposition~\ref{theorem_complexity1} and Theorem~\ref{theorem_complexity2}: (i) given an instance $(c,\mathcal{U})$, if the empirical estimates of all the arms are sufficiently accurate, i.e., $\max_{m\in[M]} \|\hat \mu_m-\mu(c_m)\|$ is smaller than a constant $\epsilon$ that depends on the problem instance, then the clustering algorithm will output a correct partition $c^{\mathrm{out}}$, i.e., one that satisfies that $c = \sigma(c^{\mathrm{out}})$ for some permutation $
\sigma$; (ii) moreover, for any cluster $k \in [K]$, $\|\mu^{\mathrm{out}}(k) - \mu(\sigma(k))\|$ is not larger than $\max_{m\in[M]} \|\hat \mu_m-\mu(c_m)\|$.
\end{remark}

\subsection{Sampling Rule}
\label{subsection_sampling_rule}
Once the estimates of the partition and the mean vectors are obtained by the \textsc{K-means--Maximin} subroutine based on the past measurements, our sampling rule utilizes the efficient method presented in Proposition~\ref{theorem_opt4} to find a plug-in approximation of the optimal oracle sampling rule. This  then informs the algorithm of  the selection of the next arm to pull. 

In particular, Algorithm~\ref{algo1} follows the so-called \emph{D-Tracking} rule, originally proposed by \citet{garivier2016optimal} for best arm identification, to track the optimal sampling rule. For the purpose of ensuring the plug-in approximation $\lambda^*(t)$ to converge to the true optimal sampling rule $\lambda^*$, the D-Tracking rule introduce a stage of \emph{forced exploration}. At each time $t$, if there exists an arm whose number of pulls is not larger than the preset threshold ($\max(\sqrt{t} - M/2, 0)$), then the agent chooses to sample that under-sampled arm. Otherwise, the agent chooses the arm according to the difference between the current proportions of arm pulls and the plug-in approximation $\lambda^*(t)$ (as described in Line~\ref{algo1_sampling_rule} of Algorithm~\ref{algo1}). Proposition~\ref{prop_sampling_rule}, proved in Appendix~\ref{appendix_proof_prop_sampling_rule}, shows the asymptotic optimality of our sampling rule, which is a joint consequence of Propositions~\ref{theorem_opt4}, \ref{prop_continuity2} and \ref{theorem_maximum}.

\begin{proposition}
\label{prop_sampling_rule}
Using the sampling rule as detailed in Lines 3-9 of Algorithm~\ref{algo1}, the proportions of arm pulls in Algorithm~\ref{algo1} converge to the optimal oracle sampling rule $\lambda^*$ almost surely, i.e., 
$$
\Pr\left(\lim_{t\rightarrow \infty} \frac {N_m(t)} t = \lambda^*_m \textup{ for all }  m\in [M] \right)=1. 
$$
\end{proposition}

\begin{remark}
We utilize the forced exploration stage in Algorithm~\ref{algo1} (Lines~\ref{line:ifAlg1_1} and~\ref{line:ifAlg1_2}) to obtain coarse estimates of the means of the arms. This appears to be necessary since we do not make any assumption on the instance of cluster bandits except that the centers for different clusters are distinct. However, the clusters being distinct does not preclude them being   arbitrarily close to one another. This hinders the adoption of  a ``more adaptive'' exploration stage. When the ratio between the minimal and the maximal pairwise distances among the centers is lower bounded by a known constant $0<r\le 1$ (see Assumption~\ref{assumption_distance}), an improved and more aggressive exploration method  is presented in Equation~\eqref{equation_newsamplingrule} of  Appendix~\ref{appendix_forced}. This  is based on a delicate quantitative analysis of $\lambda^*$ (Proposition~\ref{proposition_distance} therein). Experimental results using this more aggressive forced exploration procedure are also presented in Appendix~\ref{appendix_forced_exploration_numerical}. 
\end{remark}

\subsection{Stopping Rule}
\label{subsection_stopping_rule}
As the arm sampling proceeds, the algorithm needs to determine when to stop the sampling and recommend a partition with an error probability of at most $\delta$, namely the stopping rule. Most existing algorithms for pure exploration in the fixed-confidence setting (e.g., \citet{garivier2016optimal}, \citet{jedra2020optimal}, \citet{feng2021robust}, \citet{reda2021dealing}) consider the  \emph{Generalized Likelihood Ratio} (GLR) statistic and find  suitable task-specific threshold functions. This strategy dates back to \citet{chernoff1959sequential}. However, we show that the method based on the standard GLR is computationally intractable in the following.

Let $(c^{t*},\mathcal{U}^{t*})$ be the maximum likelihood estimate of the unknown pair $(c,\mathcal{U})$ given the past measurements up to time $t$. Then $(c^{t*},\mathcal{U}^{t*})$ is also the global minimizer to \eqref{equation_mle}, i.e., $(c^{t*},\mathcal{U}^{t*}) = \argmin_{(c',\mathcal{U}')} \sum_{m=1}^{M} N_m (t)\|\hat \mu_m(t)-\mu'(c'_m)\|^2 $. 
Using the definition of the log-likelihood function in~\eqref{equation_loglikelihood}, the logarithm of the GLR statistic (referred to as the $\log$-GLR) for testing $(c^{t*},\mathcal{U}^{t*})$ against its alternative instances can be written as 
\begin{align}
    % &\phantom{\;=\;} \log\text{-GLR} \notag \\
    %\phantom{\;=\;} 
    \log\text{-GLR}&={\ell (c^{t*},\mathcal{U}^{t*} \mid A_1,X_1,\ldots,A_t,X_t )} -\min_{(c',\mathcal{U}')\in \mathrm{Alt}(c^{t*})}  {\ell (c',\mathcal{U}' \mid A_1,X_1,\ldots,A_t,X_t)} \notag \\ 
    &=\frac 1 2 \left( -\sum_{m=1}^{M} N_m(t)\|\hat \mu_m(t)-\mu^{t*}(c^{t*}_m)\|^2 + \min_{(c',\mathcal{U}')\in \mathrm{Alt}(c^{t*})} \sum_{m=1}^{M} N_m (t)\|\hat \mu_m(t)-\mu'(c'_m)\|^2\right). \label{equation_glr}
\end{align}
There are two critical computational issues with the above expression. First, evaluating the $\log$-GLR as Equation~\eqref{equation_glr} requires the exact global minimizer to \eqref{equation_mle}, which is NP-hard to find in general. More importantly, even with the knowledge of $(c^{t*},\mathcal{U}^{t*})$, one cannot efficiently solve the optimization problem in the  second term of Equation~\eqref{equation_glr}    although it appears to be  similar to Problem~($\bigtriangleup$) discussed in Section~\ref{section_lower}. One may conjecture that an analogous combinatorial property to Lemma~\ref{theorem_opt1} holds for the second term of Equation~\eqref{equation_glr}, which might shrink the feasible set from $\mathrm{Alt}(c^{t*})$ to the alternative instances whose partitions have a Hamming distance of exactly  $1$ from $c^{t*}$. We disprove this conjecture by showing a counterexample in Example~\ref{example_counter}. As a consequence, the only feasible approach, at least for the moment, is to check all the possible partitions in $\mathrm{Alt}(c^{t*})$, which is certainly computationally intractable.

\begin{example}
\label{example_counter}
Let $K = 2$, $M = 4$ and $d = 2$. At time $t$, suppose that the empirical estimates of the $4$ arms are respectively $[0,0]^\top, [a,0]^\top,[0,b]^\top$ and $[a,b]^\top$, where $0<b/\sqrt 2 < a < b$. Then the partition that attains the minimum in~\eqref{equation_mle} is $[1,1 ,2,2]$ (or $[2,2,1,1]$) whereas the partition of the most confusing alternative instances (the minimizer to the second term of Equation~\eqref{equation_glr}) is $[1,2 ,1,2]$ (or $[2,1,2,1]$). Obviously, their Hamming distance is $2$ rather than $1$.
\end{example}

To construct a practicable stopping rule, instead of the GLR statistic, we consider the statistic
$$
Z(t) := \frac 1 2 \left(\left(-\sqrt{Z_1(t)}+\sqrt{Z_2(t)}\right)_+\right)^2 $$
with
$$
Z_1(t) := \sum_{m=1}^{M} N_m(t)\|\hat \mu_m(t)-\mu^{t-1}(c^{t-1}_m)\|^2
$$
and 
$$
Z_2(t) := \min_{(c',\mathcal{U}')\in \mathrm{Alt}(c^{t-1})} \sum_{m=1}^{M} N_m (t)\|\mu^{t-1}(c^{t-1}_m)-\mu'(c'_m)\|^2 .
$$
Note that $Z(t)$ involves $(c^{t-1},\mathcal{U}^{t-1})$, which is the estimate of the true pair $(c, \mathcal U)$ produced by the \textsc{K-means--Maximin} subroutine based on the past measurements. In particular, $(c^{t-1},\mathcal{U}^{t-1})$ is not necessarily the global minimizer to \eqref{equation_mle} and our stopping rule does not have any requirement on the quality of this estimate. For the term $Z_2(t)$, Proposition~\ref{theorem_opt2} can be utilized directly since the inherent optimization is equivalent to Problem~($\bigtriangleup$) after an appropriate normalization. 

Let $\zeta(\cdot)$ denote the Riemann zeta function, i.e., $\zeta (s)=\sum _{n=1}^{\infty }{{n^{-s}}}$. The stopping time of Algorithm~\ref{algo1} is defined as
$$
\tau_\delta := \inf\{t\in \mathbb N:Z(t)\ge \beta(\delta,t )\}
$$
where 
\begin{equation}
\label{equation_beta}
\beta(\delta,t ) = \sum_{m=1}^{M} 2 d \log(4 + \log(N_m (t))) + Md \cdot \Psi \left (\frac {\log(1/\delta)} {Md} \right)
\end{equation}
with
$$
\Psi (x) = \min_{1/2 \le h \le 1 } \left(2 - 2 \log(4h) + \frac {\log(\zeta(2h))} {h} - \frac {\log(1-h)} {2h}+\frac {x} {h} \right) 
$$
is a threshold function inspired by the concentration results for univariate Gaussian distributions \citep{kaufmann2021mixture}.

\begin{remark}
The  function $\Psi$ used in the threshold $\beta(\delta,t )$ possesses some useful properties that can be  verified in a straightforward manner or found in \citet{kaufmann2021mixture}: (i) $\Psi (x)  = x+\log (x)+ o(\log (x))$ as $x\to \infty$; (ii) $\Psi (x)
 \ge x$ for all $x>0$; (iii) let $\psi(h) := 2 - 2 \log(4h) + \frac {\log(\zeta(2h))} {h} - \frac {\log(1-h)} {2h}+\frac {x} {h} $ and then $-\psi$ is a unimodal function on $[1/2, 1]$.
\end{remark}

Overall,  our stopping rule is easy to implement and computationally efficient and furthermore, we will see that it is asymptotically optimal in the next subsection. The effectiveness of our stopping rule is shown in Proposition~\ref{theorem_stopping_rule}, which ensures that provided the algorithm stops within a finite time, the probability of recommending an incorrect partition is no more than $\delta$.

\begin{proposition}
\label{theorem_stopping_rule}
The stopping rule of \textsc{BOC} (Algorithm~\ref{algo1}) ensures that
$$
\Pr(\tau_\delta<\infty, c^{\mathrm{out}} \not \sim c) \le \delta.
$$
\end{proposition}

The proof of Proposition~\ref{theorem_stopping_rule} is deferred to Appendix~\ref{appendix_proof_theorem_stopping_rule}. To confirm that our method \textsc{BOC} is indeed a $\delta$-PAC online clustering algorithm, it remains to show it terminates within a finite time almost surely.

\begin{remark}
\label{remark_stopping_rule}
Our stopping rule is not only applicable to our sampling rule and the estimates of $(c, \mathcal U)$ produced by the \textsc{K-means--Maximin} subroutine. In fact, it also applies to any other sampling rule and the estimates by any method of estimation. In Section~\ref{section_exp}, we will experimentally compare different sampling rules with the same stopping rule.
\end{remark}

\begin{remark}
\label{remark_odd_arm}
\citet{vaidhiyan2017learning} proposed a modified GLR, where the likelihood function in the numerator of the GLR statistic is replaced by an averaged likelihood function with respect to an artificial prior probability distribution, for the odd arm identification task. We remark that this method is also not practical for the online clustering task since it requires to compute one corresponding modified GLR for each possible partition. However, the total number of possible partitions is enormous (namely the Stirling number of the second kind).
\end{remark}

\subsection{Sample Complexity Analysis}
\label{subsection_sample_complexity}
In this subsection, we analyze the correctness and sample complexity of our algorithm \textsc{BOC} (Algorithm~\ref{algo1}). Proposition~\ref{theorem_complexity1} verifies that \textsc{BOC} terminates within a finite time almost surely. Together with 
Proposition~\ref{theorem_stopping_rule}, it shows \textsc{BOC} is indeed a $\delta$-PAC online clustering algorithm. Specifically, for any instance $(c,\mathcal{U})$, \textsc{BOC} recommends a correct partition $c^{\mathrm{out}}$ based on the noisy measurements on the arm set with a probability of at least $1-\delta$. 

As shown in Proposition~\ref{theorem_complexity1} and Theorem~\ref{theorem_complexity2} respectively, the sample complexity of \textsc{BOC} asymptotically matches the instance-dependent lower bound presented in Section~\ref{section_lower}, both almost surely and in expectation, as the confidence level $\delta$ tends to zero. Therefore, \textsc{BOC} provably achieves asymptotic optimality in terms of the expected sample complexity and, at the same time, is also computationally efficient in terms of its sampling, stopping and recommendation rules. Thus, it achieves the best of both worlds.
Refer to Appendices~\ref{appendix_proof_theorem_complexity1} and \ref{appendix_proof_theorem_complexity2} for the proofs of Proposition~\ref{theorem_complexity1} and Theorem~\ref{theorem_complexity2}, respectively.

\begin{proposition}
\label{theorem_complexity1} For any instance $(c,\mathcal{U})$, Algorithm~\ref{algo1} ensures that 
$$
\Pr(\tau_\delta<\infty) = 1
$$
and
$$
\Pr\left (\limsup_{\delta \rightarrow 0 } \frac {\tau_\delta} {\log(1/\delta)} \le D^*(c,\mathcal{U}) \right) = 1.
$$
\end{proposition}

\begin{theorem}
\label{theorem_complexity2} For any instance $(c,\mathcal{U})$, Algorithm~\ref{algo1} ensures that 
$$
\limsup_{\delta \rightarrow 0 } \frac {\E[\tau_\delta]} {\log(1/\delta)} \le D^*(c,\mathcal{U}) .
$$
\end{theorem}

% \section{Fixed-budget Setting}
%  $$\Delta_{\min} :=\min_{{k,k'\in[K]: k\neq k'}} \|\mu(k)-\mu(k')\|$$

% Upper bound of error probability: 
% $$
% 2M\exp \left( - \left\lfloor \frac T M \right\rfloor \cdot \frac {\Delta_{\min}^2} {8d} \right)
% $$

% Lower bound of error probability:
% $$
% \frac 1 4 \exp \left( - \left\lfloor \frac T M \right\rfloor \cdot \frac {\Delta_{\min}^2} {2} \right)
% $$

\section{Numerical Experiments}
\label{section_exp}
%In this section, we study the empirical performance of our algorithm \textsc{BOC} and compare it with two baselines, namely \textsc{Uniform} and \textsc{Oracle}. To fairly evaluate the efficacy of our sampling rule,  \textsc{Uniform} and \textsc{Oracle} only differ from \textsc{BOC} in the sampling rules but share the same stopping and recommendation rules. As we discussed in Remark~\ref{remark_stopping_rule}, any sampling rule and recommendation rule can be cooperated with by our stopping rule, which is actually the only feasible stopping rule in hand. 

In this section, we study the empirical performance of our algorithm \textsc{BOC} and compare it with two baselines, namely \textsc{Uniform} and \textsc{Oracle}. Since our stopping rule is the only computationally tractable one available and the recommendation rule is embedded in either the sampling rule or the stopping rule, we focus on evaluating the efficacy of our sampling rule in terms of the time it takes for the algorithm(s) to stop. As we discussed in Remark~\ref{remark_stopping_rule}, any sampling rule and recommendation rule can be combined with our stopping rule. Therefore, for the sake of fairness in comparison, \textsc{Uniform} and \textsc{Oracle} only differ from \textsc{BOC} in the sampling rules and retain the other frameworks. In particular, \textsc{Uniform} samples the $M$ arms in a simple round-robin fashion while \textsc{Oracle} samples the arms based on the optimal oracle sampling rule $\lambda^*$ (i.e., the estimate $\lambda^*(t)$ in Line~\ref{algo1_sampling_rule} of Algorithm~\ref{algo1} is replaced by $\lambda^*$, which is calculated with the unknown true pair $(c, \mathcal U)$).
In each setting, the reported sample complexities of different methods are averaged over $256$ independent trials and the corresponding standard deviations are also shown as error bars or directly in the table. Finally, we mention that the partitions we learn in all our experiments are {always} correct.

\subsection{Synthetic Dataset: Verifying the Asymptotic Optimality of \textsc{BOC}} \label{subsection_synthetic}
To study the asymptotic behavior of the expected sample complexities of different methods, we construct three synthetic instances with varying difficulty levels, where $K = 4$, $M = 11$ and $d = 3$. The partitions and the first three cluster centers of all the three instances are the same, while their fourth cluster centers vary. In particular, the three instances can be expressed as follows:
\begin{equation}
    \begin{dcases}
    c = [1, 1, 2, 2, 2, 2, 3, 3, 3, 4, 4]  \\
    \mu(1) = [0, 0, 0]^\top  \\
    \mu(2) = [0, 10, 0]^\top  \\
    \mu(3) = [0, 0, 10]^\top  \\
    \mu(4) =   {\begin{dcases}
      [5, 0, 0]^\top   & \text{for the \emph{easy} instance,}  \\
      [1, 0, 0]^\top   & \text{for the \emph{moderate} instance,}  \\
      [0.5, 0, 0]^\top  & \text{for the \emph{challenging} instance.}
    \end{dcases}}
    \end{dcases} \label{eqn:instances}
\end{equation}
Moreover, motivated by \citet{garivier2016optimal}, we also consider using a heuristic threshold function $\tilde \beta(\delta, t) := \log((1+\log(t))^d / \delta)$, which is an approximation to the  original threshold function $\beta(\delta, t)$ in Equation~\eqref{equation_beta}. Although no theoretical guarantee is available when we use $\tilde \beta(\delta, t)$, it seems practical (and even conservative) in view  of the empirical error probabilities (which are always zero) for large $\delta$ (e.g., $\delta \ge 10^{-1}$).

\begin{figure}[hbpt]
\begin{minipage}{1\textwidth}
    \centering
	\subfigure[The easy instance with $\beta(\delta, t)$.]{
		\begin{minipage}[b]{0.4\textwidth}
			\includegraphics[width=1\textwidth]{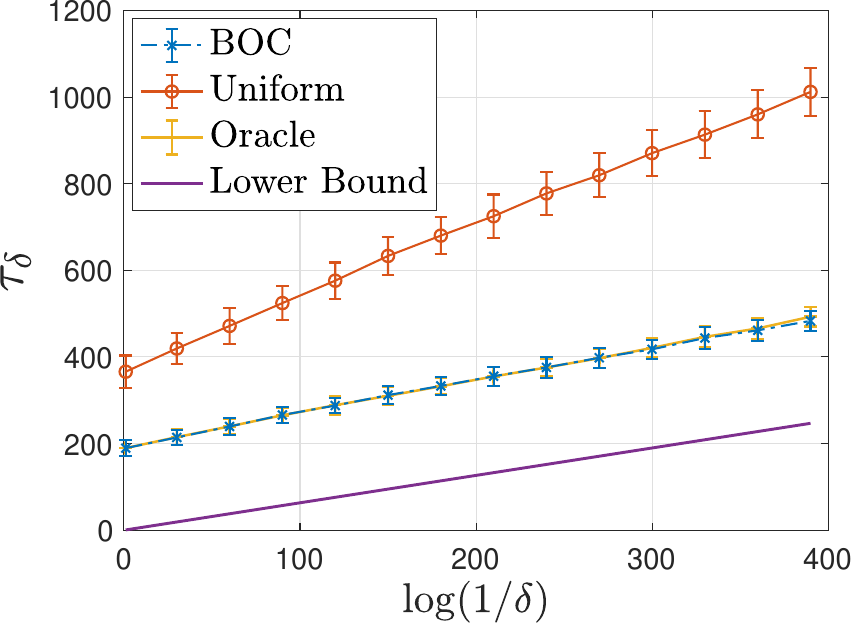} 
		\end{minipage}
		\label{figure_main1_1}
	}
	\subfigure[The easy instance with $\tilde \beta(\delta, t)$.]{
		\begin{minipage}[b]{0.4\textwidth}
			\includegraphics[width=1\textwidth]{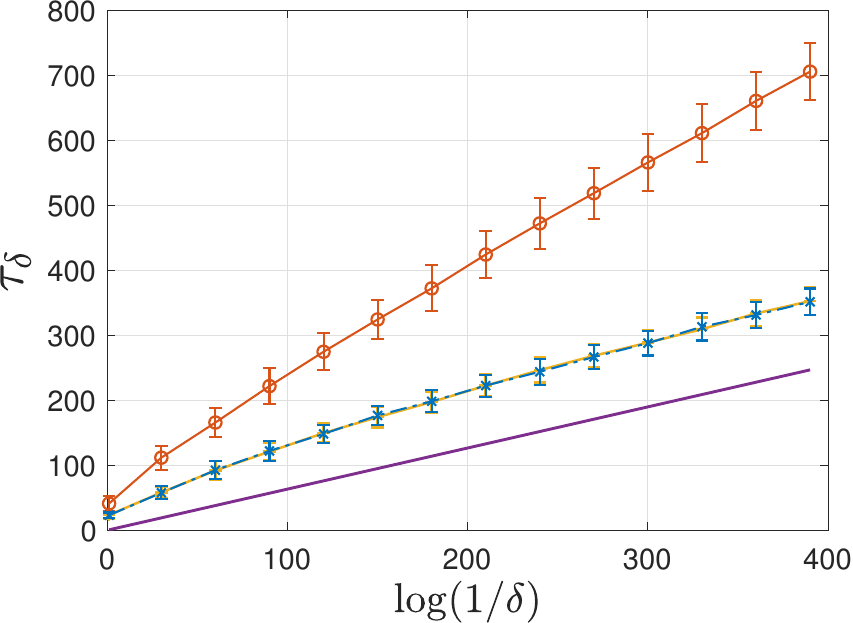} 
		\end{minipage}
		\label{figure_main1_2}
	}
    \subfigure[The moderate instance with $\beta(\delta, t)$.]{
    		\begin{minipage}[b]{0.4\textwidth}
   		 	\includegraphics[width=1\textwidth]{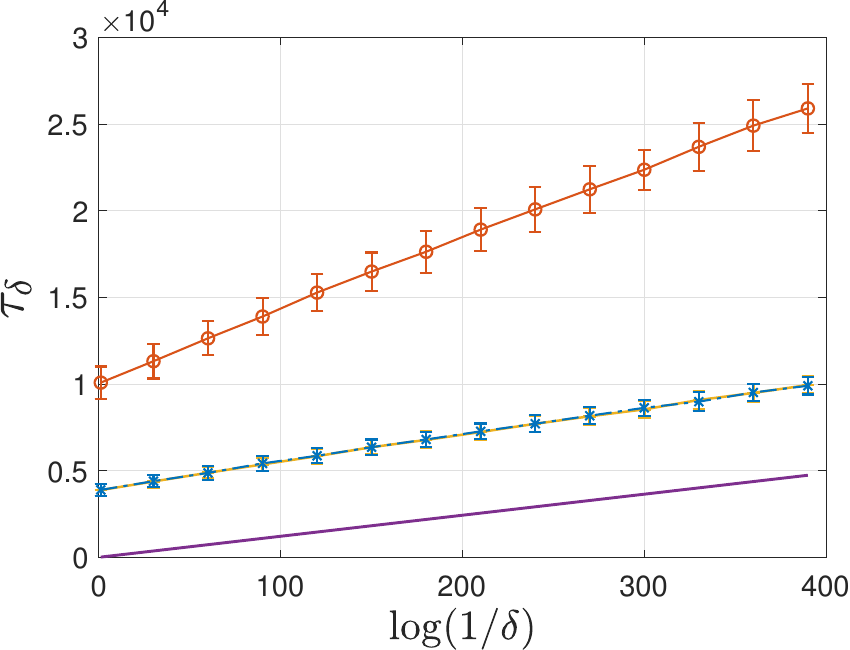}
    		\end{minipage}
		\label{figure_main2_1}
    	}
    \subfigure[The moderate instance with $\tilde \beta(\delta, t)$.]{
    		\begin{minipage}[b]{0.4\textwidth}
   		 	\includegraphics[width=1\textwidth]{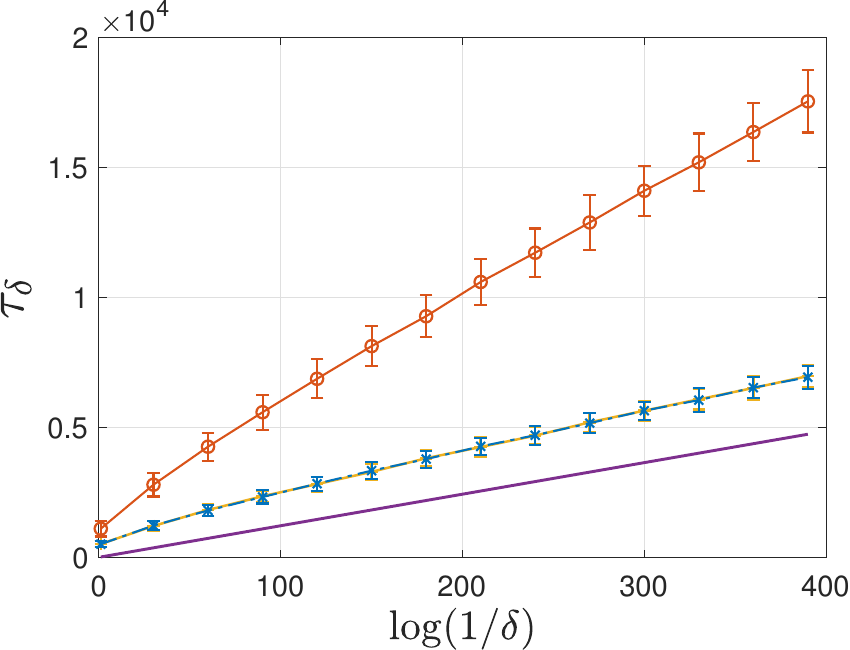}
    		\end{minipage}
		\label{figure_main2_2}
    	}
    \subfigure[The challenging instance with $\beta(\delta, t)$.]{
    		\begin{minipage}[b]{0.4\textwidth}
   		 	\includegraphics[width=1\textwidth]{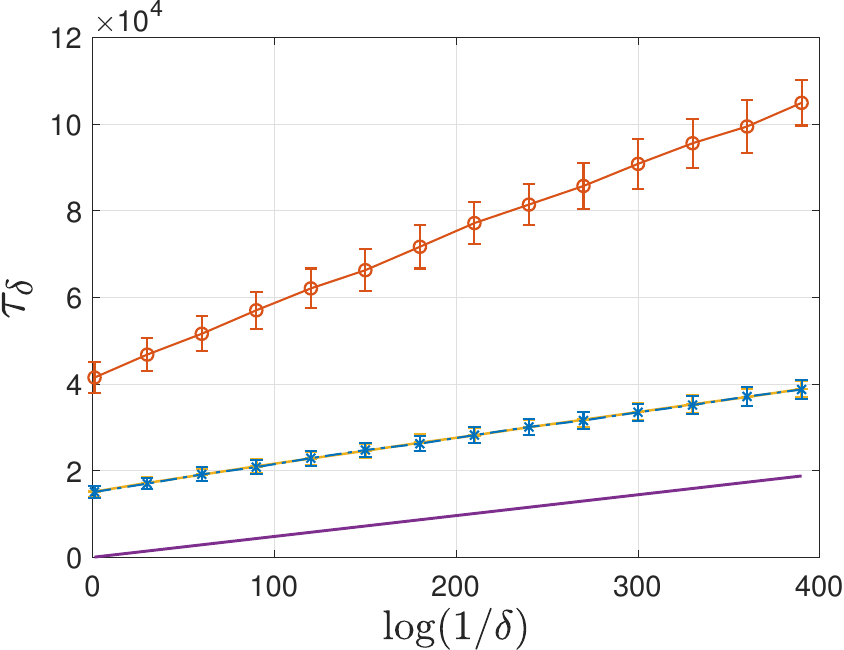}
    		\end{minipage}
		\label{figure_main3_1}
    	}
    \subfigure[The challenging instance with $\tilde\beta(\delta, t)$.]{
    		\begin{minipage}[b]{0.4\textwidth}
   		 	\includegraphics[width=1\textwidth]{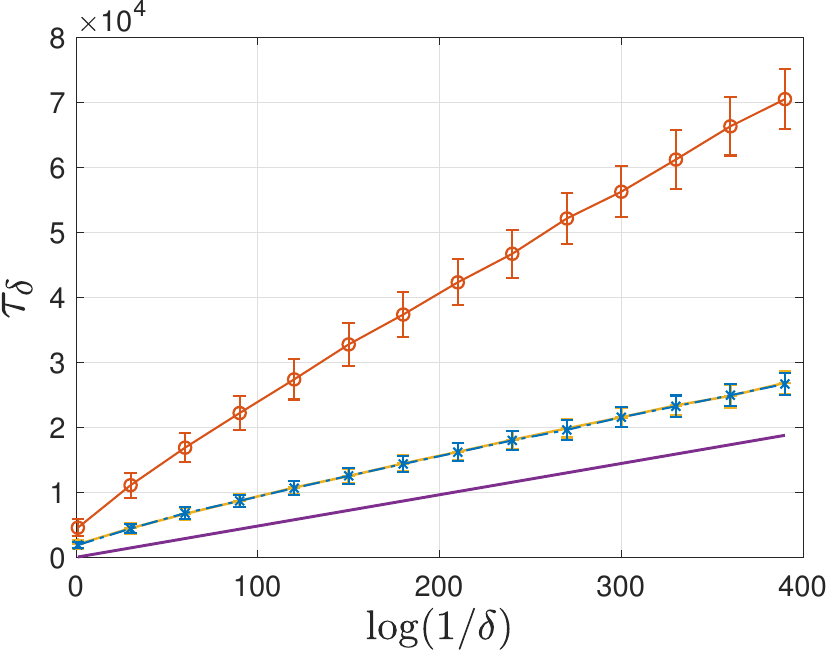}
    		\end{minipage}
		\label{figure_main3_2}
    	}
	\end{minipage}
	\caption{The empirical averaged sample complexities of the different methods with the two kinds of threshold functions for different confidence levels $\delta$ on the synthetic dataset.} \label{figure_main}
\end{figure}

The experimental results of the different methods with the two kinds of threshold functions for different confidence levels $\delta$ are presented in Figure~\ref{figure_main}. To better demonstrate the asymptotic behavior, we plot the 
empirical averaged sample complexities of the three methods as well as the instance-dependent lower bound of the expected sample complexity (see Theorem~\ref{theorem_lowerbound}) with respect to $\log(1/\delta)$ in the each sub-figure. From Figure~\ref{figure_main}, we have the following observations:
\begin{itemize}
    \item  Although our algorithm \textsc{BOC} does not require the optimal oracle sampling rule $\lambda^*$, and instead approximates it on the fly, the curves of \textsc{BOC} and \textsc{Oracle} in the each sub-figure are almost completely overlapping, which suggests the proportion of arm pulls for \textsc{BOC} converges to the distribution $\lambda^*$ very quickly.
    
    \item  
    The main observation is that as  $\delta$ decreases (or equivalently, as $\log(1/\delta)$ increases), the slope of the curve corresponding to the  \textsc{BOC} algorithm in the each sub-figure is almost equal to the slope of the lower bound, which is exactly equal to $ D^*(c,\mathcal{U})$. However, the  slope of the curve corresponding to  \textsc{Uniform} is consistently larger than that of the lower bound, exemplifying its suboptimality. This suggests that the expected sample complexity of our algorithm \textsc{BOC} matches the instance-dependent lower bound asymptotically, corroborating our theoretical results (see the lower bound and upper bound in Theorems~\ref{theorem_lowerbound} and \ref{theorem_complexity2}, respectively).
    
    \item There are unavoidable gaps between the lower bound and \textsc{BOC} (or, almost equivalently, \textsc{Oracle}), which is not an unexpected phenomenon. Although we have shown that the sampling rule resulting in $\lambda^*$ is asymptotically optimal, we have to utilize the stopping rule to evaluate the quality of the final recommendation (i.e., whether the recommended partition has an error probability of at most $\delta$). In addition, our bounds are only guaranteed to be tight {\em asymptotically} as $\delta\to 0$.
    
    \item Comparing the results for the same instance with the two  different threshold functions, it can be seen that the heuristic one $\tilde \beta(\delta, t)$ leads to lower sample complexities, especially for large $\delta$ (i.e., small $\log(1/\delta)$). Even though there are no theoretical guarantees when $\tilde \beta(\delta, t)$ is used in our stopping rule, this threshold appears to work well empirically.

\end{itemize}

\subsection{Real-world Datasets: Confirming the Non-asymptotic Superiority of \textsc{BOC}}

To complement the experiments on synthetic data and to verify that \textsc{BOC} also excels in the {\em non-asymptotic} regime, we conduct experiments on the Iris and Yeast datasets \citep{Dua:2019}, both of which are ubiquitous in  {\em offline} clustering and classification tasks. Here, we perform a novel task---{\em online} clustering  with {\em bandit} feedback.  In the Iris dataset, the number of clusters $K = 3$, the number of arms $M = 150$, and the dimension $d = 4$, while in the Yeast dataset, $K = 10$, $M = 1484$, and $d = 8$. Note that the total number of  partitions grows asymptotically as $K^M/K!$ (i.e.,  approximately $10^{70}$ or $10^{1477}$ in the Iris and Yeast datasets, respectively); hence,  it is impractical to exhaustively enumerate over all  partitions in these datasets. We emphasize that \textsc{BOC} succeeds in circumventing the need to solve any NP-hard optimization problem as a subroutine in the online clustering task. To adapt these datasets to be amenable to online clustering tasks, we choose each cluster center to be the mean of the original data points of the arms in it, and then rescale the centers so that the hardness parameter $D^*(c,\mathcal{U})$ is equal to $2$ for both datasets. Since the performances of \textsc{BOC} and \textsc{Oracle} are similar (as observed in Section~\ref{subsection_synthetic}) and the heuristic threshold function $\tilde \beta(\delta, t)$ generally achieves lower sample complexities (compared to when $\beta(t,\delta)$ is used), we only present the results of the  two methods (namely \textsc{BOC} and \textsc{Uniform}) with $\tilde \beta(\delta, t)$ on the two real datasets.

\begin{table}[t]
\centering
\setlength{\tabcolsep}{1em}
%\begin{tabular}{@{}lrrrrrrrr@{}}
\begin{tabular}{lrrrr}
\toprule
 & \multicolumn{2}{c}{\textbf{Iris Dataset}} & \multicolumn{2}{c}{\textbf{Yeast Dataset}} \\ \cmidrule(l){2-3} \cmidrule(l){4-5} 
 \multicolumn{1}{c}{$\delta$} & \multicolumn{1}{c}{\textsc{BOC}} & \multicolumn{1}{c}{\textsc{Uniform}} & \multicolumn{1}{c}{\textsc{BOC}} & \multicolumn{1}{c}{\textsc{Uniform}} \\ \midrule
$10^{-1}$ & 886.1 $\pm$ 55.9 & 1176.4 $\pm$ 69.2 & 14430.5 $\pm$ 371.3 & 19536.0 $\pm$ 530.4 \\
$10^{-2}$ & 922.5 $\pm$ 69.4 & 1208.7 $\pm$ 64.1 & 14531.2 $\pm$ 273.4 & 19697.4 $\pm$ 636.2 \\
$10^{-3}$ & 954.6 $\pm$ 80.0 & 1244.9 $\pm$ 72.7 & 14589.5 $\pm$ 175.2 & 19997.4 $\pm$ 718.7 \\
$10^{-4}$ & 993.8 $\pm$ 87.3 & 1286.2 $\pm$ 71.9 & 14631.5 $\pm$ 101.6 & 20220.3 $\pm$ 679.2 \\
$10^{-5}$ & 1026.9 $\pm$ 84.8 & 1306.0 $\pm$ 72.0 & 14639.6 $\pm$ 150.0 & 20467.7 $\pm$ 591.0 \\
$10^{-6}$ & 1059.0 $\pm$ 68.4 & 1335.7 $\pm$ 67.2 & 14686.3 $\pm$ 218.1 & 20564.5 $\pm$ 499.3 \\
$10^{-7}$ & 1075.8 $\pm$ 62.4 & 1368.5 $\pm$ 65.5 & 14723.8 $\pm$ 270.3 & 20686.5 $\pm$ 338.6 \\
$10^{-8}$ & 1090.0 $\pm$ 56.9 & 1389.4 $\pm$ 71.7 & 14797.3 $\pm$ 348.5 & 20733.1 $\pm$ 240.6 \\
$10^{-9}$ & 1104.8 $\pm$ 50.4 & 1415.6 $\pm$ 77.1 & 14844.6 $\pm$ 385.7 & 20762.0 $\pm$ 132.4 \\
$10^{-10}$ & 1120.2 $\pm$ 48.2 & 1447.0 $\pm$ 73.3 & 14977.1 $\pm$ 445.0 & 20766.8 $\pm$ 107.1 \\ \bottomrule
\end{tabular}
\caption{The  averaged empirical sample complexities of \textsc{BOC} and \textsc{Uniform} with the heuristic threshold function $\tilde \beta(\delta, t)$ for different confidence levels $\delta$ on the real-world datasets.}  \label{table_real}

\end{table}

The sample complexities for different confidence levels $\delta$ are presented in Table~\ref{table_real}.\footnote{The instance-dependent lower bound is not presented in Table~\ref{table_real} since it is not informative for these two real-world datasets in the non-asymptotic regime. For instance, even when the confidence level $\delta$ is equal to $10^{-10}$, the lower bound is only $D^*(c,\mathcal{U})\cdot \log(1/\delta)\approx 46$, which is much smaller than the total number of arms (i.e., $150$ or $1484$ in the Iris and Yeast datasets, respectively).} From the table, we see that \textsc{BOC} significantly outperforms  the non-adaptive baseline method \textsc{Uniform} for all $\delta$ in terms of sample complexities. This demonstrates that \textsc{BOC} is able to effectively learn the clusters in an online manner given bandit feedback. 

{In Appendix~\ref{appendix_experiment}, we consider a  dataset with a much higher dimensionality $d$, namely, the MNIST dataset \citep{lecun1998gradient}. From Table~\ref{table1} therein, we observe that the same conclusions apply.} 
\section{Conclusion}
\label{section_conclusion}

In this paper, we proposed a novel online clustering with bandit feedback framework  in which there is a set of arms that can be clustered into non-overlapping groups, and at each time, one arm is pulled, and a sample from the  distribution it is associated with is observed. We proposed and analyzed \textsc{Bandit Online Clustering} (or \textsc{BOC}) that, as discussed in Section~\ref{subsection_stopping_rule}, overcomes some critical computational limitations that a standard and natural GLR statistic suffers from due to the combinatorial search space of partitions.  In addition to its computational efficiency, we proved that \textsc{BOC} is asymptotically optimal in the sense that it attains an instance-dependent information-theoretic lower bound as the confidence level $\delta$ tends to zero.

%Contributions

There are some limitations of the current model and theoretical contributions that  serve as fertile avenues for future research. Firstly, in real-world applications such as recommendation systems and   online market segmentation, it is  often the case that the absolute correct clustering does not have to be found; an approximate clustering, with the advantage of further computational reductions, is usually sufficient. Developing computationally efficient and statistically optimal algorithms that allow for some distortion from the optimal clustering is thus of practical and theoretical importance. Secondly, our results are asymptotic in nature; they are only tight when the confidence level  $\delta$ tends to zero. As we have seen in Section~\ref{section_exp}, this results in a gap between the upper bound and the actual performance of \textsc{BOC} when $\delta$ is not vanishingly small. It would thus be instructive to develop {\em non-asymptotic} or {\em refined asymptotic} bounds, perhaps  by leveraging   the ``second-order'' results in \citet{malyutov2001second} and  \citet{litan}. 
Thirdly, our methodology can be generalized to the situation that the distributions of the observations are in the \emph{multivariate exponential families}, whereas it requires  efforts to preserve our computational efficiency.
Finally, it is worth developing bandit feedback models and algorithms for other generalizations of  clustering, such as hierarchical clustering, fuzzy or soft clustering, or community detection on graphs \citep{abbe2017community}.

% Acknowledgements should go at the end, before appendices and references
\newpage
\acks{We express our sincere gratitude to the reviewers and action editor for their meticulous reading and insightful comments. We are also thankful for the valuable discussions with Karthik Periyapattana Narayanaprasad and Yunlong Hou.\\
This research work is funded by the Singapore Ministry of Education Academic Research Fund (AcRF) Tier 2 under grant number A-8000423-00-00 and the Singapore Ministry of Education AcRF Tier 1 under grant number A-8000980-00-00.}


%\newpage

\appendix
% \section*{Appendix A.}
% \label{app:theorem}

% Note: in this sample, the section number is hard-coded in. Following
% proper LaTeX conventions, it should properly be coded as a reference:

%In this appendix we prove the following theorem from
%Section~\ref{sec:textree-generalization}:

\section{Auxiliary Lemmas}
\begin{lemma}[{The Maximum Theorem \citep{berge1963topological, sundaram1996first}}]
\label{lemma_maximum_theorem}
Let $f: S \times  \Theta \to \mathbb R $ be a continuous function , and $\mathcal{D}:\Theta \rightrightarrows S$ be a compact-valued continuous correspondence. Let $f^*: \Theta \to \mathbb R $ and $D^*: \Theta \rightrightarrows S $ be defined by
$$
f^{*}(\theta)=\max \{f(x, \theta) : x \in \mathcal{D}(\theta)\}
$$
and
$$
\mathcal{D}^{*}(\theta)=\argmax \{f(x, \theta) : x \in \mathcal{D}(\theta)\}=\{x \in \mathcal{D}(\theta):f(x, \theta)=f^{*}(\theta)\}.
$$
Then $f^{*}$ is a continuous function on $\Theta$, and $\mathcal{D}^{*}$ is a compact-valued, upper hemicontinuous\footnote{{\em Hemicontinuity} of a correspondence (resp.\ the adjective {\em hemicontinuous}) is also termed as {\em semicontinuity} (resp.\ {\em semicontinuous}) in some books, e.g.,~\citet{sundaram1996first}.} correspondence on $\Theta$.

\end{lemma}

\begin{lemma}[{\citet[Theorem~9.12]{sundaram1996first}}]
\label{lemma_correspondence_continuity}
A single-valued correspondence that is hemicontinuous (whether upper or lower hemicontinuous) is continuous when viewed as a function. Conversely, every continuous function, when viewed as a single-valued correspondence, is both upper and lower hemicontinuous.
\end{lemma}
\begin{lemma}
\label{lemma_sampling_rule}
Let $\lambda^*$ denote the oracle optimal sampling rule of the instance $(c,\mathcal{U})$. If there exists $\epsilon > 0$ and $t_0(\epsilon)$ such that 
$$
\sup_{t\ge t_0(\epsilon)} \max_{m\in[M]} |\lambda^*_m(t)- \lambda^*_m| \le \epsilon,
$$
then there exists $t_1(\epsilon) \ge t_0(\epsilon)$ such that 
$$
\sup_{t\ge t_1(\epsilon)} \max_{m\in[M]} \left |\frac {N_m(t)} t- \lambda^*_m \right | \le 3(M-1)\epsilon.
$$
Furthermore, a valid choice of $t_1(\epsilon)$ is $\frac {t_0(\epsilon)} {\epsilon^3}$.
\end{lemma}

%Remark (Junwen): I think the main reason why we cannot prove a non-asymptotic upper bound for our sample complexity is this lemma (we cannot deal with this 1 / \epsilon term).

\begin{proof}%[Proof of Lemma~\ref{lemma_sampling_rule}.] 
Our sampling rule satisfies the assumptions of   \citet[Lemma~17]{garivier2016optimal}, and the existence of $t_1(\epsilon)$ follows. The choice of $t_1(\epsilon)$ can be obtained by working through the proof of the same lemma.  %For the choice of $t_1(\epsilon)$, one needs to check the proof of the same lemma.
\end{proof}

\begin{lemma} 
\label{lemma_triangular1}
For any $x,y\in \mathbb R^+$, 
$$
\left(\left(-\sqrt{x}+\sqrt{y}\right)_+\right)^2 = \max_{\alpha \ge 0} \left( -\alpha x + \frac{\alpha} {\alpha+1}y \right) .
$$
\end{lemma}
\begin{proof}%[Proof of Lemma~\ref{lemma_triangular1}.]
Notice that
\begin{align*}
    -\alpha x + \frac{\alpha} {\alpha+1}y  = - (1+\alpha)x- \frac 1 {1+\alpha} y + (x + y) .
\end{align*}

If $y \ge x$, then
\begin{align*}
    \max_{\alpha \ge 0} \left( -\alpha x + \frac{\alpha} {\alpha+1}y \right) &= \left. \left(- (1+\alpha)x- \frac 1 {1+\alpha} y \right)\right |_{\alpha = \sqrt{\frac y x }-1} + (x + y) \\
    &= -2 \sqrt{xy}+ (x + y) \\
    &= \left(\left(-\sqrt{x}+\sqrt{y}\right)_+\right)^2.
\end{align*}

If $y < x$, then
\begin{align*}
    \max_{\alpha \ge 0} \left( -\alpha x + \frac{\alpha} {\alpha+1}y \right) &= \left. \left(- (1+\alpha)x- \frac 1 {1+\alpha} y \right)\right |_{\alpha = 0} + (x + y) \\
    &= 0 \\
    &= \left(\left(-\sqrt{x}+\sqrt{y}\right)_+\right)^2.
\end{align*}

\end{proof}

\begin{lemma} 
\label{lemma_triangular2}
For any $ x, y \in \mathbb R^d$ and $\alpha \ge 0$, 
$$
-\alpha \|x\|^2+\frac \alpha {\alpha+1} \|y\|^2 \le \|x-y \|^2.
$$
\end{lemma}
\begin{proof}%{Proof of Lemma~\ref{lemma_triangular2}.}
Notice that the above inequality is equivalent to
\begin{align*}
    &\phantom{\;\iff\;}  (1+\alpha) \|x\|^2-2x\cdot y+\frac 1 {\alpha+1} \|y\|^2  \ge 0 ,
\end{align*}
which is equivalent to 
\begin{align*}
    %\\ & \iff 
    \left\|\sqrt{1+\alpha} \cdot x - \frac 1 {\sqrt{\alpha+1}} \cdot y \right\|^2 \ge 0.
\end{align*}
Thus, the result  obviously holds.

\end{proof}

\begin{lemma}
\label{lemma_beta}
It holds that
$$\Pr\left (\exists\, t \in \mathbb N:  \frac 1 2 \sum_{m=1}^{M} N_m (t)\|\hat \mu_m(t) -\mu(c_m)\|^2 \ge \beta(\delta,t ) \right) \le \delta$$
where $\beta(\delta,t )$ is defined in Equation~\eqref{equation_beta}.
\end{lemma}

\begin{proof}%{Proof of Lemma~\ref{lemma_beta}.}
%For ease of notation, w
In this proof, we use $[x]_i$ to denote the $i^{\mathrm{th}}$ component of the vector $x \in \mathbb R^d$.

Recall that at each time $t$, the agent selects an arm $A_t$ and observes
$
X_t = \mu(c_{A_t})+\eta_t
$,
where $\eta_t$ follows the standard $d$-dimensional Gaussian distribution $\mathcal N (\mathbf 0, \mathbf I_d)$. Since each  individual component of $\eta_t$ independently follows the standard univariate Gaussian distribution $\mathcal N (0, 1)$, one arbitrary arm can be treated as $d$ independent sub-arms. Equivalently, the agent selects a group of $d$ sub-arms and observes 
$$
[X_t]_i = \left[\mu(c_{A_t})\right]_i+[\eta_t]_i 
$$
for all $ i\in[d] $. 

Note that there are $Md$ sub-arms in total. Then Lemma~\ref{lemma_beta} follows from the concentration inequality for the empirical means of sub-arms \citep[Theorem~9]{kaufmann2021mixture}.
\end{proof}

\begin{lemma}[{Adapted from \citet[Lemma 18]{garivier2016optimal}}]
\label{lemma_log}
For any two constants $a>0$ and $b\in \mathbb R$ such that $b+ \log\left( \frac 1 a \right) > 0 $,
$$
x= \frac 1 a \left( b + \log\left( \frac e a \right)+ \log \left( b+ \log\left( \frac 1 a \right) \right)\right)
$$
satisfies $ax\ge\log(x)+b$.
\end{lemma}

\section{Proofs of Section~\ref{section_lower}}
\label{appendix_lower}
\subsection{Proof of Theorem~\ref{theorem_lowerbound}}
\label{appendix_proof_theorem_lowerbound}
\begin{proof}%{Proof of Theorem~\ref{theorem_lowerbound}.} 
For fixed $\delta\in (0,1)$ and instance $(c,\mathcal{U})$, consider any $\delta$-PAC online clustering algorithm. 

We will see in Proposition~\ref{theorem_opt3} that $D^*(c,\mathcal{U})$ is finite so the situation that $\E[\tau_\delta]$ is infinite is trivial. Henceforth, we assume that $\E[\tau_\delta]$ is finite. 

 For any arm $m \in [M]$, let $N_m(t)$ denote the number of pulls of arm $m$ up to time $t$. Consider an arbitrary instance $(c',\mathcal{U}')$ in $\mathrm{Alt}(c)$. By applying the \emph{transportation} inequality \citep[Lemma 1]{kaufmann2016complexity} and the KL-divergence for the multivariate normal distribution, we have 
$$
\frac 1 2 \sum_{m=1}^{M}\E[N_m(\tau_\delta)] \|\mu(c_m)-\mu'(c'_m)\|^2 \ge  { d_{\mathrm{KL}}(\delta, 1-\delta)}.
$$

Since the above displayed inequality holds for all instances in $\mathrm{Alt}(c)$ and  $[\E[N_1(\tau_\delta)],\allowbreak \E[N_2(\tau_\delta)],\ldots,\E[N_M(\tau_\delta)]]^{\top}/ \E[\tau_\delta] $ forms a probability distribution in $\mathcal P _{M}$, we obtain
\begin{align*}
    { d_{\mathrm{KL}}(\delta, 1-\delta)} &\le \frac 1 2  \inf_{(c',\mathcal{U}')\in \mathrm{Alt}(c)} \sum_{m=1}^{M} \E[N_m(\tau_\delta)] \|\mu(c_m)-\mu'(c'_m)\|^2  \\
    &= \frac 1 2 \E[\tau_\delta] \inf_{(c',\mathcal{U}')\in \mathrm{Alt}(c)} \sum_{m=1}^{M} \frac {\E[N_m(\tau_\delta)]} {\E[\tau_\delta]} \|\mu(c_m)-\mu'(c'_m)\|^2   \\
    &\le \frac 1 2 \E[\tau_\delta] \sup_{\lambda \in \mathcal P _{M}} \inf_{(c',\mathcal{U}')\in \mathrm{Alt}(c)} \sum_{m=1}^{M} \lambda_m\|\mu(c_m)-\mu'(c'_m)\|^2 \\
    &= \E[\tau_\delta]   D^*(c,\mathcal{U})^{-1}.
\end{align*}

Since $\lim_{\delta \rightarrow 0 }  { d_{\mathrm{KL}}(\delta, 1-\delta)} / {\log(1/\delta)} =1 $, letting $\delta \to 0 $ yields 
$$
\liminf_{\delta \rightarrow 0 } \frac {\E[\tau_\delta]} {\log(1/\delta)} \ge D^*(c,\mathcal{U}) 
$$
as desired. 
\end{proof}

\subsection{Proof of Lemma~\ref{theorem_opt1}}
\label{appendix_proof_theorem_opt1}
\begin{proof}%{Proof of Lemma~\ref{theorem_opt1}.}
For any fixed $\lambda \in \mathcal P _{M}$ and $(c',\mathcal{U}') $, let 
$$\mathrm{Dist}(c',\mathcal{U}') := \sum_{m=1}^{M} \lambda_m\|\mu(c_m)-\mu'(c'_m)\|^2.$$

To prove Lemma~\ref{theorem_opt1}, we only need to show for any instance $(c^\dagger,\mathcal{U}^\dagger) \in \mathrm{Alt}(c)$ such that $d_{\mathrm{H}}(c^\dagger, c) > 1$, there exists another instance $(c^*,\mathcal{U}^*) \in \mathrm{Alt}(c)$ such that $d_{\mathrm{H}}(c^*, c) = 1$ and $\mathrm{Dist}(c^*,\mathcal{U}^*) \le \mathrm{Dist}(c^\dagger,\mathcal{U}^\dagger)$. The proof consists of four steps and Figure~\ref{figure_lemma1} serves as an illustration to help understand the various  constructions.

\begin{figure}[t]
\begin{minipage}{1\textwidth}
\centering
\begin{overpic}[width=1\textwidth]{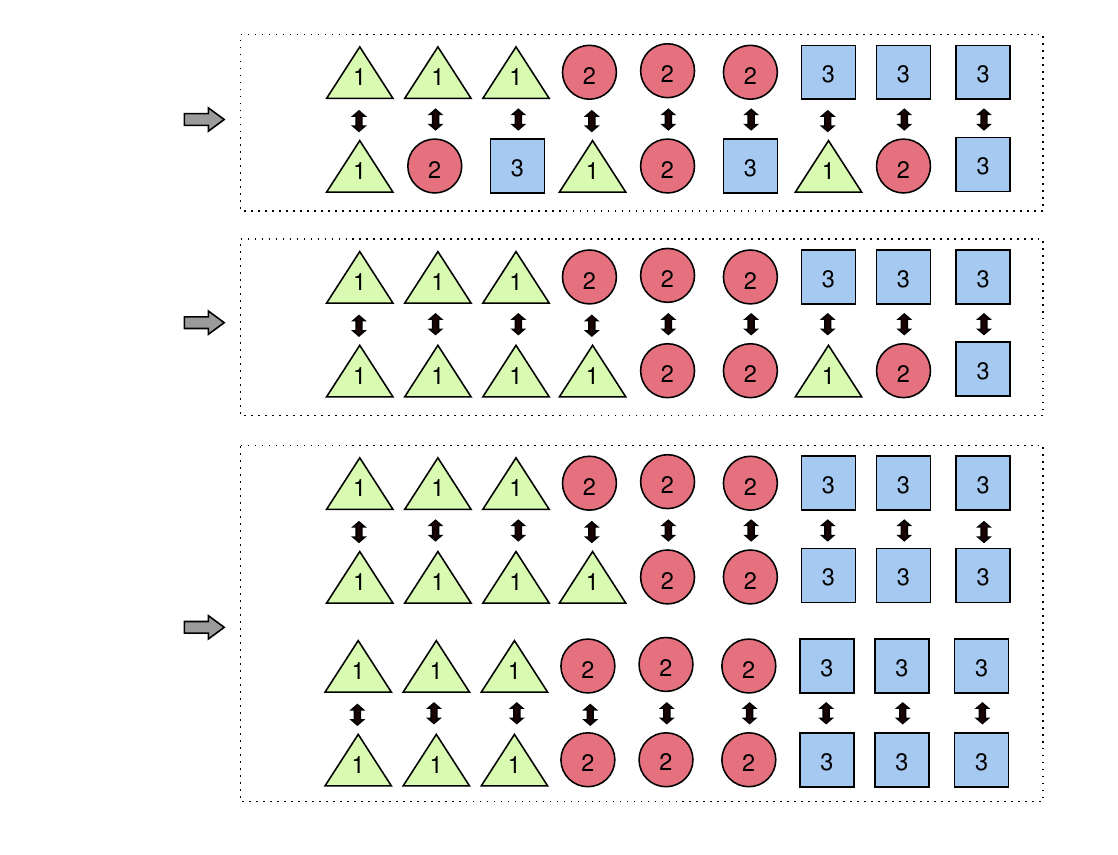}
\put(7.8, 65.28){\textbf{Step 1}}
\put(7.8, 47){\textbf{Step 2}}
\put(7.8, 19.68){\textbf{Step 3}}
\put(24, 69){{$c^{\phantom{(0)}}:$}}
\put(24, 60.6){{$c^{(0)}:$}}
\put(24, 50.7){{$c^{\phantom{(1)}}:$}}
\put(24, 42.3){{$c^{(1)}:$}}
\put(24, 32.2){{$c^{\phantom{(2)}}:$}}
\put(24, 23.8){{$c^{(2)}:$}}
\put(24, 15.9){{$c^{\phantom{(3)}}:$}}
\put(24, 7.4){{$c^{(3)}:$}}
\end{overpic}
\end{minipage}
\caption{An illustration of the proof of Lemma~\ref{theorem_opt1}. Figure~\ref{figure_lemma1} illustrates the  construction of the  sequence of instances when the number of clusters $K = 3$ and the number of arms (or items) $M = 9$. Each pair of items connected by a double arrow represents the cluster indices of one arm in the true partition $c$ and one of the newly constructed partitions $c^{(\bar{k})}$ for $\bar{k}=0,1,2,3$. After the application of the  permutation $\sigma$  as defined in~\eqref{eqn:permute} in Step 1, each of the arms can have any cluster index. However, due to the desirable property of the permutation as stated in~\eqref{theorem_opt1_firststep}, in Step 2, we are able to construct a new partition $c^{(1)}$ such that for any arm, its cluster index in $c^{(1)}$ is not larger than that of $c$. Next, we modify the new partition from right to left in Step 3 (see Equations~\eqref{eqn:right2left1} and~\eqref{eqn:right2left2}) and we finally return to a partition that is identical to $c$.}
\label{figure_lemma1}
\end{figure}

\paragraph{Step 1 (Permute the partition of the given instance).} To construct a new instance, we construct a permutation $\sigma$ on $[K]$ such that 
\begin{align} \label{eqn:permute}
    \left \{ \begin{array}{l}
    \sigma^{-1}(1) = \argmin\limits_{k\in[K]} \|\mu^\dagger(k)-\mu(1)||^2\\
    \sigma^{-1}(2) = \argmin\limits_{k\in[K] \setminus \{\sigma^{-1}(1) \}} \|\mu^\dagger(k)-\mu(2)||^2\\
    \sigma^{-1}(3) = \argmin\limits_{k\in[K] \setminus \{\sigma^{-1}(1),\sigma^{-1}(2)\}} \|\mu^\dagger(k)-\mu(3)||^2 \\
    \cdots\\
    \sigma^{-1}(K-1) = \argmin\limits_{k\in[K] \setminus \{\sigma^{-1}(1),\sigma^{-1}(2),\ldots \sigma^{-1}(K-2)\}} \|\mu^\dagger(k)-\mu(K-1)||^2.
    \end{array} \right. 
\end{align}

Let $(c^{(0)},\mathcal{U}^{(0)}) = \sigma(c^\dagger,\mathcal{U}^\dagger)$ and hence we have
\begin{align}
    \label{theorem_opt1_firststep}
    \left \{ \begin{array}{l}
    1 = \argmin\limits_{1\le k \le K} \|\mu^{(0)}(k)-\mu(1)||^2\\
    2 = \argmin\limits_{2\le k \le K} \|\mu^{(0)}(k)-\mu(2)||^2\\
    3 = \argmin\limits_{3\le k \le K} \|\mu^{(0)}(k)-\mu(3)||^2 \\
    \cdots\\
    K-1 = \argmin\limits_{K-1\le k \le K} \|\mu^{(0)}(k)-\mu(K-1)||^2.
    \end{array} \right. 
\end{align}

Obviously, $d_{\mathrm{H}}(c,c^{(0)}) \ge 1$, otherwise $(c^\dagger,\mathcal{U}^\dagger) \not \in \mathrm{Alt}(c)$.

\paragraph{Step 2 (Update the clustering).} Now we construct another instance $(c^{(1)},\mathcal{U}^{(0)})$, in which $c^{(1)}_m = \min(c_m, c^{(0)}_m)$ for all $ m \in [M]$. It holds that
\begin{align*}
    &\phantom{=} \mathrm{Dist}(c^{(0)},\mathcal{U}^{(0)}) \\ &= \sum_{m=1}^{M} \lambda_m\|\mu(c_m)-\mu^{(0)}(c^{(0)}_m)\|^2 \\
    &= \sum_{m=1}^{M} \lambda_m\|\mu(c_m)-\mu^{(0)}(c^{(0)}_m)\|^2 \mathbbm 1 \{c_m\le c^{(0)}_m\} + \sum_{m=1}^{M} \lambda_m\|\mu(c_m)-\mu^{(0)}(c^{(0)}_m)\|^2 \mathbbm 1 \{c_m>c^{(0)}_m\} \\
    &\ge \sum_{m=1}^{M} \lambda_m\|\mu(c_m)-\mu^{(0)}(c_m)\|^2 \mathbbm 1 \{c_m\le c^{(0)}_m\} + \sum_{m=1}^{M} \lambda_m\|\mu(c_m)-\mu^{(0)}(c^{(0)}_m)\|^2 \mathbbm 1 \{c_m>c^{(0)}_m\} \\
    &= \sum_{m=1}^{M} \lambda_m\|\mu(c_m)-\mu^{(0)}(c^{(1)}_m)\|^2 \\
    &= \mathrm{Dist}(c^{(1)},\mathcal{U}^{(0)})
\end{align*}
where the inequality is due to our construction of $(c^{(0)},\mathcal{U}^{(0)})$ in Equation~\eqref{theorem_opt1_firststep}.

\paragraph{Step 3 (Construct a sequence of instances).} For any $(c',\mathcal{U}') $ and $k,k' \in [K]$, let
$$w_{k,k'}(c') :=   \sum_{m=1}^{M} \lambda_m \mathbbm 1\{c_m=k, c'_m=k'\}.$$
Then 
$$
\mathrm{Dist}(c',\mathcal{U}') = \sum_{k=1}^K \sum_{k'=1}^K w_{k,k'}(c') \|\mu(k)-\mu'(k')\|^2.
$$

Moreover, the minimization problem $\min_{\mathcal{U}'}  \mathrm{Dist}(c',\mathcal{U}')$ has a unique solution   $\mathcal{U}^{\prime \star}$ which is defined by its components as 
\begin{equation}
\label{theorem_opt1_thirdstep}
\mu^{\prime \star}(k') = \frac {\sum_{k=1}^K w_{k,k'}(c') \mu(k)} {\sum_{k=1}^K w_{k,k'}(c')} 
\end{equation}
for all $k'\in[K]$.

Consider an instance $(c^{(1)},\mathcal{U}^{(1)})$, in which $\mathcal{U}^{(1)} = \argmin_{\mathcal{U}'}  \mathrm{Dist}(c^{(1)},\mathcal{U}')$. Clearly, it holds that $\mathrm{Dist}(c^{(1)},\mathcal{U}^{(0)}) \ge \mathrm{Dist}(c^{(1)},\mathcal{U}^{(1)})$. From the construction of $c^{(1)}$, we also know that $w_{k,K}(c^{(1)}) = 0$ for any $k\neq K$. Therefore, by \eqref{theorem_opt1_thirdstep}, $\mu^{(1)}(K) = \mu(K)$. 

Then consider another instance $(c^{(2)},\mathcal{U}^{(1)})$, where 
\begin{equation}
c^{(2)}_m = \begin{cases}
K & \text{if }c_m = K \\
c^{(1)}_m & \text{otherwise.}
\end{cases} \label{eqn:right2left1}
\end{equation}
Since $\mu^{(1)}(K) = \mu(K)$, $\mathrm{Dist}(c^{(1)},\mathcal{U}^{(1)}) \ge \mathrm{Dist}(c^{(2)},\mathcal{U}^{(1)})$. Next, we construct $(c^{(2)},\mathcal{U}^{(2)})$, in which  $\mathcal{U}^{(2)} = \argmin_{\mathcal{U}'}  \mathrm{Dist}(c^{(2)},\mathcal{U}')$. From the construction of $c^{(2)}$, we also know that $w_{k,K-1}(c^{(2)}) = 0$ for any $k\neq K-1$. Therefore, $\mu^{(2)}(K-1)= \mu(K-1)$ and $\mathrm{Dist}(c^{(2)},\mathcal{U}^{(2)}) \ge \mathrm{Dist}(c^{(3)},\mathcal{U}^{(2)})$ where 
\begin{equation}
c^{(3)}_m = \begin{cases}
K-1 & \text{if }c_m = K-1 \\
c^{(2)}_m & \text{otherwise.}
\end{cases} \label{eqn:right2left2}
\end{equation}

Following the same method,   we can then construct a sequence of instances $\{(c^{(3)},\mathcal{U}^{(3)}), \allowbreak (c^{(4)},\mathcal{U}^{(3)}), \ldots, (c^{(K)},\mathcal{U}^{(K)})\}$. Note that eventually $(c^{(K)},\mathcal{U}^{(K)}) = (c,\mathcal{U})$ since the cluster indices of $c^{(K)}$ and $c$ are the same and $\mathcal{U}^{(K)} = \argmin_{\mathcal{U}'}  \mathrm{Dist}(c^{(K)},\mathcal{U}')$.

\paragraph{Step 4 (Find the desired instance within the constructed sequence).} Consider the whole sequence $\{(c^{(0)},\mathcal{U}^{(0)}), (c^{(1)},\mathcal{U}^{(0)}), (c^{(1)},\mathcal{U}^{(1)}), (c^{(2)},\mathcal{U}^{(1)}), \ldots, (c^{(K)},\mathcal{U}^{(K)})\}$. Both the distance function $\mathrm{Dist}(\cdot,\cdot)$ and  the Hamming distance $d_{\mathrm{H}}(c,\cdot)$ are non-increasing. Furthermore, $d_{\mathrm{H}}(c,c^{(K)}) = 0$. 

If $d_{\mathrm{H}}(c,c^{(0)}) = 1$, we can simply choose $(c^*,\mathcal{U}^*)= (c^{(0)},\mathcal{U}^{(0)})$. 

If $d_{\mathrm{H}}(c,c^{(0)}) > 1$, from the sequence of instances, we can select the first instance whose partition is exactly the same as $c$, denoted as $(c^{(\bar k)}, \mathcal{U}^{(\bar k -1)})$ ($1\le \bar k \le K$), i.e., $c^{(\bar k)}= c$. We also know $(c^{(\bar k-1)}, \mathcal{U}^{(\bar k -1)})$ satisfies that  $d_{\mathrm{H}}(c,c^{(\bar k-1)}) \ge 1$. 

Our construction of the sequence of instances is ``consecutive'' in the sense that we can modify the cluster indices of $c^{(\bar k-1)}$ one by one until 
we get exactly the partition $c^{(\bar k)}$. Therefore, we can construct another sequence of instances $\{(c^{(\bar k-1)}, \mathcal{U}^{(\bar k -1)}), (c^{(\bar k'-1)}, \mathcal{U}^{(\bar k -1)}), \allowbreak (c^{(\bar k''-1)}, \mathcal{U}^{(\bar k -1)}), \ldots, (c^{(\bar k)}, \mathcal{U}^{(\bar k -1)})\}$, where the Hamming distance $d_{\mathrm{H}}(c,\cdot)$ is strictly decreasing by $1$ in each step in this sequence. Besides, the distance function $\mathrm{Dist}(\cdot,\cdot)$ is also non-increasing.  
As a result, we can always find an instance $(c^*,\mathcal{U}^*)$ in this sequence such that $d_{\mathrm{H}}(c^*, c) = 1$. 
\end{proof}

\subsection{Proof of Proposition~\ref{theorem_opt2}}
\label{appendix_proof_theorem_opt2}
\begin{proof}%{Proof of Proposition~\ref{theorem_opt2}.}
Due to Lemma~\ref{theorem_opt1}, we only need to consider the set of alternative instances whose partitions have a Hamming distance of $1$ from the given partition $c$, i.e., $\{(c',\mathcal{U}') \in \mathrm{Alt}(c): d_{\mathrm{H}}(c', c)=1\}$. 

If $\lambda \in \mathcal P _{M}^+$, suppose that $c$ and $c'$ only differ in the label of arm $m$, which is changed from $c_{m}$ to $k'$. Since $c'$ is a valid partition, we have $n(c_m)>1$. To minimize the objective function, for any $k\in[K]\setminus \{k'\}$, $\mu'(k)$ can be exactly set to be $\mu(k)$. 
Therefore, the objective function can be simplified to 
$$
\lambda_m \|\mu(c_m)-\mu'(k')\|^2 + w(k') \|\mu(k')-\mu'(k')\|^2,
$$
which shows the mean vector of cluster $k'$ should be chosen as $\mu'(k') = \frac{\lambda_m \mu(c_m) + w(k')\mu(k')} {\lambda_m+w(k')}$, a weighted sum of $\mu(c_m)$ and $\mu(k')$.

Altogether, the optimal value can be derived as follows:
\begin{align}
    &\phantom{\;=\;}\inf_{(c',\mathcal{U}')\in \mathrm{Alt}(c)} \sum_{m=1}^{M} \lambda_m\|\mu(c_m)-\mu'(c'_m)\|^2 \notag \\
    &=\inf_{\substack{(c',\mathcal{U}')\in \mathrm{Alt}(c):\\d_{\mathrm{H}}(c', c)=1 }} \sum_{m=1}^{M} \lambda_m\|\mu(c_m)-\mu'(c'_m)\|^2  \notag \\
    &= \inf_{\substack{m\in[M],k'\in [K]:\\ n(c_{m})>1,k'\neq c_m}} \frac {\lambda_m w(k')} { \lambda_m +w(k')} \|\mu(c_m)-\mu(k')\|^2 \label{theorem_opt1_1} \\
    &= \min_{\substack{m\in[M],k'\in [K]:\\ n(c_{m})>1,k'\neq c_m}} \frac {\lambda_m w(k')} { \lambda_m +w(k')} \|\mu(c_m)-\mu(k')\|^2 \label{theorem_opt1_11} \\
    &= \min_{\substack{k,k'\in[K]:\\ n(k)>1,k'\neq k}}  \frac {\bar  w(k) w(k')} { \bar w(k) +w(k')} \|\mu(k)-\mu(k')\|^2 \label{theorem_opt1_2}
\end{align}
where Equation~\eqref{theorem_opt1_1} follows from our choice of the mean vectors above and Equation~\eqref{theorem_opt1_2} follows from the fact that for any $y,z>0$, $f(x)= \frac {xy} {x+z}$ is increasing on $(0, \infty)$. As a consequence, the infimum in Problem~($\bigtriangleup$) can be replaced with a minimum since the infimum can be attained by some ${(c',\mathcal{U}')\in \mathrm{Alt}(c)}$.

If  $\lambda \in \mathcal P _{M} \setminus \mathcal P _{M}^+$, suppose that $\lambda_m = 0$ for some $m \in [M]$. Notice that the objective function is always non-negative.  

We consider two situations. First, if $n(c_m) > 1$, we will construct an instance $(c',\mathcal{U}')\in \mathrm{Alt}(c)$ such that the objective function is zero. In particular,
we can change the label of the arm $m$ from $c_m$ to any $k \neq c_m$, and keep all the mean vectors the same. Then the objective function for the new instance is exactly zero, which shows $$\inf_{(c',\mathcal{U}')\in \mathrm{Alt}(c)} \sum_{m=1}^{M} \lambda_m\|\mu(c_m)-\mu'(c'_m)\|^2 = 0.$$

Second, if $n(c_m) = 1$, we will construct an instance $(c',\mathcal{U}')\in \mathrm{Alt}(c)$ such that the objective function is arbitrarily close to zero. In particular, we can change the label of any other arm $m'$ (subject to $n(c_{m'}) > 1$) from $c_{m'}$ to $c_m$. Thus the objective function can be simplified to 
$$
\lambda_{m'} \|\mu(c_{m'})-\mu'(c_m)\|^2.
$$
For the mean vector of $c_m$, note that we cannot set $\mu'(c_m)$  to be exactly $\mu(c_{m'})$ otherwise  $(c',\mathcal{U}')\notin \mathrm{Alt}(c)$. However, we can let $\mu'(c_m)$ be arbitrarily close to $\mu(c_{m'})$, which yields 
$$\inf_{(c',\mathcal{U}')\in \mathrm{Alt}(c)} \sum_{m=1}^{M} \lambda_m\|\mu(c_m)-\mu'(c'_m)\|^2 = 0.$$

This completes the proof of Proposition~\ref{theorem_opt2}. 
\end{proof}

\begin{remark}
In fact, from the above proof, we   know more about the optimal solution if $\lambda \in \mathcal P _{M}^+$. Let
$$
(k^*,k'^{*}) := \argmin_{\substack{k,k'\in[K]:\\ n(k)>1,k'\neq k}}  \frac {\bar  w(k) w(k')} { \bar w(k) +w(k')} \|\mu(k)-\mu(k')\|^2.
$$
Then there exists $(c^*,\mathcal{U}^*)$, where 
$$
c^*_m = 
\begin{cases}
k'^{*} &\text{if } m=\argmin_{m\in[M]:c_m=k^*} \lambda_m \\
c_m &\text{otherwise}
\end{cases}
$$
and
$$
\mu^*(k) =  
\begin{cases}
\frac {\bar w(k^*)\mu(k^*)+w(k'^{*})\mu(k'^{*})} {\bar w(k^*)+w(k'^{*})} &\text{if } m=k'^{*}\\
\mu(k) &\text{otherwise}
\end{cases}
$$
such that 
$$(c^*,\mathcal{U}^*) \in \argmin_{(c',\mathcal{U}')\in \mathrm{Alt}(c)} \sum_{m=1}^{M} \lambda_m\|\mu(c_m)-\mu'(c'_m)\|^2.$$
 
\end{remark}

\subsection{Proof of Proposition~\ref{prop_continuity1}}
\label{appendix_proof_prop_continuity1}
\begin{proof}%{Proof of Proposition~\ref{prop_continuity1}.}
First, we define $f_1:\mathbb R^+ \times\mathbb R^+ \to \mathbb R^+ $ as
$$
f_1(x,y):= \begin{cases}
\frac {xy} {x+y} &\text{ if } x+y \neq 0 \\
0 &\text{ otherwise }
\end{cases}
$$
which is   continuous on $\mathbb R^+ \times\mathbb R^+ $.

Next, we will prove that for all $( \lambda,\mathcal{U}) \in \mathcal P _{M} \times \mathscr U $,  $g(\lambda,\mathcal{U})$ as defined in Proposition~\ref{prop_continuity1} can be written as
\begin{align}
\label{prop_continuity1_1}
g( \lambda,\mathcal{U}) = \min_{\substack{m\in[M],k\in [K]:\\ n(c_{m})>1,k\neq c_m}} f_1\left (\lambda_m, \sum_{\bar m=1}^{M} \lambda_{\bar m} \mathbbm 1\{c_{\bar m}=k\}\right )  \|\mu(c_m)-\mu(k)\|^2.
\end{align}

For any $\lambda \in \mathcal P _{M} ^+$, Equation~\eqref{prop_continuity1_1} follows directly from  Equation~\eqref{theorem_opt1_11} in the proof of Proposition~\ref{theorem_opt2}. 

For any $\lambda \in \mathcal P _{M} \setminus \mathcal P _{M}^+$, it suffices to find $m\in[M],k\in [K]$ subject to $n(c_{m})>1$ and $k\neq c_m$ such that  $f_1\left (\lambda_m, \sum_{\bar m=1}^{M} \lambda_{\bar m} \mathbbm 1\{c_{\bar m}=k\}\right )  = 0 $. Suppose that $\lambda_{\hat m}= 0$ for some ${\hat m} \in [M]$ and we will consider two situations. If $n(c_{\hat m}) > 1$, we can simply choose $m = \hat m$ and any $k \neq c_{\hat m}$. If $n(c_{\hat m}) = 1$, we can choose any $m$ subject to $n(c_m)>1$, and $k = c_{\hat m}$. Altogether, Equation~\eqref{prop_continuity1_1} holds for all $( \lambda,\mathcal{U}) \in \mathcal P _{M} \times \mathscr U $. 

Finally, for any $m\in[M],k\in [K]$ such that $n(c_{m})>1$ and $k\neq c_m$, we consider the function
$$
f_2(m,k;\lambda,\mathcal{U}):=f_1\left (\lambda_m, \sum_{\bar m=1}^{M} \lambda_{\bar m} \mathbbm 1\{c_{\bar m}=k\}\right )  \|\mu(c_m)-\mu(k)\|^2
$$
so that 
$$
g( \lambda,\mathcal{U}) = \min_{\substack{m\in[M],k\in [K]:\\ n(c_{m})>1,k\neq c_m}} f_2(m,k;\lambda,\mathcal{U}). 
$$
Since $(\lambda,\mathcal{U})\mapsto f_2(m,k;\lambda,\mathcal{U})$ is continuous on $\mathcal P _{M} \times \mathscr U$ for fixed $m$ and $k$ and the finite minimum operation preserves continuity, $(\lambda,\mathcal{U})\mapsto g(\lambda,\mathcal{U})$ is also continuous on $\mathcal P _{M} \times \mathscr U$.
\end{proof}

\subsection{Proof of Proposition~\ref{theorem_opt3}}
\label{appendix_proof_theorem_opt3}
\begin{proof}%{Proof of Proposition~\ref{theorem_opt3}.}
For any $(c,\mathcal{U})$, $ D^*(c,\mathcal{U})$ can be written as: 
\begin{align}
 D^*(c,\mathcal{U}) &=  \left\{ \frac 1 2 \sup_{\lambda \in \mathcal P _{M}} \inf_{(c',\mathcal{U}')\in \mathrm{Alt}(c)} \sum_{m=1}^{M} \lambda_m\|\mu(c_m)-\mu'(c'_m)\|^2 \right\}^{-1} \notag \\
 &=  \left\{ \frac 1 2 \max_{\lambda \in \mathcal P _{M}} \inf_{(c',\mathcal{U}')\in \mathrm{Alt}(c)} \sum_{m=1}^{M} \lambda_m\|\mu(c_m)-\mu'(c'_m)\|^2 \right\}^{-1} \label{theorem_opt3_1} \\
 &= \left\{ \frac 1 2  \max_{\lambda \in \mathcal P _{M}^+} \min_{\substack{k,k'\in[K]:\\ n(k)>1,k'\neq k}}  \frac {\bar  w(k) w(k')} { \bar w(k) +w(k')} \|\mu(k)-\mu(k')\|^2  \right\}^{-1} \label{theorem_opt3_2}  \\
 &= 2  \min_{\lambda \in \mathcal P _{M}^+} \max_{\substack{k,k'\in[K]:\\ n(k)>1,k'\neq k}}   \left( \frac 1 {\bar  w(k) } +  \frac 1 { w(k') } \right) \|\mu(k)-\mu(k')\|^{-2} \notag.
\end{align}
where Equation~\eqref{theorem_opt3_1} follows from the continuity of $g(\lambda, \mathcal{U})$ as shown in Proposition~\ref{prop_continuity1} and the compactness of $\mathcal P_M$, and Equation~\eqref{theorem_opt3_2} follows from Proposition~\ref{theorem_opt2} and the non-negativity of $D^*(c,\mathcal{U})$.

Notice that both $\{w(k)\}_{k=1}^K$ and $\{\bar w(k)\}_{k=1}^K$ depend on $\lambda$. For any $k$ and fixed $w(k)$, $\bar  w(k) = \min_{m\in[M]:c_m=k} \lambda_m$ is maximized \emph{if and only if} for all $m\in[M]$ such that $c_m = k$, $\lambda_m$ are equal. Therefore, we can solve the outer minimization on a ``smaller'' probability simplex $\mathcal P _{K}^+$. This yields 
\begin{align*}
 D^*(c,\mathcal{U})  &=  2 \min_{w \in \mathcal P _{K}^+} \max_{\substack{k,k'\in[K]:\\ n(k)>1,k'\neq k}}  \left( \frac {n(k)} {w(k)} + \frac 1 { w(k')}\right) \|\mu(k)-\mu(k')\|^{-2}
\end{align*}
as desired.
\end{proof}

\subsection{Proof of Proposition~\ref{theorem_opt4}}
\label{appendix_proof_theorem_opt4}
\begin{proof}%{Proof of Proposition~\ref{theorem_opt4}.}
According to the proof of Proposition~\ref{theorem_opt3}, there exists a bijective map (which is specified in the proof of Proposition~\ref{theorem_opt3} as well as the statement of Proposition~\ref{theorem_opt4}) between the solution(s) to 
$$ \argmax\limits_{\lambda \in \mathcal P _{M}} \inf\limits_{(c',\mathcal{U}')\in \mathrm{Alt}(c)} \sum\limits_{m=1}\limits^{M} \lambda_m\|\mu(c_m)-\mu'(c'_m)\|^2$$
and the solution(s) to
\begin{align}
\label{minimax_convex}
    \argmin_{w \in \mathcal P _{K}^+} \max_{\substack{k,k'\in[K]:\\ n(k)>1,k'\neq k}}  \left( \frac {n(k)} {w(k)} + \frac 1 { w(k')}\right) \|\mu(k)-\mu(k')\|^{-2}.
\end{align}

Therefore, it suffices to show the solution to \eqref{minimax_convex} is unique. We will show this by contradiction as follows. 

Suppose that $w^*$ and  $w^{**}$ are different solutions to \eqref{minimax_convex} and $w^*({\hat k}) \neq  w^{**}({\hat k})$ for some $\hat k\in [K]$.  

For any $k,k'\in[K]$ such that $n(k)>1$ and $k'\neq k$, consider the function 
$$
f_1(k,k';w, \mathcal U) := 2 \left( \frac {n(k)} {w(k)} + \frac 1 { w(k')}\right) \|\mu(k)-\mu(k')\|^{-2}.
$$
Since $ \|\mu(k)-\mu(k')\|^{-2}>0$ by the definition of $\mathscr U$, $f_1(k,k';w, \mathcal U)$ is convex in $w$. In particular, $f_1(k,k';w, \mathcal U)$ is strictly convex in the pair $(w(k),w(k'))$.
Besides, since the pointwise maximum operation preserves convexity, 
$$
f_2(w,\mathcal U ) := \max_{\substack{k,k'\in[K]:\\ n(k)>1,k'\neq k}} f_1(k,k';w, \mathcal U)
$$
is also convex in $w$. Therefore, $w^{***} := \frac 1 2 (w^*+w^{**}) $ is also a solution to \eqref{minimax_convex} and 
$$
 D^*(c,\mathcal{U})  = f_2(w^*,\mathcal U) = f_2(w^{**},\mathcal U) = f_2( w^{***},\mathcal U).
$$

We claim that for any $k,k'\in[K]$ such that $n(k)>1$ and $k'\neq k$, if ${\hat k} \in \{k,k'\}$, then
$$
f_1(k, k' ;w^{***}, \mathcal U) < D^*(c,\mathcal{U}).
$$
Otherwise, using the fact $w^*({\hat k}) \neq  w^{**}({\hat k})$ and the strict convexity of $f_1(k,k' ;w^{***}, \mathcal U) $ in the pair $(w^{***}(k),w^{***}(k'))$, 
\begin{align*}
    D^*(c,\mathcal{U}) &= f_1(k, k' ;w^{***}, \mathcal U) \\
    &< \frac 1 2 \left( f_1(k, k'  ;w^{*}, \mathcal U) +f_1(k, k'  ;w^{**}, \mathcal U) \right) \\
    &\le \frac 1 2  \left( f_2(w^*,\mathcal U) + f_2(w^{**},\mathcal U) \right) \\
    &= D^*(c,\mathcal{U})
\end{align*}
which leads to a contradiction. Therefore, our claim holds.

In view of this claim, we can identify a $w^{\dagger} \in \mathcal P _K^+$ such that $ f_2(w^{\dagger},\mathcal U) <  D^*(c,\mathcal{U})$, which results in a contradiction to the optimality of $D^*(c,\mathcal{U})$. In particular,  such a $w^{\dagger}$ can be defined through its components as 
\begin{align*}
    w^{\dagger}(k) = \begin{cases}
    w^{***}(k) - (K-1) \epsilon  &\text{if } k= \hat k \\
    w^{***}(k) + \epsilon &\text{otherwise}
    \end{cases}
\end{align*}
with a sufficiently small $\epsilon > 0$ such that 
\begin{align*}
    f_1(k,k';w^{\dagger}, \mathcal U) < D^*(c,\mathcal{U})
\end{align*}
for all $k,k'\in[K]$ such that $n(k)>1$, $k'\neq k$ and ${\hat k} \in \{k,k'\}$. The existence of such a sufficiently small $\epsilon > 0$ is guaranteed by the continuity of $f_1(k, k' ;w^{***}, \mathcal U)$ in the pair $(w^{***}(k),w^{***}(k'))$. 

In addition, for any $k,k'\in[K]$ such that $n(k)>1$, $k'\neq k$ and ${\hat k} \not \in \{k,k'\}$, 
\begin{align*}
    f_1(k,k';w^{\dagger}, \mathcal U) &= 2 \left( \frac {n(k)} {w^{\dagger}(k)} + \frac 1 { w^{\dagger}(k')}\right) \|\mu(k)-\mu(k')\|^{-2} \\ 
    &< 2 \left( \frac {n(k)} {w^{***}(k)} + \frac 1 { w^{***}(k')}\right) \|\mu(k)-\mu(k')\|^{-2} \\
    &=
    f_1(k,k';w^{***}, \mathcal U) \\
    &\le D^*(c,\mathcal{U}).
\end{align*}

In view of the fact that under both cases ($\hat{k} \in \{k,k'\}$ and $\hat{k} \notin \{k,k'\}$), we have $f_1 (k,k'; w^\dagger, U)< D^*(c,\mathcal{U})$, we conclude that 
$$ f_2(w^{\dagger},\mathcal U)= \max_{\substack{k,k'\in[K]:\\ n(k)>1,k'\neq k}} f_1(k,k';w^{\dagger}, \mathcal U)
<  D^*(c,\mathcal{U})$$
which contradicts the fact that $D^*(c,\mathcal{U})$ is the optimal value.

Altogether, the solution to \eqref{minimax_convex} is unique and hence the solution to 
$$ \argmax\limits_{\lambda \in \mathcal P _{M}} \inf\limits_{(c',\mathcal{U}')\in \mathrm{Alt}(c)} \sum\limits_{m=1}\limits^{M} \lambda_m\|\mu(c_m)-\mu'(c'_m)\|^2 $$
is also unique. 
\end{proof}

\subsection{Proof of Proposition~\ref{prop_continuity2}}
\label{appendix_proof_prop_continuity2}
\begin{proof}%{Proof of Proposition~\ref{prop_continuity2}}
Recalling Proposition~\ref{prop_continuity1}, 
$$
g( \lambda,\mathcal{U})  = \inf_{(c',\mathcal{U}')\in \mathrm{Alt}(c)} \sum_{m=1}^{M} \lambda_m\|\mu(c_m)-\mu'(c'_m)\|^2
$$
is continuous on $\mathcal P _{M} \times \mathscr U$. 

Since $\mathcal P _{M}$ does not depend on $\mathcal U$ and is compact, by Lemma~\ref{lemma_maximum_theorem}, 
$$
\Lambda(\mathcal{U} ) =\argmax_{\lambda \in \mathcal P _{M}} g( \lambda,\mathcal{U}) 
$$
is an upper hemicontinuous correspondence on $\mathscr U$. 

According to Proposition~\ref{theorem_opt4}, $\Lambda(\mathcal{U} )$ is single-valued. Consequently, by Lemma~\ref{lemma_correspondence_continuity}, $\Lambda$ is continuous on~$\mathscr U$. 
\end{proof}

\section{On the Forced Exploration Stage of Algorithm~\ref{algo1}}
\label{appendix_forced}
In this appendix, we study the situation that 
the ratio between
the minimal and the maximal pairwise distances among the cluster centers is bounded, and propose a novel exploration method, which is more adaptive than the original forced exploration used in Algorithm~\ref{algo1}. 
In particular, we assume that $(c,\mathcal U)$, the instance of cluster bandits to be partitioned, satisfies Assumption~\ref{assumption_distance}.

\begin{assumption}
\label{assumption_distance}
The ratio between the minimal and the maximal pairwise distances among the $K$ centers of the clusters $\{\mu(k)\}_{k\in [K]}$ is lower-bounded by a known constant $r \in (0, 1]$, i.e.,
$$
 \frac{\min_{{k,k'\in[K]: k\neq k'}} \|\mu(k)-\mu(k')\| }{\max_{{k,k'\in[K]: k\neq k'}} \|\mu(k)-\mu(k')\| } \ge r .
$$
\end{assumption}

We start from a quantitative result on the optimal oracle sampling rule $\lambda^*$, which is proved in Appendix~\ref{appendix_proof_proposition_distance}. 

\begin{proposition}
\label{proposition_distance}
Under Assumption~\ref{assumption_distance}, $\lambda^*$, which is the unique solution to 
\begin{align*}
    \argmax\limits_{\lambda \in \mathcal P _{M}} \inf\limits_{(c',\mathcal{U}')\in \mathrm{Alt}(c)} \sum\limits_{m=1}\limits^{M} \lambda_m\|\mu(c_m)-\mu'(c'_m)\|^2,
\end{align*}
satisfies $\lambda^*_m \ge \frac {r^2} {2M} $ for all ${m\in[M]}$.
\end{proposition}

Basically, Proposition~\ref{proposition_distance} shows that the proportions of arm pulls for the optimal sampling rule $\lambda^*$ are uniformly lower-bounded by a positive constant. Therefore, at each time step $t$, we no longer need to check whether there exists an arm whose number of pulls is not larger than a preset threshold. Once we obtain the plug-in approximation $\lambda^*(t)$ with the estimate produced by \textsc{K-means--Maximin}, we project it onto the constrained probability simplex $\mathcal P _{M} (\frac {r^2} {2M} ) := \{\lambda \in [\frac {r^2} {2M} , 1]^M: \|\lambda \|_1=1 \}$. Then instead of the original plug-in approximation $\lambda^*(t)$ which may result in under-sampling, we can track the optimal sampling rule based on this projection (which is denoted as $\tilde \lambda^*(t)$). In 
summary, the sampling rule (Line~\ref{algo1_sampling_rule} of Algorithm~\ref{algo1}) becomes
\begin{align}
\label{equation_newsamplingrule}
    A_{t+1} = \argmax_{m\in[M]} \; t \tilde \lambda^*_m(t) - N_m (t)
\end{align}
where 
\begin{align*}
    \tilde \lambda^*(t) = \operatorname{Proj}_{\mathcal P _{M} (\frac {r^2} {2M} ) }(\lambda^*(t))= \argmin_{\lambda \in {\mathcal P _{M} (\frac {r^2} {2M} ) }} \|\lambda -\lambda^*(t)\|.
\end{align*}
% (specifically, $\frac {r^2} {2M} t$)

We refer to the improved sampling rule in Equation~\eqref{equation_newsamplingrule} as the linear-exploration sampling rule henceforth. Compared to the original one proposed in Section~\ref{subsection_sampling_rule}, the linear-exploration sampling rule circumvents the forced exploration stage and always maintains a linear minimal exploration rate, which ensures that the plug-in approximation $\lambda^*(t)$  converges to the oracle optimal sampling rule $\lambda^*$ (almost surely). Moreover, since Proposition~\ref{proposition_distance} indicates that the optimal sampling rule always lies in the constrained probability simplex $\mathcal P _{M} (\frac {r^2} {2M} )$ under Assumption~\ref{assumption_distance}, the projection $\tilde \lambda^*(t)$ also converges to $\lambda^*$ (almost surely). Therefore,   asymptotic optimality is preserved. Besides, one can easily see that our subsequent results on sample complexity (Proposition~\ref{theorem_complexity1} and Theorem~\ref{theorem_complexity2}) also hold. 
However, since the forced exploration stage in the original sampling rule of \textsc{BOC} might take effect even if the plug-in approximation $\lambda^*(t)$ is close to the optimal sampling rule $\lambda^*$, the linear-exploration sampling rule may result in better empirical performance in the non-asymptotic regime, as demonstrated in the numerical results in Appendix~\ref{appendix_forced_exploration_numerical}.

Finally, we remark that the linear-exploration sampling rule requires extra knowledge of the pairwise distances among the cluster centers. This is unlikely to be available in most practical applications and hence we adopt an arguably less elegant forced exploration strategy in the main text.
%Although such information might be available in some real-world applications, we tend to handle more general situations and hence adopt a less elegant forced exploration stage in the main text.

\subsection{Proof of Proposition~\ref{proposition_distance}}
\label{appendix_proof_proposition_distance}
\begin{proof}
By Proposition~\ref{theorem_opt4}, it suffices to show that $w^*$, which is the unique solution to
$$
\argmin_{w \in \mathcal P _{K}^+} \max_{\substack{k,k'\in[K]:\\ n(k)>1,k'\neq k}}  \left( \frac {n(k)} {w(k)} + \frac 1 { w(k')}\right) \|\mu(k)-\mu(k')\|^{-2},
$$
satisfies $w^*(k) \ge \frac{r^2n(k)} {2M}$ for all $k\in[K]$. 

In the following, we will show this result by contradiction. 

Suppose that there exists $\hat k\in[K]$ such that $w^*(\hat k) < \frac{r^2n(\hat k)} {2M}$. For ease of notation, we denote 
\begin{align*}
    (\spadesuit) &: = \min_{w \in \mathcal P _{K}^+} \max_{\substack{k,k'\in[K]:\\ n(k)>1,k'\neq k}}  \left( \frac {n(k)} {w(k)} + \frac 1 { w(k')}\right) \|\mu(k)-\mu(k')\|^{-2}\\
    &\phantom{:}= \max_{\substack{k,k'\in[K]:\\ n(k)>1,k'\neq k}}  \left( \frac {n(k)} {w^*(k)} + \frac 1 { w^*(k')}\right) \|\mu(k)-\mu(k')\|^{-2}.
\end{align*}

If $n(\hat k) = 1$, then
\begin{align*}
(\spadesuit) &\ge \max_{\substack{k\in[K]:\\ n(k)>1,k\neq \hat k}}  \left( \frac {n(k)} {w^*(k)} + \frac 1 { w^*(\hat k)}\right) \|\mu(k)-\mu(\hat k)\|^{-2} \\
&> \frac 1 { w^*(\hat k)} \min_{{k,k'\in[K]: k\neq k'}} \|\mu(k)-\mu(k')\| ^{-2} \\
&> \frac  {2M}{r^2} \min_{{k,k'\in[K]: k\neq k'}} \|\mu(k)-\mu(k')\| ^{-2}.
\end{align*}

If $n(\hat k) > 1$, then 
\begin{align*}
(\spadesuit) &\ge \max_{k'\in[K]: k'\neq \hat k}  \left( \frac {n(\hat k)} {w^*(\hat k)} + \frac 1 { w^*(k')}\right) \|\mu(\hat k)-\mu(k')\|^{-2} \\
&> \frac {n(\hat k)} { w^*(\hat k)} \min_{{k,k'\in[K]: k\neq k'}} \|\mu(k)-\mu(k')\| ^{-2} \\
&> \frac {2M} {r^2} \min_{{k,k'\in[K]: k\neq k'}} \|\mu(k)-\mu(k')\| ^{-2}.
\end{align*}

Therefore, in both cases, we have 
\begin{align}
\label{equation_contradict}
(\spadesuit) > \frac  {2M}{r^2} \min_{{k,k'\in[K]: k\neq k'}} \|\mu(k)-\mu(k')\| ^{-2}.
\end{align}

On the other hand, we construct $\tilde w \in \mathbb R^{K}$ such that $\tilde w(k) = \frac {n(k)} {M}$ for all $k\in [K]$.  Since $\sum_{k\in [K]}  {n(k)} = M $, $\tilde w$ is a valid element in $\mathcal P _{K}^+$. Thus, under Assumption~\ref{assumption_distance}, it holds that
\begin{align*}
    (\spadesuit) &\le \max_{\substack{k,k'\in[K]:\\ n(k)>1,k'\neq k}}  \left( \frac {n(k)} {\tilde w(k)} + \frac 1 { \tilde w(k')}\right) \|\mu(k)-\mu(k')\|^{-2} \\
    &= \max_{\substack{k,k'\in[K]:\\ n(k)>1,k'\neq k}}  \left( M + \frac M { n(k')}\right) \|\mu(k)-\mu(k')\|^{-2} \\
    &\le 2M \max_{{k,k'\in[K]: k\neq k'}} \|\mu(k)-\mu(k')\| ^{-2} \\
    &\le \frac {2M} {r^2} \min_{{k,k'\in[K]: k\neq k'}} \|\mu(k)-\mu(k')\| ^{-2}
\end{align*}
which contradicts Inequality~\eqref{equation_contradict}. 

The proof of Proposition~\ref{proposition_distance} is thus completed.
\end{proof}

\begin{table}[t]
\centering
\setlength{\tabcolsep}{1em}
\begin{tabular}{lrrrr}
\toprule
 & \multicolumn{2}{c}{\textbf{\small{Rescaled Challenging Instance}}} & \multicolumn{2}{c}{\textbf{Iris Dataset}} \\ \cmidrule(l){2-3} \cmidrule(l){4-5} 
 \multicolumn{1}{c}{$\delta$} & \multicolumn{1}{c}{\textsc{Original}} & \multicolumn{1}{c}{\textsc{LE}} & \multicolumn{1}{c}{\textsc{Original}} & \multicolumn{1}{c}{\textsc{LE}} \\ \midrule
$10^{-1}$ & 102.9 $\pm$ 29.5 & 78.9 $\pm$ 20.3 & 886.1 $\pm$ 55.9 & 886.8 $\pm$ 47.5\\
$10^{-2}$ & 119.0 $\pm$ 28.9 & 86.7 $\pm$ 21.3 & 922.5 $\pm$ 69.4 & 920.7 $\pm$ 71.8\\
$10^{-3}$ & 132.3 $\pm$ 29.4 & 95.9 $\pm$ 22.3 & 954.6 $\pm$ 80.0 & 953.4 $\pm$ 77.8 \\
$10^{-4}$ & 142.1 $\pm$ 31.5 & 102.8 $\pm$ 23.6 & 993.8 $\pm$ 87.3 & 990.1 $\pm$ 88.8\\
$10^{-5}$ & 158.3 $\pm$ 32.7 & 114.7 $\pm$ 25.4 & 1026.9 $\pm$ 84.8 & 1023.6 $\pm$ 83.7 \\
$10^{-6}$ & 165.8 $\pm$ 32.3 & 120.1 $\pm$ 26.5 & 1059.0 $\pm$ 68.4 & 1055.3 $\pm$ 71.8 \\
$10^{-7}$ & 176.1 $\pm$ 31.2 & 128.9 $\pm$ 27.1 & 1075.8 $\pm$ 62.4 & 1079.1 $\pm$ 58.8 \\
$10^{-8}$ & 189.1 $\pm$ 35.3 & 135.4 $\pm$ 26.7 & 1090.0 $\pm$ 56.9 & 1092.2 $\pm$ 54.1 \\
$10^{-9}$ & 199.6 $\pm$ 32.8 & 143.3 $\pm$ 27.3 & 1104.8 $\pm$ 50.4 & 1109.6 $\pm$ 48.7 \\
$10^{-10}$ & 209.7 $\pm$ 33.8 & 152.2 $\pm$ 29.6 & 1120.2 $\pm$ 48.2 & 1120.0 $\pm$ 49.7 \\ \bottomrule
\end{tabular}
\caption{The averaged empirical sample complexities of \textsc{BOC} with the original forced-exploration sampling rule described in Algorithm~\ref{algo1} (and indicated by \textsc{Original} in the table) or the linear-exploration Sampling Rule (indicated by \textsc{LE}) with the heuristic threshold function $\tilde \beta(\delta, t)$ for different confidence levels $\delta$.}  \label{table_linear}
\end{table}

\subsection{Numerical Experiments on the Linear-Exploration Sampling Rule}
\label{appendix_forced_exploration_numerical}

In this section, we numerically compare the more aggressive linear-exploration sampling rule described in Equation~\eqref{equation_newsamplingrule} to the  the original sampling rule proposed in Section~\ref{subsection_sampling_rule}. We experiment this on the challenging synthetic instance in Equation~\eqref{eqn:instances} and the Iris dataset as these datasets behave differently. For the challenging synthetic instance, we rescale the cluster centers such that its hardness parameter $D^*(c,\mathcal{U})$ is equal to that of  the Iris dataset. For the linear-exploration sampling rule, the exact information of the ratio between the minimal and the maximal pairwise distances is given to the algorithm as an input.

The empirical stopping times for various (non-vanishing) failure probabilities $\delta$ are presented in Table~\ref{table_linear}. The behaviors on both datasets are different. The linear-exploration rule outperforms the original one for the rescaled challenging synthetic instance, while there are no obvious gaps between the two sampling rules for the Iris dataset. This is because for the rescaled challenging instance, there exists  some arms  that require a few arm pulls to learn their means sufficiently accurately. However, due to the forced exploration rule, they are pulled more often than necessary and this results in the inefficiency. On the other hand, this phenomenon is not observed for the Iris dataset and both Algorithm~\ref{algo1} and the linear-exploration rule work well in ensuring that the approximate proportions of arm pulls ($\lambda^*(t)$ or $\tilde \lambda^*(t)$) converge to the optimal one dictated by the lower bound. 

%which has a comparatively large number of arms ($M=150$), the forced exploration step in Algorithm~\ref{algo1} has a rather minimal effect on the entire algorithm due to the fact that the max in Line~\ref{line:ifAlg1_1} often assumes the value $0$. 
%both exploration methods suffer from the fact that 
%re consumed inefficiently by the forced exploration, which does not occur for the Iris dataset

% \section{Extensions to Other Noise Distributions}
% \label{appendix_extension}
% %In this appendix, 

% \paragraph{Sub-Gaussian Noise.} Two kinds of definitions of Sub-Gaussian Noise; stopping rule. 

% \paragraph{Multivariate Exponential Families.} Definitions (independent single-parameter exponential families); Computation(optimization lemma); K-means; stopping rule. 

\section{Proofs of Section~\ref{section_algo}}
\label{appendix_algo}
\subsection{Proof of Proposition~\ref{theorem_maximum}}
\label{appendix_proof_theorem_maximum}
\begin{proof}%{Proof of Proposition~\ref{theorem_maximum}.}
The proof consists of three steps, using the triangle inequality frequently.

\textbf{First}, we show that the $K$ arms chosen in Maximin Initialization are selected from $K$ disjoint clusters. 

Let $m_1, m_2,\ldots,m_{K}$ denote the arms chosen in Maximin Initialization in order. Assuming that we have already taken the empirical estimates of $\bar k$ arms as the cluster centers, consider $\mathcal A _{\bar k}$, the set of the arms that shares the same true cluster index with at least one of the existing centers. For any arm $m' \in \mathcal A _{\bar k}$, the Euclidean distance to the nearest existing center can be upper bounded as follows:
\begin{align*}
    \min_{1\le k \le \bar k} \| \hat \mu_{m'} - \hat \mu(k)\|  &=  \min_{1\le k \le \bar k} \| \hat \mu_{m'} - \hat \mu_{m_k}\| \\
    &\le \min_{1\le k \le \bar k} \left (\| \hat \mu_{m'} - \mu(c_{m'})\| + \|  \mu(c_{m'}) - \mu(c_{m_k})\| + \|  \mu(c_{m_k}) - \hat \mu_{m_k}\| \right)\\
    &\le  \min_{1\le k \le \bar k} \left ( 2\max_{m\in[M]} \|\hat \mu_m-\mu(c_m)\| + \|  \mu(c_{m'}) - \mu(c_{m_k})\|  \right) \\
    &= 2\max_{m\in[M]} \|\hat \mu_m-\mu(c_m)\| + \min_{1\le k \le \bar k}  \|  \mu(c_{m'}) - \mu(c_{m_k})\|  \\
    &= 2\max_{m\in[M]} \|\hat \mu_m-\mu(c_m)\|
\end{align*}
where the last equality results from the fact that at least one of the existing centers shares the same true cluster index as that of arm $m'$.

On the other hand, for any arm $m' \in [M]\setminus \mathcal A _{\bar k}$, the Euclidean distance to the nearest existing center can be lower bounded as follows:
\begin{align*}
    \min_{1\le k \le \bar k} \| \hat \mu_{m'} - \hat \mu(k)\|  &=  \min_{1\le k \le \bar k} \| \hat \mu_{m'} - \hat \mu_{m_k}\| \\
    &\ge \min_{1\le k \le \bar k} \left (-\| \hat \mu_{m'} - \mu(c_{m'})\| + \|  \mu(c_{m'}) - \mu(c_{m_k})\| - \|  \mu(c_{m_k}) - \hat \mu_{m_k}\| \right)\\
    &\ge  \min_{1\le k \le \bar k} \left ( -2\max_{m\in[M]} \|\hat \mu_m-\mu(c_m)\| + \|  \mu(c_{m'}) - \mu(c_{m_k})\|  \right) \\
    &= -2\max_{m\in[M]} \|\hat \mu_m-\mu(c_m)\| + \min_{1\le k \le \bar k}  \|  \mu(c_{m'}) - \mu(c_{m_k})\|  \\
    &\ge -2\max_{m\in[M]} \|\hat \mu_m-\mu(c_m)\| + \min_{{k,k'\in[K]: k\neq k'}} \|\mu(k)-\mu(k')\|
\end{align*}
where the last inequality follows from the fact that none of the existing centers shares the same true cluster index with arm $m'$. 

The above upper bound and lower bound, together with the constraint on the accuracy of the empirical estimates, show that the $\bar k$-th center must come from $[M]\setminus \mathcal A _{\bar k}$. Therefore,  Maximin Initialization succeeds in choosing $K$ centers from $K$ disjoint clusters.

\textbf{Second}, we prove that after the first step (Line~\ref{algo2_step1}) in the first iteration of Weighted $K$-means, $\hat c$ is a correct partition, i.e., $\hat c \sim c$. 

For any arm $m' \in [M]$, we can always find exactly one arm (denoted as $m_{\bar k}$) from $m_1, m_2,\ldots,m_{K}$ that shares the same true cluster index since $m_1, m_2,\ldots,m_{K}$ are selected from $K$ disjoint clusters. The distance between $\hat \mu_{m'}$ and $\hat \mu (\bar k)$ can be upper bounded as follows:
\begin{align*}
    \|\hat \mu_{m'} -\hat \mu(\bar k)\| &\le \|\hat \mu_{m'} -  \mu(c_{m'})\| + \| \mu(c_{m'}) -\hat \mu(\bar k)\|  \\
    &=  \|\hat \mu_{m'} -  \mu(c_{m'})\| + \| \mu(c_{m_{\bar k}})-\hat \mu_{m_{\bar k}}\|  \\
    &\le 2\max_{m\in[M]} \|\hat \mu_m-\mu(c_m)\|.
\end{align*}
However, with respect to the remaining cluster centers, we have 
\begin{align*}
    \min_{k\in[K]:k\neq\bar k} \|\hat \mu_{m'} -\hat \mu(k)\| &\ge  \min_{k\in[K]:k\neq\bar k}  \left(-\|\hat \mu_{m'} -  \mu(c_{m'})\| + \|\mu(c_{m'})- \mu(c_{m_k})\|  - \|\mu(c_{m_k})- \hat \mu(k)\|\right) \\
    &= \min_{k\in[K]:k\neq\bar k}  \left(-\|\hat \mu_{m'} -  \mu(c_{m'})\| + \|\mu(c_{m'})- \mu(c_{m_k})\|  - \|\mu(c_{m_k})- \hat \mu_{m_{ k}}\|\right)\\
    &\ge \min_{k\in[K]:k\neq\bar k}  \left( -2  \max_{m\in[M]} \|\hat \mu_m-\mu(c_m)\| + \|\mu(c_{m'})- \mu(c_{m_k})\|  \right) \\
    &=  -2  \max_{m\in[M]} \|\hat \mu_m-\mu(c_m)\| + \min_{k\in[K]:k\neq\bar k}   \|\mu(c_{m_{\bar k}})- \mu(c_{m_k})\|  \\
    &\ge -2  \max_{m\in[M]} \|\hat \mu_m-\mu(c_m)\| + \min_{{k,k'\in[K]: k\neq k'}} \|\mu(k)-\mu(k')\| \\
    &> -2\max_{m\in[M]} \|\hat \mu_m-\mu(c_m)\| + 4\max_{m\in[M]} \|\hat \mu_m-\mu(c_m)\|\\
    &=  2\max_{m\in[M]} \|\hat \mu_m-\mu(c_m)\| \\
    &\ge \|\hat \mu_{m'} -\hat \mu(\bar k)\|.
\end{align*}
Hence, $$\hat c _{m'} = \argmin_{k\in[K]} \|\hat \mu_{m'} -\hat \mu(k)\|= \bar k .$$

Since the arm $m'$ is arbitrary, the above argument shows those arms that share the same true cluster index still share the same cluster index in $\hat c$. Besides, there are $K$ disjoint non-empty clusters in $\hat c$. Therefore, $\hat c$ is a correct partition. 

\textbf{Finally}, we prove that the partition $\hat c$ no longer changes after the first iteration of Weighted $K$-means, which we term as the Clustering is {\em stabilized}. 

We need to show after the update (Line~\ref{algo2_step2}) of $\hat \mu(k)$, $\argmin_{k\in[K]} \|\hat \mu_m -\hat \mu(k)\|$ still returns $\hat c _m$ for any arm $m \in [M]$.

Since $\hat c \sim c$, there exists a permutation $\sigma$ on $[K]$ such that $c = \sigma(\hat c)$. For any cluster $\bar k \in [K]$,
\begin{align}
    \|\hat \mu(\bar k) - \mu (\sigma(\bar k ))\| &= \left \| \frac {\sum_{m\in[M]} N_m \hat \mu_m \mathbbm 1 \{ \hat c_m = \bar k \}} {\sum_{m\in[M]} N_m  \mathbbm 1 \{ \hat c_m = \bar k \}} -\mu (\sigma(\bar k )) \right\| \notag \\
    &= \left \| \frac {\sum_{m\in[M]} N_m  \mathbbm 1 \{  c_m = \sigma(\bar k) \} (\hat \mu_m - \mu (\sigma(\bar k )) )} {\sum_{m\in[M]} N_m  \mathbbm 1 \{ c_m = \sigma(\bar k) \}}  \right\| \notag \\
    &= \left \| \frac {\sum_{m\in[M]} N_m  \mathbbm 1 \{  c_m = \sigma(\bar k) \} (\hat \mu_m - \mu (c_m) )} {\sum_{m\in[M]} N_m  \mathbbm 1 \{ c_m = \sigma(\bar k) \}}  \right\|  \notag \\
    &\le \frac {\sum_{m\in[M]} N_m  \mathbbm 1 \{  c_m = \sigma(\bar k) \} \|\hat \mu_m - \mu (c_m) \|} {\sum_{m\in[M]} N_m  \mathbbm 1 \{ c_m = \sigma(\bar k) \}} \notag\\
    &\le \max_{m\in[M]} \|\hat \mu_m-\mu(c_m)\|.\label{equation_theorem_maximum}
\end{align}

Then for any arm $m' \in [M]$, the distance between $\hat \mu_{m'}$ and $\hat \mu (\hat c_{m'})$ can be upper bounded as follows
\begin{align*}
    \|\hat \mu_{m'} -\hat \mu(\hat c_{m'})\| &\le \|\hat \mu_{m'} -  \mu(c_{m'})\| + \| \mu(c_{m'}) -\hat \mu(\hat c_{m'})\|  \\
    &=  \|\hat \mu_{m'} -  \mu(c_{m'})\| + \| \mu(\sigma(\hat c_{m'}))-\hat \mu(\hat c_{m'})\|  \\
    &\le 2\max_{m\in[M]} \|\hat \mu_m-\mu(c_m)\|
\end{align*}
while the minimal distance between $\hat \mu_{m'}$ and other centers can be lower bounded as follows 
\begin{align*}
    &\phantom{=} \min_{k\in[K]:k\neq\hat c_{m'}} \|\hat \mu_{m'} -\hat \mu(k)\| \\ &\ge  \min_{k\in[K]:k\neq\hat c_{m'}}  \left(-\|\hat \mu_{m'} -  \mu(c_{m'})\| + \|\mu(c_{m'})- \mu(\sigma(k))\|  - \|\mu(\sigma(k))- \hat \mu(k)\|\right) \\
    &= \min_{k\in[K]:k\neq\hat c_{m'}}  \left(-\|\hat \mu_{m'} -  \mu(c_{m'})\| + \|\mu(\sigma(\hat c_{m'}))- \mu(\sigma(k))\|  - \|\mu(\sigma(k))- \hat \mu(k)\|\right)       \\
    &\ge \min_{k\in[K]:k\neq\hat c_{m'}}  \left( -2  \max_{m\in[M]} \|\hat \mu_m-\mu(c_m)\| + \|\mu(\sigma(\hat c_{m'}))- \mu(\sigma(k))\|  \right) \\
    &=  -2  \max_{m\in[M]} \|\hat \mu_m-\mu(c_m)\| + \min_{k\in[K]:k\neq\hat c_{m'}}   \|\mu(\sigma(\hat c_{m'}))- \mu(\sigma(k))\| \\
    &\ge -2  \max_{m\in[M]} \|\hat \mu_m-\mu(c_m)\| + \min_{{k,k'\in[K]: k\neq k'}} \|\mu(k)-\mu(k')\| \\
    &>  -2\max_{m\in[M]} \|\hat \mu_m-\mu(c_m)\| + 4\max_{m\in[M]} \|\hat \mu_m-\mu(c_m)\|  \\
    &=  2\max_{m\in[M]} \|\hat \mu_m-\mu(c_m)\| \\
    &\ge  \|\hat \mu_{m'} -\hat \mu(\hat c_{m'})\| .
\end{align*}

Therefore, $\argmin_{k\in[K]} \|\hat \mu_{m'} -\hat \mu(k)\|$ remains equal to $\hat c _{m'}$. Since the arm $m'$ is arbitrary, the partition is proved to be stabilized after the first iteration of Weighted $K$-means.

Moreover, $\hat \mu (k)$ for all $ k \in [K]$ will stay the same once the partition stabilizes. Hence, by Equation~\eqref{equation_theorem_maximum},
$$\max_{k\in[K]}\|\hat \mu(k) - \mu(\sigma(k))\| \le \max_{m\in[M]} \|\hat \mu_m-\mu(c_m)\|.$$

Now the proof of Proposition~\ref{theorem_maximum} is completed.
\end{proof}

\subsection{Proof of Proposition~\ref{prop_sampling_rule}}
\label{appendix_proof_prop_sampling_rule}
\begin{proof}%{Proof of Proposition~\ref{prop_sampling_rule}.}
Due to the forced exploration in Algorithm~\ref{algo1} and the strong law of large numbers, for all $ m\in [M]$, $\hat \mu_m(t)$ converges almost surely to $\mu(c_m)$, i.e., $\hat \mu_m(t) \as \mu(c_m)$, as $t$ tends to infinity.

In the following, we condition on the event 
$$
\mathcal E = \left \{ \lim_{t\to \infty} \hat \mu_m(t) = \mu(c_m) \text{ for all }  m\in [M] \right\}
$$
which has probability $1$.

Note that in finite-dimensional spaces, pointwise convergence and convergence in Euclidean norm are equivalent. Therefore, by Proposition~\ref{theorem_maximum}, we know that for sufficiently large $t$, Algorithm~\ref{algo2} will output a correct partition $c^t$ such that $c^t = \sigma^t(c)$ for some permutation $
\sigma^t$ on $[K]$; moreover, $\mu^t(\sigma^t(k)) \to \mu(k)$ for all $ k\in [K]$. Therefore, $\mathrm{Alt}(c^t) = \mathrm{Alt}(c)$ for sufficiently large $t$, and $\sigma^t(\mathcal U^t) \to \mathcal U$. Hence, for sufficiently large $t$, 
\begin{align*}
    \lambda^*(t) &= \argmax_{\lambda \in \mathcal P _{M}} \inf_{(c',\mathcal{U}')\in \mathrm{Alt}(c^t)} \sum_{m=1}^{M} \lambda_m\|\mu^t(c^t_m)-\mu'(c'_m)\|^2 \\
    &= \argmax_{\lambda \in \mathcal P _{M}} \inf_{(c',\mathcal{U}')\in \mathrm{Alt}(c)} \sum_{m=1}^{M} \lambda_m\|\mu^t(\sigma^t(c_m))-\mu'(c'_m)\|^2.
\end{align*}
By Proposition~\ref{prop_continuity2},
\begin{align*}
    \lambda^*(t) 
    &\to \argmax_{\lambda \in \mathcal P _{M}} \inf_{(c',\mathcal{U}')\in \mathrm{Alt}(c)} \sum_{m=1}^{M} \lambda_m\|\mu(c_m)-\mu'(c'_m)\|^2  \\
    &= \lambda^*.
\end{align*}

Consequently,  $\lambda^*(t)$ converges pointwisely to $\lambda^*$. That is to say, for any $\epsilon > 0$, there exists 
$t_0(\epsilon)$ such that 
$$
\sup_{t>t_0(\epsilon)} \max_{m\in[M]} |\lambda^*_m(t)- \lambda^*_m| \le \frac {\epsilon} {3(M-1)}.
$$
By Lemma~\ref{lemma_sampling_rule}, there further exists $t_1(\epsilon) \ge t_0(\epsilon)$ such that 
$$
\sup_{t>t_1(\epsilon)} \max_{m\in[M]} \left |\frac {N_m(t)} t- \lambda^*_m \right | \le \epsilon .
$$
This is also equivalent to  
$$
\lim_{t\rightarrow \infty} \frac {N_m(t)} t = \lambda^*_m \;\text{ for all }\;  m\in [M]
$$
as desired.

\end{proof}

\subsection{Proof of Proposition~\ref{theorem_stopping_rule}}
\label{appendix_proof_theorem_stopping_rule}
\begin{proof}%{Proof of Proposition~\ref{theorem_stopping_rule}.}
Recall the definitions of $Z(t)$, $Z_1(t)$, and $Z_2(t)$ from Section~\ref{subsection_stopping_rule} or Algorithm~\ref{algo1}.

%Let us recall the definition of $Z(t)$:
%$$
%Z(t) = \frac 1 2 \left(\left(-\sqrt{Z_1(t)}+\sqrt{Z_2(t)}\right)_+\right)^2
%$$
%with
%$$
%Z_1(t) = \sum_{m=1}^{M} N_m(t)\|\hat \mu_m(t)-\mu^{t-1}(c^{t-1}_m)\|^2
%$$
%and 
%$$
%Z_2(t) = \min_{(c',\mathcal{U}')\in \mathrm{Alt}(c^{t-1})} \sum_{m=1}^{M} N_m %(t)\|\mu^{t-1}(c^{t-1}_m)-\mu'(c'_m)\|^2 .
%$$

We can then bound the error probability as follows: 
\begin{align}
     &\phantom{\;=\;} \Pr(\tau_\delta<\infty, c^{\mathrm{out}} \not \sim c) \notag \\
     &\le \Pr(\exists\,  t \in \mathbb N: c^{t-1}\not\sim c, Z(t)\ge \beta(\delta,t )) \notag \\
     &= \Pr\left(\exists\,  t \in \mathbb N: c^{t-1}\not\sim c, \frac 1 2 \max_{\alpha \ge 0} \left( -\alpha Z_1(t) + \frac{\alpha} {\alpha+1}Z_2(t) \right) \ge \beta(\delta,t ) \right) 
     \label{theorem_stopping_rule_0}\\
     &\le \Pr\left(\exists\,  t \in \mathbb N: \frac 1 2\max_{\alpha \ge 0} \left( -\alpha Z_1(t) + \frac{\alpha} { \alpha +1} \sum_{m=1}^{M} N_m (t)\|\mu^{t-1}(c^{t-1}_m)-\mu(c_m)\|^2  \right) \ge \beta(\delta,t )\right) 
     \label{theorem_stopping_rule_1}\\
     &\le \Pr\left (\exists\,  t \in \mathbb N:  \frac 1 2\sum_{m=1}^{M} N_m (t)\|\hat \mu_m(t) -\mu(c_m)\|^2 \ge \beta(\delta,t ) \right) \label{theorem_stopping_rule_2}\\
     &\le \delta.\label{theorem_stopping_rule_3}
\end{align}

Line \eqref{theorem_stopping_rule_0} follows from Lemma~\ref{lemma_triangular1}.

Line \eqref{theorem_stopping_rule_1} follows from the fact that if $c^{t-1}\not\sim c$, then $(c,\mathcal{U}) \in \mathrm{Alt}(c^{t-1})$ and hence
$$
Z_2(t) \le \sum_{m=1}^{M} N_m (t)\|\mu^{t-1}(c^{t-1}_m)-\mu(c_m)\|^2 .
$$

Line \eqref{theorem_stopping_rule_2} follows from Lemma~\ref{lemma_triangular2} and thus
\begin{align*}
    &\phantom{\;=\;}  \max_{\alpha \ge 0} \left( -\alpha Z_1(t) + \frac{\alpha} {\alpha+1} \sum_{m=1}^{M} N_m (t)\|\mu^{t-1}(c^{t-1}_m)-\mu(c_m)\|^2  \right) 
    \\
    &=  \max_{ \alpha \ge 0} \left(  \sum_{m=1}^{M} N_m (t) \left(-\alpha\|\hat \mu_m(t)-\mu^{t-1}(c^{t-1}_m)\|^2 + \frac{\alpha} {\alpha+1}\|\mu^{t-1}(c^{t-1}_m)-\mu(c_m)\|^2 \right) \right) \\
    &\le \max_{\alpha \ge 0} \left(  \sum_{m=1}^{M} N_m (t)  \|\hat \mu_m(t) -\mu(c_m)\|^2  \right) \\
    &=  \sum_{m=1}^{M} N_m (t)  \|\hat \mu_m(t) -\mu(c_m)\|^2  .
\end{align*}

Line \eqref{theorem_stopping_rule_3} follows from Lemma~\ref{lemma_beta}.

\end{proof}

\subsection{Proof of Proposition~\ref{theorem_complexity1}}
\label{appendix_proof_theorem_complexity1}
\begin{proof}%{Proof of Proposition~\ref{theorem_complexity1}.}
Similarly to the proof of Proposition~\ref{prop_sampling_rule}, in the following, we condition on the event 
$$
\mathcal E = \left \{ \lim_{t\to \infty} \hat \mu_m(t) = \mu(c_m) \text{ for all }  m\in [M] \right\}
$$
which has probability $1$. 

Note that in finite-dimensional spaces, pointwise convergence and convergence in Euclidean norm are equivalent. By Proposition~\ref{theorem_maximum}, there exists $t_1>0$ such that for all $t\ge t_1$, Algorithm~\ref{algo2} will output a correct partition $c^t = \sigma^t(c)$ for some permutation $
\sigma^t$ on $[K]$; moreover, $\mu^t(\sigma^t(k)) \to \mu(k)$ for all $ k\in [K]$. Therefore, for all $ m\in [M]$, $\mu^{t-1}(c^{t-1}_m) \to \mu(c_m)$. 

Accordingly, on the event $\mathcal E$, as $t$ tends to infinity, $\|\hat \mu_m(t)-\mu^{t-1}(c^{t-1}_m)\|^2\to 0$ for all $ m\in [M]$. Since $N_m (t)/t$ is uniformly bounded by $1$ for all $ m\in [M]$, we have 
$$
\frac {Z_1(t)} t  = \sum_{m=1}^{M} \frac {N_m(t)} t \|\hat \mu_m(t)-\mu^{t-1}(c^{t-1}_m)\|^2 \to 0.
$$

Now we consider $Z_2(t)/t$.  By Proposition~\ref{prop_sampling_rule} and its proof, conditioned on the event $\mathcal E$, ${N_m(t)} /t \to \lambda^*_m $ for all $ m\in [M]$.

When $t> t_1$, $c^{t-1}$ is always correct (i.e., $c^{t-1}\sim c$) and hence $\mathrm{Alt}(c^{t-1}) = \mathrm{Alt}(c)$. Thus, for $t> t_1$, 
\begin{align*}
    \frac {Z_2(t)} t &=   \min_{(c',\mathcal{U}')\in \mathrm{Alt}(c^{t-1})} \sum_{m=1}^{M} \frac {N_m(t)} t\|\mu^{t-1}(c^{t-1}_m)-\mu'(c'_m)\|^2\\
    &= \min_{(c',\mathcal{U}')\in \mathrm{Alt}(c)} \sum_{m=1}^{M} \frac {N_m(t)} t\|\mu^{t-1}(\sigma^{t-1}(c_m))-\mu'(c'_m)\|^2.
\end{align*}

By Proposition~\ref{prop_continuity1}, as $t$ tends to infinity,
\begin{align*}
    \frac {Z_2(t)} t &\to  \min_{(c',\mathcal{U}')\in \mathrm{Alt}(c)} \sum_{m=1}^{M}  \lambda^*_m  \|\mu(c_m)-\mu'(c'_m)\|^2 \\
    &= 2D^*(c,\mathcal{U})^{-1}.
\end{align*}

Consequently, 
$$
\frac {Z(t)} t = \frac 1 2 \left(\left(-\sqrt{\frac {Z_1(t)} t}+\sqrt{\frac {Z_2(t)} t}\right)_+\right)^2 \to D^*(c,\mathcal{U})^{-1}. 
$$

So for any $0<\epsilon<1$, there exists $t_2>t_1$ such that for all $t\ge t_2$, 
$$\frac  {Z(t)} t \ge (1-\epsilon) D^*(c,\mathcal{U})^{-1}.$$

Now consider $\beta(\delta,t )$. Since $N_m(t) \le t$ for $ m\in [M]$, it holds that there exists $t_3>0$ such that for all $t \ge t_3$, 
$$
 \sum_{m=1}^{M} 2 d \log(4 + \log(N_m (t))) \le \log(t) 
$$
which implies that 
$$
\beta(\delta,t ) \le \log(t) + Md \cdot \Psi\left (\frac {\log(1/\delta)} {Md} \right).
$$

Altogether, we have 
\begin{align*}
    \tau_\delta &= \inf\{t\in \mathbb N:Z(t)\ge \beta(\delta,t )\} \\
    &\le \max\{t_2,t_3\} \vee \inf \left \{t\in \mathbb N: t(1-\epsilon) D^*(c,\mathcal{U})^{-1} \ge \log(t) + Md \cdot \Psi \left (\frac {\log(1/\delta)} {Md}\right)\right \} 
\end{align*}
which shows $\tau_\delta$ is finite conditioned on $\mathcal E$. Since $\Pr(\mathcal E) = 1$, we have 
$$
\Pr(\tau_\delta<\infty) = 1.
$$

For ease of notation, let $a := {(1-\epsilon) D^*(c,\mathcal{U})^{-1}}$ and $b := Md \cdot \Psi \left (\frac {\log(1/\delta)} {Md}\right)$. For sufficiently small $\delta$, since $b\ge \log (1/\delta)$, $b+ \log\left( \frac 1 a \right) > 0 $. Thus, by Lemma~\ref{lemma_log}, 
\begin{align*}
    \tau_\delta 
    &\le \max \left\{t_2, t_3, \frac 1 a \left( b + \log\left( \frac e a \right)+ \log \left( b+ \log\left( \frac 1 a \right) \right)\right) \right\}.
\end{align*}

Note that $t_2$ and $t_3$ do not depend on $\delta$ and $\Psi (x)  = x+\log (x)+ o(\log (x))$ as $x\to \infty$. Thus, 
\begin{align*}
    \limsup_{\delta \rightarrow 0 } \frac {\tau_\delta} {\log(1/\delta)} &\le  \limsup_{\delta \rightarrow 0 }  \frac 1 {a\log(1/\delta)} \left( b + \log\left( \frac e a \right)+ \log \left( b+ \log\left( \frac 1 a \right) \right)\right)  \\
    &= \frac 1 a \\
    &= {(1-\epsilon)^{-1} D^*(c,\mathcal{U})}.
\end{align*}

Since the above inequality holds for any $0< \epsilon < 1$, by letting $\epsilon \to 0^+$,
\begin{align*}
    \limsup_{\delta \rightarrow 0 } \frac {\tau_\delta} {\log(1/\delta)} &\le  D^*(c,\mathcal{U}).
\end{align*}

\end{proof}

\subsection{Proof of Theorem~\ref{theorem_complexity2}}
\label{appendix_proof_theorem_complexity2}
\begin{proof}%{Proof of Theorem~\ref{theorem_complexity2}.}
Let $0< \epsilon <1$. 

First, we consider the sampling rule. According to Proposition~\ref{theorem_maximum} and Proposition~\ref{prop_continuity2}, there exists a function 
$$
\xi: (0,1) \to \left(0, \frac 1 4 \min_{{k,k'\in[K]: k\neq k'}} \|\mu(k)-\mu(k')\|\right)
$$
such that $\lim_{\epsilon \to 0^+} \xi(\epsilon) = 0 $ and for any $t > 0$, if 
$$
\max_{m\in[M]} \|\hat \mu_m(t)-\mu(c_m)\| \le \xi(\epsilon),
$$
then:
\begin{enumerate}[label = (\roman*)]
    \item Algorithm~\ref{algo2} outputs a correct partition $c^t$ such that $c^t = \sigma^t(c)$ for some permutation $\sigma^t$ on~$[K]$;
    \item $\mathcal U ^t$ satisfies that $$\max_{k\in[K]}\|\mu^{t}(\sigma^t(k)) - \mu(k)\| \le   \xi(\epsilon);$$
    \item $\lambda^*(t)$ satisfies that
    $$
    \max_{m\in[M]} |\lambda^*_m(t)- \lambda^*_m|  \le \epsilon.
    $$
\end{enumerate}

Point (iii) is, in fact, a consequence of the continuity of $\lambda^*(t)$ (see Proposition~\ref{prop_continuity2}):
\begin{align*}
    \lim_{\sigma^t(\mathcal U^t) \to \mathcal U} \lambda^*(t) &= \lim_{\sigma^t(\mathcal U^t) \to \mathcal U}   \argmax_{\lambda \in \mathcal P _{M}} \inf_{(c',\mathcal{U}')\in \mathrm{Alt}(c^t)} \sum_{m=1}^{M} \lambda_m\|\mu^t(c^t_m)-\mu'(c'_m)\|^2 \\
    &= \lim_{\sigma^t(\mathcal U^t) \to \mathcal U}  \argmax_{\lambda \in \mathcal P _{M}} \inf_{(c',\mathcal{U}')\in \mathrm{Alt}(c)} \sum_{m=1}^{M} \lambda_m\|\mu^t(\sigma^t(c_m))-\mu'(c'_m)\|^2 \\
    &= \argmax_{\lambda \in \mathcal P _{M}} \inf_{(c',\mathcal{U}')\in \mathrm{Alt}(c)} \sum_{m=1}^{M} \lambda_m\|\mu(c_m)-\mu'(c'_m)\|^2  \\
    &= \lambda^*.
\end{align*}

In addition, 
let $T  \ge \lceil  {M} /{\epsilon^3}\rceil  \in \mathbb N$ and define the event 
$$
\mathcal{E}_{T}(\epsilon) = \bigcap_{t=\lfloor T \epsilon^3\rfloor }^{\infty} \left \{  \max_{m\in[M]} \|\hat \mu_m(t)-\mu(c_m)\| \le \xi(\epsilon) \right\}.
$$

Conditioned on the event $\mathcal{E}_{T}(\epsilon)$, by Lemma~\ref{lemma_sampling_rule} and the definition of $\xi(\epsilon)$, for all $t \ge T$,
$$
\max_{m\in[M]} \left |\frac {N_m(t)} t- \lambda^*_m \right | \le 3(M-1)\epsilon.
$$

Now we introduce $D^*_{\epsilon}(c,\mathcal{U})^{-1}$, which is an $\epsilon$-approximation of $D^*(c,\mathcal{U})^{-1}$, and defined as 
% \begin{align*}
%     D^*_{\epsilon}(c,\mathcal{U})^
%     {-1} = \frac 1 2 \inf_{\substack{   \max\limits_{m\in[M]} \left | \hat \lambda _m - \lambda^*_m \right | \le 3(M-1)\epsilon,  \\  \max\limits_{k\in[K]}\|\hat \mu(k) - \mu(\sigma(k))\| \le   \xi(\epsilon)   }} \min_{(c',\mathcal{U}')\in \mathrm{Alt}(c^{t-1})} \sum_{m=1}^{M} \hat \lambda _m \|\hat \mu(c^{t-1}_m)-\mu'(c'_m)\|^2 .
% \end{align*}
\begin{align*}
    D^*_{\epsilon}(c,\mathcal{U})^{-1} &:= \frac 1 2 \inf_{\hat \lambda, \hat {\mathcal U}} \min_{(c',\mathcal{U}')\in \mathrm{Alt}(c)} \sum_{m=1}^{M} \hat \lambda _m \|\hat \mu(c_m)-\mu'(c'_m)\|^2  
\end{align*}
where the infimum is over all $\hat \lambda$ and $\hat {\mathcal U}$ such that
\begin{align*}
    \begin{cases}
    \max\limits_{m\in[M]} \left | \hat \lambda _m - \lambda^*_m \right | \le 3(M-1)\epsilon, \\   \max\limits_{k\in[K]}\|\hat \mu(k) - \mu(k)\| \le   \xi(\epsilon).
    \end{cases}
\end{align*}

By the continuity of $\epsilon \mapsto D^*_{\epsilon}(c,\mathcal{U})^{-1}$ at $0$ (which is a consequence of Proposition~\ref{prop_continuity1}), 
\begin{align*}
\lim_{\epsilon \to 0^+} D^*_{\epsilon}(c,\mathcal{U})^{-1} 
&=\frac 1 2 \min_{(c',\mathcal{U}')\in \mathrm{Alt}(c)} \sum_{m=1}^{M}  \lambda^*_m  \|\mu(c_m)-\mu'(c'_m)\|^2 \\
&= D^*(c,\mathcal{U})^{-1}.
\end{align*}

Therefore, conditioned on the event $\mathcal{E}_{T}(\epsilon)$, for all $t >T$, $Z_2(t) / t \ge 2 D^*_{\epsilon}(c,\mathcal{U})^{-1}$. Concerning $Z_1(t)$, conditioned on the event $\mathcal{E}_{T}(\epsilon)$, for all $t >T$, using the inequality in Lemma~\ref{lemma_triangular2} (with $\alpha = 1$),
\begin{align*}
    \frac {Z_1(t)} t  &= \sum_{m=1}^{M} \frac {N_m(t)} t \|\hat \mu_m(t)-\mu^{t-1}(c^{t-1}_m)\|^2 \\
    &\le \sum_{m=1}^{M} \frac {2N_m(t)} t \left( \|\hat \mu_m(t)-\mu(c_m)\|^2 + \| \mu(c_m)-\mu^{t-1}(c^{t-1}_m)\|^2\right) \\
    & = \sum_{m=1}^{M} \frac {2N_m(t)} t \left( \|\hat \mu_m(t)-\mu(c_m)\|^2 + \| \mu(c_m)-\mu^{t-1}(\sigma^{t-1}(c_m))\|^2\right) \\
    &\le \sum_{m=1}^{M} \frac {2N_m(t)} t \left(\max_{m\in[M]} \|\hat \mu_m(t)-\mu(c_m)\| + \max_{k\in[K]}\|\mu^{t-1}(\sigma^{t-1}(k)) - \mu(k)\| \right) \\
    &\le 4 \xi(\epsilon).
\end{align*}

Altogether, conditioned on $\mathcal{E}_{T}(\epsilon)$, for all $t > T$, %taking sufficiently small $\epsilon$,
\begin{align*}
    \frac {Z(t)} t &= \frac 1 2 \left(\left(-\sqrt{\frac {Z_1(t)} t}+\sqrt{\frac {Z_2(t)} t}\right)_+\right)^2 \\
    &\ge \left(\left(\sqrt{ D^*_{\epsilon}(c,\mathcal{U})^{-1}} - \sqrt{2\xi(\epsilon)}\right)_+\right)^2.
\end{align*}

Then consider $\beta(\delta,t )$. Since $N_m(t) \le t$ for $ m\in [M]$, it holds that there exists $t_2>0$ such that for all $t \ge t_2$, 
$$
 \sum_{m=1}^{M} 2 d \log(4 + \log(N_m (t))) \le \log(t) 
$$
which implies that 
$$
\beta(\delta,t ) \le \log(t) + Md \cdot \Psi\left (\frac {\log(1/\delta)} {Md} \right).
$$

For ease of notation, let $a := \left(\left(\sqrt{ D^*_{\epsilon}(c,\mathcal{U})^{-1}} - \sqrt{2\xi(\epsilon)}\right)_+\right)^2 $ and $b := Md \cdot \Psi \left (\frac {\log(1/\delta)} {Md}\right)$. Consequently, conditioned on $\mathcal{E}_{T}(\epsilon)$, 
\begin{align*}
    \tau_\delta &= \inf\{t\in \mathbb N:Z(t)\ge \beta(\delta,t )\} \\
    &\le \max\{T+1,t_2\} \vee \inf \left \{t\in \mathbb N: at \ge \log(t) + b \right \}. 
\end{align*}

Let $t_3:= \max\{ \lceil  {M} /{\epsilon^3}\rceil +1 ,t_2\} \vee \inf \left \{t\in \mathbb N: at \ge \log(t) + b \right \}$. Then for all $T+1 \ge t_3$, $\mathcal{E}_{T}(\epsilon)\subseteq \{ \tau_\delta \le T+1 \}$ and hence $\Pr(\tau_\delta > T+1) \le \Pr( \mathcal{E}_{T}^c(\epsilon))$. Therefore, $\E[\tau_\delta]$ can be upper bounded as follows:
\begin{align*}
  \E[\tau_\delta] &= \sum_{t=0}^{\infty} \Pr(\tau_\delta > t) \\
  &= \sum_{t=0}^{t_3-1} \Pr(\tau_\delta > t) + \sum_{t=t_3}^{\infty} \Pr(\tau_\delta > t) \\
  &\le t_3 + \sum_{T=t_3-1}^{\infty} \Pr( \mathcal{E}_{T}^c(\epsilon))\\
  &\le t_3 + \sum_{T=\lceil {M} /{\epsilon^3}\rceil}^{\infty} \Pr( \mathcal{E}_{T}^c(\epsilon)).
\end{align*}

Since $\lim_{\epsilon \to 0^+} a = D^*(c,\mathcal{U})^{-1}$, $a>0$ for sufficiently small $\epsilon$. Together with $b\ge \log (1/\delta)$, we have $b+ \log\left( \frac 1 a \right) > 0 $ for sufficiently small $\delta$. Thus, by Lemma~\ref{lemma_log}, 
\begin{align*}
    \E[\tau_\delta]
    &\le \max \left\{\left \lceil\frac {M} {\epsilon^3}\right \rceil+1,t_2, \frac 1 a \left( b + \log\left( \frac e a \right)+ \log \left( b+ \log\left( \frac 1 a \right) \right)\right) \right\} +  \sum_{T=\lceil {M} /{\epsilon^3}\rceil}^{\infty} \Pr( \mathcal{E}_{T}^c(\epsilon)).
\end{align*}

For the moment, let us assume that the final term $\sum_{T=\lceil {M} /{\epsilon^3}\rceil}^{\infty} \Pr( \mathcal{E}_{T}^c(\epsilon))$ is finite. Note that $\lceil {M} /{\epsilon^3}\rceil$, $t_2$ and $\sum_{T=\lceil {M} /{\epsilon^3}\rceil}^{\infty} \Pr( \mathcal{E}_{T}^c(\epsilon))$ do not depend on $\delta$ and $\Psi (x)  = x+\log (x)+ o(\log (x))$ as $x\to \infty$. Thus, 
\begin{align*}
    \limsup_{\delta \rightarrow 0 } \frac {\E[\tau_\delta]} {\log(1/\delta)} &\le  \limsup_{\delta \rightarrow 0 }  \frac 1 {a\log(1/\delta)} \left( b + \log\left( \frac e a \right)+ \log \left( b+ \log\left( \frac 1 a \right) \right)\right)  \\
    &= \frac 1 a \\
    &= \left(\left(\sqrt{ D^*_{\epsilon}(c,\mathcal{U})^{-1}} - \sqrt{\xi(\epsilon)}\right)_+\right)^{-2}.
\end{align*}

Since the above inequality holds for any $\epsilon >0$ such that $a > 0$, by letting $\epsilon \to 0^+$,
\begin{align*}
    \limsup_{\delta \rightarrow 0 } \frac {\tau_\delta} {\log(1/\delta)} &\le  D^*(c,\mathcal{U}).
\end{align*}

It remains to show that $\sum_{T=\lceil {M} /{\epsilon^3}\rceil}^{\infty} \Pr( \mathcal{E}_{T}^c(\epsilon))$ is finite. Since 
$$
\mathcal{E}_{T}^c(\epsilon) = \bigcup_{t=\lfloor T \epsilon^3\rfloor }^{\infty} \left \{  \max_{m\in[M]} \|\hat \mu_m(t)-\mu(c_m)\| > \xi(\epsilon) \right\},
$$
by a union bound, we have 
\begin{align*}
    \Pr( \mathcal{E}_{T}^c(\epsilon)) &\le \sum_{t=\lfloor T \epsilon^3\rfloor }^{\infty} \sum_{m\in[M]} \Pr(\|\hat \mu_m(t)-\mu(c_m)\| > \xi(\epsilon) )\\
    &\le \sum_{t=\lfloor T \epsilon^3\rfloor }^{\infty} \sum_{m\in[M]} \Pr\left(\exists\,  i\in[d]: |[\hat \mu_m(t)-\mu(c_m)]_i| > \frac {\xi(\epsilon)} {\sqrt d}\right)\\
    &\le \sum_{t=\lfloor T \epsilon^3\rfloor }^{\infty} \sum_{m\in[M]} \sum_{i\in[d]}\Pr\left(\left|[\hat \mu_m(t)-\mu(c_m)]_i\right| > \frac {\xi(\epsilon)} {\sqrt d}\right),
\end{align*}
where we use $[x]_i$ to denote the $i^{\mathrm{th}}$ component of the vector $x \in \mathbb R^d$.

Considering sufficiently large $T$, for any $t \ge \lfloor T \epsilon^3\rfloor$,  $N_m (t) \ge \sqrt{t}/2$  for all $ m \in [M]$. Then, by Hoeffding's inequality for sub-Gaussian random variables, 
\begin{align*}
    \Pr( \mathcal{E}_{T}^c(\epsilon)) &\le 2 M d\sum_{t=\lfloor T \epsilon^3\rfloor }^{\infty} \exp \left(-\frac { \xi(\epsilon)^2\sqrt{t} } d \right) \\
    &\le 2 M d \int_{x =\lfloor T \epsilon^3\rfloor -1  }^{\infty}\exp \left(-\frac {\xi(\epsilon)^2 \sqrt{x}} d \right)\, \mathrm{d} x\\
    &= \frac {4Md^3} {\xi(\epsilon)^4} \left(\frac {\xi(\epsilon)^2 \sqrt{\lfloor T \epsilon^3\rfloor -1 }} d +1 \right)\exp \left(-\frac { \xi(\epsilon)^2 \sqrt{\lfloor T \epsilon^3\rfloor -1 } } d \right) \\
    &\le \frac {4Md^3} {\xi(\epsilon)^4} \left(\frac {\xi(\epsilon)^2 \sqrt{T \epsilon^3 }} d +1 \right)\exp \left(-\frac { \xi(\epsilon)^2 \sqrt{T \epsilon^3 } } {2d} \right) .
\end{align*}

Since for any constant $c>0$, the infinite sum $\sum_{n=1}^{\infty} \sqrt{n} \exp(-c\sqrt{n})$ is convergent, $\sum_{T=\lceil {M} /{\epsilon^3}\rceil}^{\infty} \Pr( \mathcal{E}_{T}^c(\epsilon))$ is also convergent.

This completes the proof of Theorem~\ref{theorem_complexity2}.\end{proof}

\section{Additional Numerical Results on a Higher-Dimensional Dataset}
\label{appendix_experiment}
In this appendix, we extend our numerical experiments to  the MNIST dataset \citep{lecun1998gradient}, which comprises data points of higher dimensions. Specifically, we conduct our online clustering task as follows. First, we randomly sample $500$ images of digits from the dataset, vectorize them, and, akin to \citet{sheikh20}, we employ Principal Component Analysis (PCA) to project the data onto different number of dimensions. Subsequently, we preprocess the dataset consistently with our approach to the Iris and Yeast datasets (see Section~\ref{section_exp}). This yields the number of clusters $K = 10$ and the number of arms $M = 500$, with the problem dimension $d$ set to be in $\{10, 50, 100, 200, 400\}$.  Note that the original dimensionality of the data points in the MNIST dataset is $784=28\times 28$.

The sample complexities for confidence level $\delta = 0.01$ are presented in Table~\ref{table1}.  It is evident from the table that \textsc{BOC}  outperforms the non-adaptive baseline method, \textsc{Uniform}, by a factor of $\approx 2$ in terms of their sample complexities, underscoring the advantage of employing an adaptive sampling approach, even on this dataset with large dimensionality.

    \begin{table}[t]
    	\begin{center}  
    		\begin{tabular}{crr}
    			\toprule
    		\multicolumn{1}{c}{$d$}	& \multicolumn{1}{c}{\textsc{BOC}}  & \multicolumn{1}{c}{\textsc{Uniform}}   \\
    			\midrule
    			{$10$} &  $2932.0 \, \pm \, 52.4 \phantom{1}$ & $6991.8 \, \pm \, 120.8 $    \\ 
    			{$50$} &  $8782.6 \, \pm \, 110.8$ & $16418.7 \, \pm \, 173.8 $    \\
                    {$100$} &  $16663.2 \, \pm \, 162.8$ & $31270.8 \, \pm \, 273.5 $  \\
                    {$200$} &  $32840.5 \, \pm \, 138.8$ & $61543.7 \, \pm \, 275.3 $  \\
                    {$400$} &  $ 65414.2 \, \pm \, 209.6 $ & $ 122508.3 \, \pm \, 369.9 $  \\
    			\bottomrule
    		\end{tabular}
    	\end{center}
    	\caption{The  averaged empirical sample complexities of \textsc{BOC} and \textsc{Uniform} for $\delta =0.01$ on the MNIST dataset.}
     \label{table1}
    \end{table}

\vskip 0.4in
\bibliography{references}

\end{document}